\documentclass[ruledheader,normaltoc]{abnt}
\usepackage[brazilian]{babel}
\usepackage[utf8]{inputenc}
\usepackage[none]{hyphenat}
\usepackage[T1]{fontenc}
\usepackage[alf,abnt-etal-list=0,abnt-etal-text=it,abnt-repeated-author-omit=yes,bibjustif]{abntcite}
\usepackage{abnt-alf}
\usepackage[ruled,portuguese]{algorithm2e}
\usepackage{dissertacao}
\usepackage{cite}
\usepackage{acronym}
\usepackage{etex}
\usepackage{latexsym}
\usepackage{graphicx}
\usepackage{chngcntr}
\counterwithout{footnote}{chapter}

\usepackage{blindtext}
\usepackage{sectsty} 
\sectionfont{\normalsize\bfseries} 
\subsectionfont{\normalsize\bfseries} 

\pdfhorigin=\dimexpr1in+(210mm-8.5in)/2\relax
\pdfvorigin=\dimexpr1in+(297mm-11in)/2\relax

\newfloat{figura}{!htb}{fig}[chapter]
\floatname{figura}{Figura}
\newfloat{grafico}{!htb}{graf}[chapter]
\floatname{grafico}{Gráfico}
\newfloat{tabela}{!htb}{tab}[chapter]
\floatname{tabela}{Tabela}
\usepackage[Conny]{fncychap}
\makeatletter
\ChNameVar{\fontsize{12}{12}\bfseries\centering}
\ChNumVar{\fontsize{12}{12}\rm}
\ChTitleVar{\fontsize{12}{12}\bfseries\centering}
\ChRuleWidth{2pt}
\makeatother

\usepackage{dropping}
\usepackage{rotating}

\usepackage{xcolor}
\usepackage{tikz}
\usepackage[a4paper,left=3cm,right=2cm,top=3cm,bottom=2cm]{geometry}
\begin{document}

\thispagestyle{empty}%
\definecolor{oneblue}{rgb}{0.72,0.88,1.1}
\begin{center}
    \includegraphics[width=2.5cm]{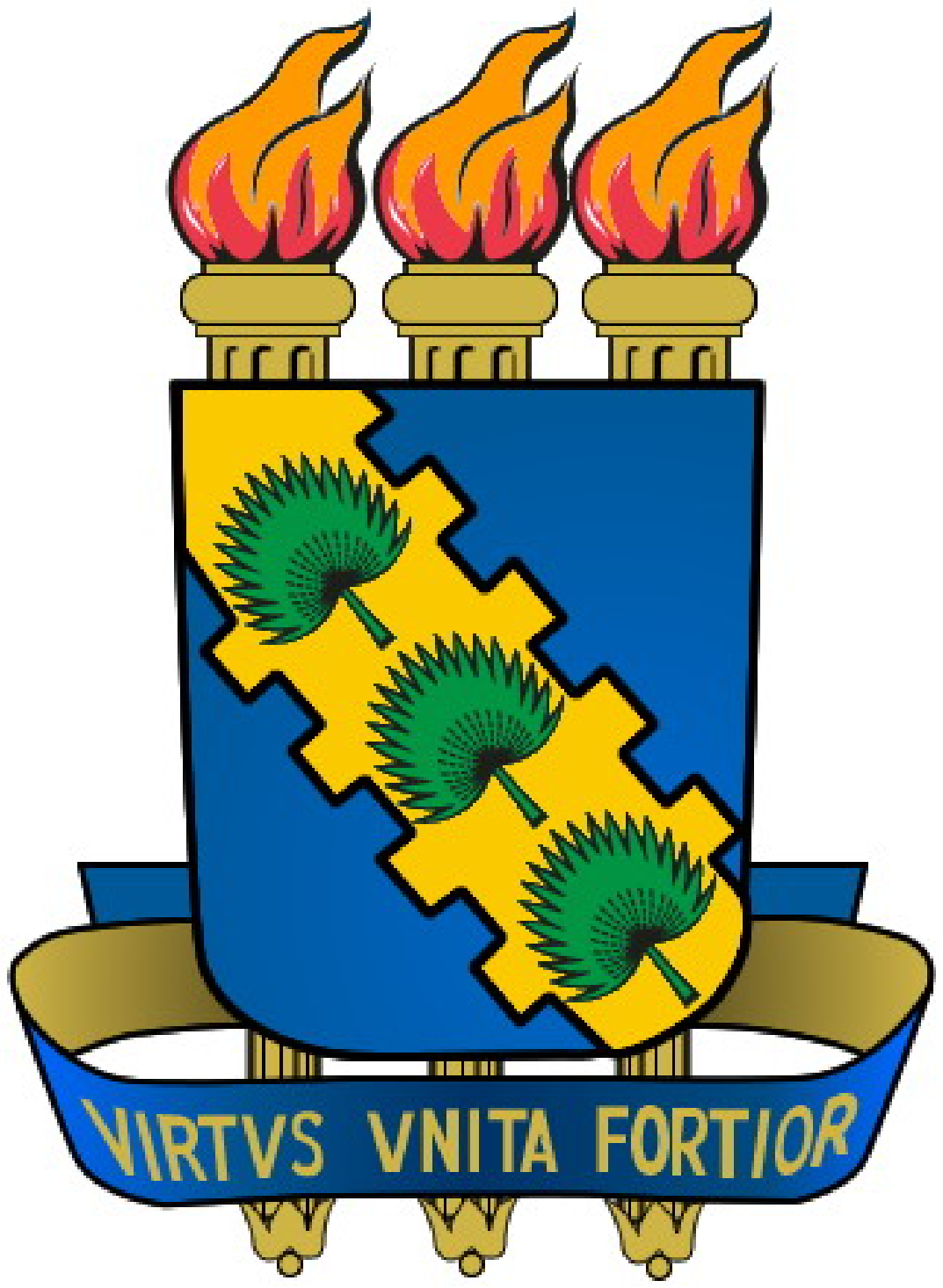} \\%
    \textsc{
    Universidade Federal do Ceará \\%
    Campus de Sobral \\
    Programa de Pós-Graduação em Engenharia Elétrica e de Computação\\%
    }

    \null\vfill
    \vspace{.3cm}
    \textbf{\textsc{\large{Elvys Linhares Pontes}}} \\%
    \null\vfill

    \vspace{.5cm}%
        \textbf{ \textsc{\large{UTILIZAÇÃO DE GRAFOS E MATRIZ DE SIMILARIDADE NA SUMARIZAÇÃO AUTOMÁTICA DE DOCUMENTOS BASEADA EM EXTRAÇÃO DE FRASES\\}}}

    \null\vfill
    \vspace{3cm}%

    \null\vfill
    \vspace{.25cm}%

    {\normalsize    \textsc{Sobral \\%
                            2015}
    }
\end{center}

\pagecolor{white}
\begin{titlepage}
    \vfill
    \begin{center}
        \textbf{\textsc{\large{Elvys Linhares Pontes}}}\\[5cm]
        \textbf{ \textsc{\large{UTILIZAÇÃO DE GRAFOS E MATRIZ DE SIMILARIDADE NA SUMARIZAÇÃO AUTOMÁTICA DE DOCUMENTOS BASEADA EM EXTRAÇÃO DE FRASES}}}\\[1cm]
        \hspace{0.45\textwidth}
            \begin{minipage}{.5\textwidth}
                \begin{espacosimples}
                    Dissertação submetida à coordenação do programa de Pós-Graduação em Engenharia Elétrica e de Computação da Universidade 
Federal do Ceará, como requisito parcial para obtenção do grau de mestre em Engenheiro Elétrico e de Computação.\\ \\
                    Orientadora: Profa. Dra. Andréa Carneiro Linhares \\
                    Co-orientador: Prof. Dr. Juan-Manuel Torres-Moreno
                \end{espacosimples}
            \end{minipage}
        \vfill
        \textsc{Sobral \\%
        2015}
    \end{center}
\end{titlepage}




\chapter*{Agradecimentos}
\label{ch_agradecimentos}
\thispagestyle{empty}

\dropping{2}{A}gradeço primeiramente a Deus. A minha orientadora Andréa e meu co-orientador Juan-Manuel pela ajuda e paciência no trabalho. 
À FUNCAP pelo fomento e à Universidade Federal do Ceará. A todos os meus professores, amigos e a minha querida família. Em especial, eu 
quero agradecer a minha namorada Polyanna por sempre me apoiar e incentivar a enfrentar os obstáculos e aventuras da vida e superá-los.
\vspace{1.4cm}

\begin{flushright}
  \textit{Elvys Linhares.}
\end{flushright}

\label{ch_resumo}

\begin{resumo}
\dropping{2}{A} internet possibilitou o aumento da quantidade de informação disponível. Entretanto, as práticas de ler e compreender essas informações são tarefas  dispendiosas. Nesse cenário, as aplicações de Processamento de Linguagem Natural (PLN) possibilitam soluções muito importantes, destacando-se a Sumarização Automática de Textos (SAT), que  produz um resumo a partir de um ou mais textos-fontes. 
Resumir um ou mais textos de forma automática, contudo, é uma tarefa complexa devido às dificuldades inerentes à análise e geração desse resumo. Esta dissertação descreve as principais técnicas e metodologias (PLN e heurísticas) para a geração de sumários. São igualmente abordados e propostos alguns métodos heurísticos baseados em Grafos e em Matriz de Similaridade para mensurar a relevância das sentenças e gerar resumos por extração de sentenças. Foram utilizados os corpus multi-idioma (Espanhol, Francês e Inglês), CSTNews (Português do 
Brasil), RPM (Francês) e DECODA (Francês) para avaliar os sistemas desenvolvidos e os resultados assim obtidos foram bastante interessantes.

\noindent\textbf{Palavras-chave:} Processamento da Linguagem Natural, Sumarização Automática de Textos, Grafos, Matriz de Similaridade.

\end{resumo}

\begin{abstract}
\label{ch_abstract}

\dropping{2}{T}he internet increased the amount of information available. However, the reading and understanding of this information are costly tasks. In this scenario, the Natural Language Processing (NLP) applications enable very important solutions, highlighting the Automatic Text Summarization (ATS), which produce a summary from one or more source texts. Automatically summarizing one or more texts, however, is a complex task because of the difficulties inherent to the analysis and generation of this summary. This master's thesis describes the main techniques and methodologies (NLP and heuristics) to generate summaries. We have also addressed and proposed some heuristics based on graphs and similarity matrix to measure the relevance of judgments and to generate summaries by extracting sentences. We used the multiple languages (English, French and Spanish), CSTNews (Brazilian Portuguese), RPM (French) and DECODA (French) corpus to evaluate the developped systems. The results obtained were quite interesting.

\noindent \textbf{Keywords:} Natural Language Processing, Automatic Text Summarization, Graph, Similarity Matrix.

\end{abstract}

\listof{figure}{Lista de Figuras}

\listof{grafico}{Lista de Gráficos}

\listof{tabela}{Lista de Tabelas}

\chapter*{Lista de Siglas e Acr\^onimos}

\begin{acronym}[XXXXXXXX]\addtolength{\itemsep}{-0.5\baselineskip}
\acro{Artex}{\textit{Autre Resumeur TEXtuel}}
\acro{CAS}{\textit{Chemical Abstracts Service}}
\acro{CST}{\textit{Cross-document Structure Theory}}
\acro{DVS}{Decomposição de Valores Singulares}
\acro{FMN}{Fatorização de Matrizes Não-negativas}
\acro{FRESA}{\textit{FRamework for Evaluating Summaries Automatically}}
\acro{JS}{Jensen-Shannon}
\acro{ISF}{\textit{Inverse Sentence Frequency}}
\acro{KL}{Kullback-Leibler}
\acro{LSA}{\textit{Latent Semantic Analysis}}
\acro{MEV}{Modelo Espaço Vetorial}
\acro{MRS}{\textit{Multi-document Rhetorical Structure}}
\acro{PCM}{Problema do Caminho Mínimo}
\acro{PLI}{Programação Linear Inteira}
\acro{PLN}{Processamento da Linguagem Natural}
\acro{RAG}{\textit{Résumeur Avec de Graphes}}
\acro{ROUGE}{\textit{Recall-Oriented Understudy for Gisting Evaluation}}
\acro{SASI}{Sumarizador Automático baseado em Subconjunto Independente}
\acro{SAT}{Sumarização Automática de Textos}
\acro{SIM}{Subconjunto Independente Máximo}
\acro{SIV}{Subconjunto Independente de Vértices}
\acro{SUMMatrix}{\textit{SUMmarizer based on Matrix model}}
\acro{SVR}{\textit{Support Vector Regression}}
\acro{TF}{\textit{Term Frequency}}
\acro{TF-IDF}{\textit{Term Frequency - Inverse Document Frequency}}
\acro{TF-ISF}{\textit{Term Frequency - Inverse Sentence Frequency}}
\acro{VSM}{\textit{Vector Space Model}}
\end{acronym}

\newpage
\chapter*{Símbolos e Notações}
\dropping{2}{N}esta seção são apresentados os símbolos e as notações utilizadas nesta dissertação. De forma geral, os escalares e os 
vetores serão representados por letras em itálico e as matrizes por letras romanas maiúsculas em negrito. Outras convenções são listadas 
a seguir:

\setlongtables
\begin{longtable}{lX}

\multicolumn{2}{l}{\textbf{Conjuntos e Espaços Vetoriais}}\\

$\mathbb{R}$            & Conjunto dos números reais\\%



\multicolumn{2}{l}{\textbf{Vetores e Matrizes}}\\

$\mathbf{S}$            & Matriz de saco de palavras\\

$\mathbf{M}$            & Matriz de palavras\\


\multicolumn{2}{l}{\textbf{Funções e outros operadores}}\\


$ \log_{b}{(\cdot)} $           &  Função logarítmica de base $b$\\




\end{longtable}
\addtocounter{table}{-1}

\tableofcontents

\chapter{Introdução}
\label{ch:int}

O avanço tecnológico possibilitou melhorar e aumentar a velocidade de comunicação mundial através da transmissão de vídeos, imagens e 
sons. Atualmente, grande parte dos livros possui uma versão digital ou áudio e a popularização das redes sociais (como \textit{Facebook}, 
\textit{Twitter}, \textit{Youtube}, entre outros) e sítios de notícias possibilitaram um grande aumento da quantidade de informações 
trafegadas pela internet acerca dos mais diversos assuntos. Todos os dias, uma quantidade considerável de informações é adicionada em 
vários sítios com comentários, fotos, vídeos e áudio. Dessa forma, um acontecimento é rapidamente divulgado na Web por diferentes 
fontes de notícias e formatos (áudio, imagem, texto e vídeo). 

O leitor, além de não possuir tempo hábil para ler essa quantidade de informações, não se interessa por todos os assuntos propostos e, 
geralmente, seleciona o conteúdo do seu interesse. Vale destacar que grande parte da informação criada a todo momento é pessoal como, por 
exemplo, comentários da vida cotidiana, fotos e vídeos pessoais divulgados em redes sociais e blogs. Assim, parte dessas informações não é 
importante para o público em geral. Por isso, os jornais, filmes, livros, revistas, \textit{sites} e blogs possuem manchetes, sumários ou sinopses 
dos assuntos abordados. O leitor, ao ler as manchetes de um jornal, identifica o assunto de cada notícia e, então, pode escolher qual delas 
ler na íntegra. O processo é similar para livros e filmes com suas sinopses e os destaques nos sítios e blogs. Dessa forma, o leitor 
identifica rapidamente o assunto de seu interesse para em seguida dar sequência à leitura. 

As manchetes e sinopses de livros e filmes (entre outros) são tipos de resumos. De forma geral, o ``resumo'' é composto pela ideia principal 
apresentada no documento original de forma curta e objetiva. Para uma melhor compreensão dessa palavra seguem algumas definições existentes 
na literatura e em dicionários:

\begin{itemize}
 \item Condensação em poucas palavras do que foi dito ou escrito mais extensamente\footnote{Dicionário Michaelis online: http://michaelis.uol.com.br};
 \item É a abreviação, representação precisa do conteúdo do documento original, preferencialmente feita pelo autor do documento;
 \item É um produzido de um ou mais textos, que contém as informações importantes do texto original, e não pode ser maior que a metade do texto original.
\end{itemize}

O resumo deve conter as principais informações do texto original de forma clara e concisa possibilitando a compreensão do mesmo. Ele pode 
ser criado utilizando diferentes estratégias. Entretanto, cada uma delas cria um resumo diferente com características específicas (boas ou 
ruins). O homem geralmente segue uma metodologia para criar um resumo. Inicialmente, ele lê o texto e 
seleciona as principais informações. Com esses dados e seu conhecimento, ele constrói novas frases de tamanho menor contendo as informações 
relevantes e a ideia geral contida no texto original.

Outra característica fundamental do resumo é o seu tamanho. O resumo pode ser produzido em diferentes tamanhos dependendo do fim almejado. 
Por exemplo, as manchetes de notícias de jornais e sítios possuem poucas palavras para chamar a atenção do leitor e transmitir a ideia chave 
da notícia. Entretanto, os textos mais longos necessitam um resumo mais extenso para o leitor compreender o assunto do 
texto, como é o caso de livros, que necessitam resumos maiores que uma notícia do cotidiano para ter sua ideia geral transmitida. Uma 
forma de mensurar o tamanho do resumo é analisar sua quantidade de palavras ou caracteres. Outra forma possível é a taxa de compressão 
$tc$, que é responsável por definir o tamanho do resumo em relação ao seu texto original. Ela é definida pela equação \ref{eq:txcp}.

\begin{equation}
 \label{eq:txcp}
 tc = \frac{|resumo|}{|texto|}
\end{equation}

O resumo é um processo de compressão que remove o conteúdo não relevante e mantém as informações fundamentais do texto. Quanto menor o
valor da $tc$, menor será o tamanho do resumo do texto analisado. Essa redução, até um certo nível, melhora a qualidade do resumo pois 
ressalta as principais informações do mesmo. Entretanto, a redução exagerada do texto ocasiona a perda de informações relevantes e de sua 
compreensibilidade. Vários trabalhos analisaram a melhor taxa de compressão do documento fonte para manter um resumo pequeno com 
ideias claras e concisas. Segundo \cite{Lin99}, a taxa de compressão variando entre $0,15$ e $0,30$ possibilita criar bons resumos sem 
perder as principais informações do texto.

Alguns resumos possuem estrutura e tamanho específicos, como é caso dos \textit{Twitter} e \textit{SMS} \cite{twitter}, que 
contêm um limite de caracteres pré-determinado e diferentes gírias e abreviações. Nesses casos, é necessário uma filtragem de expressões 
(\textit{hashtags}, \textit{smilefaces}, links, entre outros) e um novo vocabulário contendo-as para analisar as mensagens e criar 
os resumos utilizando-as.

O resumo facilita e acelera a obtenção de informações relevantes para o leitor. Entretanto, a grande quantidade de textos e o custo elevado 
de resumidores profissionais tornam impossível resumir muitos documentos em tempo hábil e com custo acessível. Uma solução para esse 
problema apresenta-se através da análise e da criação automática de resumos de textos.

\section{Processamento da Linguagem Natural}
\label{sc:pln}

O \ac{PLN} é uma área de pesquisa envolvendo Linguística, Inteligência Artificial e Ciências da Computação, com o propósito de fazer a 
interação entre a linguagem natural humana e das máquinas. É necessário um grande conhecimento dos idiomas e expressões corporais, orais e 
escritas para realizar a análise correta de um texto, imagem ou video. Os computadores ainda não conseguem realizar esse processo 
perfeitamente, pois a complexidade para representar e identificar completamente a estrutura sintática e semântica de um documento é elevada. 
O \ac{PLN} necessita da \cite{Dyer}: 

\begin{itemize}
 \item Representação de conhecimentos abstratos;
 \item Aprendizagem de módulos de processamento e troca de informações entre eles;
 \item Criação e propagação de ligações dinâmicas;
 \item Manipulação de estruturas;
 \item Aquisição e acesso a memórias lexicais, semânticas e episódicas.
\end{itemize}

O \ac{PLN} possui diversos campos de estudo: análise do discurso, separação morfológica, tradução automática, geração e compreensão da 
linguagem natural (mensagem escrita e oral), reconhecimento de caracteres, reconhecimento de discurso e análise de sentimentos, etc. As 
principais aplicações do \ac{PLN} são: \ac{SAT}, tradução automática, reconhecimento de discurso, compressão textual, corretor
automático e pergunta resposta. 

A \ac{SAT} consiste em resumir um ou mais textos de forma automática. Esses textos podem estar
presentes em um banco de dados, repositório local ou remoto. O sistema sumarizador analisa os textos identificando os dados relevantes e 
cria um novo texto baseado nessas informações. Para isso, é necessário reconhecer as palavras, a estrutura das sentenças e analisar a 
estrutura sintática das frases, verificando as palavras e seus significados em cada sentença. Dessa forma, o sistema identifica os dados 
relevantes e cria um novo texto através de paráfrases (ou extração de frases) com o conteúdo dessas informações.

A tradução automática consiste no processo de traduzir um texto de qualquer idioma para outro. Essa tarefa é bem complexa, pois é 
necessário saber a semântica e a gramática de cada idioma para compreender o texto com suas variâncias linguísticas regionais e traduzir as 
sentenças para outro idioma. Dessa forma, é possível traduzir um texto mantendo a sua coesão e coerência.

A extração de dados obtém as informações específicas de um ou mais textos. Elas podem ser obtidas de forma simples ou através de 
sinônimos, da reordenação de expressões ou da análise de expressões similares para identificar os dados desejados.

O reconhecimento do discurso realiza o reconhecimento textual de uma mensagem por meio de uma conversa entre pessoas ou de um áudio.
Já a compressão textual analisa um texto e reduz o seu conteúdo mantendo sua informatividade e gramaticalidade. Uma problemática dessa 
tarefa consiste em como definir mecanismos para avaliar as informações mais importantes de um documento e reduzir o texto mantendo-o 
claro e conciso.

O corretor automático analisa o texto e verifica a gramática, concordância verbal e nominal das sentenças identificando erros e sugerindo 
possíveis correções. O sistema pergunta resposta analisa automaticamente os textos com o intuito de encontrar informações sobre um 
determinado tema de uma pergunta e retorna uma resposta (com base nos textos analisados).

Dentre as aplicações de \ac{PLN} citadas anteriormente, a \ac{SAT} foi escolhida como problemática base desta dissertação. Atualmente, 
existem diversos sistemas sumarizadores automáticos de textos que analisam diferentes tipos de documentos e estruturas, e criam resumos
em formatos distintos. \cite{Archana} classificam os sistemas sumarizadores como:

\begin{itemize}
 \item Os resumos são criados pelos processos de abstração ou extração de sentenças. A abstração gera um resumo a partir da criação de novas 
frases com base no texto original. Ela requer um conhecimento mais avançado da estrutura do idioma incluindo suas regras gramaticais e de 
construção de sentenças. Ela possibilita ainda uma melhor análise do texto e da mensagem. A extração geralmente utiliza menos recursos das 
linguagens e se concentra em criar resumos a partir das estruturas já existentes no texto. Usualmente, são utilizadas regras estatísticas 
para avaliar a relevância das sentenças e o resumo é criado a partir daquelas consideradas mais relevantes.
 \item Há sumários genéricos ou voltados a um tema específico. Os resumos genéricos não diferenciam o conteúdo abordado nos
 documentos e realizam o mesmo tipo de análise em todos os textos. Os direcionados a um tema possuem regras e conceitos mais avançados 
 para uma área de análise. Por exemplo, os resumos voltados para a área da saúde necessitam e consideram conceitos e regras mais 
 específicos dos termos encontrados nesses documentos.
 \item A criação de resumos pode ser mono ou multidocumento. A sumarização monodocumento analisa um texto e cria o resumo com as principais informações do mesmo. A multidocumento analisa geralmente um \textit{cluster} de documentos. Esse \textit{cluster} pode conter informações similares (ou não) de uma mesma época (ou não).
 \item Os resumos podem ser indicativos ou informativos. Os indicativos fornecem a informação geral do texto sem conter informações 
específicas. Os informativos fornecem a informação geral do texto juntamente com dados específicos do documento original.
 \item O sumário pode conter somente as informações-chave sem descrever dados básicos, como também pode conter dados básicos para
 auxiliar o leitor a compreender melhor o assunto descrito.
\end{itemize}

\section{Justificativa}
\label{sc:just}

Como mencionado anteriormente, a grande quantidade de informação existente e o tempo limitado para ler os textos por uma pessoa 
se tornou um problema na sociedade moderna. A \ac{SAT} é uma ferramenta bastante útil a fim de propor soluções para esse problema e agilizar 
o processo de identificação e leitura de textos. Algumas facilidades e/ou vantagens em utilizar a sumarização são \cite{Archana}:

\begin{itemize}
 \item O resumo economiza o tempo de leitura possibilitando ao leitor identificar rapidamente o assunto principal do texto.
 \item O resumo facilita a busca e a seleção de textos na literatura.
 \item Melhora a qualidade de indexação dos documentos.
 \item Os algoritmos de sumarização automática não têm preferências sobre assuntos ou coisas possibilitando a criação de um resumo sem uma 
possível análise tendenciosa da opinião de um resumidor profissional.
\end{itemize}

\section{Objetivos}
\label{sc:obj}

Os principais objetivos desta dissertação são pesquisar e desenvolver sistemas de \ac{SAT} por extração de sentenças nos idiomas Francês e Português. Inicialmente, serão analisadas diferentes metodologias para criar resumos como, por exemplo, matriz de similaridade, conjunto estável e algoritmos em grafos. Serão explorados e analisados diversos sistemas sumarizadores na literatura a fim de estudar seus funcionamentos e utilizá-los como referência para a avaliação qualitativa dos sistemas aqui propostos. 

Serão desenvolvidos quatro sistemas sumarizadores independentes de idioma: \ac{RAG}, LIA-RAG, \ac{SASI} e \ac{SUMMatrix} para sumarizar (multi)documentos em Francês e Português. O \ac{RAG}, LIA-RAG e \ac{SASI} utiliza a Teoria de Grafos e o \ac{SUMMatrix} integra o uso de matrizes de similaridade. Outra parte do trabalho analisará os sistemas responsáveis pela análise qualitativa automática de resumos. Serão estudados e utilizados os sistemas \ac{FRESA} e \ac{ROUGE} para avaliar a qualidade dos resumos através das métricas precisão, cobertura e $medida$-$f$.

\section{Estrutura}
\label{sc:estrutura}

O restante desta dissertação está estruturado da seguinte forma:
\begin{itemize}
 \item O capítulo 2 aborda o estado da arte envolvendo os trabalhos em \ac{SAT}. São descritas diferentes metodologias para analisar os 
textos e produzir seus respectivos resumos.
 \item O capítulo 3 descreve a modelagem matemática e as fórmulas utilizadas para calcular a relevância, similaridade, divergência e 
identificar as sentenças principais dos textos.
 \item O capítulo 4 descreve o funcionamento dos principais tipos de sistemas de \ac{SAT} na literatura.
 \item O capítulo 5 analisa a estrutura e a metodologia dos sistemas propostos nesta dissertação (LIA-RAG, RAG, SASI e SUMMatrix).
 \item O capítulo 6 detalha a estrutura e as características de cada corpus\footnote{O corpus é um conjunto estruturado de documentos com 
determinadas características.} utilizado na avaliação dos sistemas.
 \item O capítulo 7 descreve os resultados obtidos para cada corpus e a análise geral dos sistemas.
 \item O capítulo 8 conclui esta dissertação destacando os pontos positivos e negativos dos resultados obtidos e os possíveis trabalhos futuros.
\end{itemize}
\chapter{Estado da arte}
\label{ch:est}

Os primeiros trabalhos acerca da sumarização automática de documentos abordavam somente textos jornalísticos através de técnicas simples baseadas na frequência das palavras para avaliar a relevância das frases \cite{Luhn}. Os resumos de profissionais possuem ótima qualidade em termos de informatividade e legibilidade. Entretanto, sua produção é mais lenta, cara e sujeita à subjetividade do profissional. Em contrapartida, os resumos produzidos de modo automático têm custo de produção bem reduzido, inexistência de problemas de subjetividade e de 
variabilidade observados nas proposições dos resumidores profissionais, dentre outros. Verificou-se, assim, a necessidade de gerar resumos automáticos visando essas facilidades e a fim de lidar com o crescimento da quantidade de informações.

\cite{Edmundson} deu continuidade aos trabalhos de Luhn, adicionando ao processo de produção de resumos considerações sobre posição das frases e presença de palavras provenientes da estrutura do documento (por exemplo, títulos, sub-títulos, etc.). As pesquisas desenvolvidas em \ac{CAS} \cite{Pollock}, concernentes à produção de sumários a partir de artigos científicos de Química, permitiram validar a viabilidade das abordagens de extração automática de frases. Uma ``limpeza''\ das frases através de operações de eliminação foi introduzida pela primeira vez. No intuito de adequar os resumos aos padrões impostos pela \ac{CAS}, uma normalização do vocabulário era efetuada. A normalização incluiu a substituição de palavras/frases por suas abreviações e uma padronização das variantes ortográficas. Os estudos sobre \ac{SAT} podem ser divididos em dois grupos, extração e abstração de texto. No primeiro grupo, há a identificação das partes mais relevantes de um ou mais textos, através de técnicas de recuperação de informação estatística para mensurar a relevância das frases e gerar o resumo através da concatenação das mesmas. O trabalho de \cite{Zhi} descreve vários métodos estatísticos utilizados no processo de sumarização de textos. A abstração analisa o texto original de uma forma linguística profunda, interpreta o texto semanticamente em uma representação formal, encontra novos conceitos mais concisos para descrevê-lo e, em seguida, gera um novo texto mais curto com o mesmo sentido do texto original com as informações relevantes apresentadas de forma concisa (fusão, paráfrase, etc) \cite{summarist}.

Existem diversos métodos para analisar a relevância das sentenças e construir um resumo. \cite{Saranyamol} classificaram esses métodos em: 
\ac{TF-IDF}, \textit{Cluster}, Teoria dos Grafos, Aprendizado de Máquina, \ac{LSA}, Redes Neurais e Lógica \textit{Fuzzy}.  Normalmente, 
esses métodos baseiam-se em cálculos estatísticos para verificar a relevância das sentenças. Por isso eles não analisam corretamente a 
estrutura e a semântica do texto, ocasionando erros gramaticais e sintáticos e reduzindo a coesão e coerência do resumo. 

De forma geral, os trabalhos voltados à extração de frases utilizam a seguinte metodologia de produção de sumários:
\begin{itemize}
\item Pré-processamento do texto;
\item Identificação das frases em destaque no documento;
\item Construção do sumário por concatenação das frases extraídas.
\end{itemize}

O pré-processamento do texto ou do \textit{cluster} geralmente se inicia com o processo de segmentação do texto para identificar frases, 
palavras e pontuações. Em seguida, identifica-se qual é o idioma do texto e realiza-se o processo de filtragem, em que são removidos os 
\textit{stopwords} (palavras sem relevância para o texto e para o conteúdo), e o processo de \textit{stemming}, em que as palavras são 
reduzidas a seus radicais, evitando-se suas flexões (singular e plural, masculino e feminino, derivações de palavras e verbos).

A identificação das frases relevantes no texto é normalmente feita através de uma ponderação que faz uso de métodos estatísticos: análise da 
frequência das palavras nas sentenças e no texto em geral, determinação da energia textual das sentenças \cite{enertex}, modelagem das 
sentenças como vetores \cite{artex}, grupos de vértices num grafo \cite{SASI}, entre outros.

A parte final do processo de sumarização é a geração de textos, que pode ser realizada através de diversas técnicas. Os métodos de sumarização por extração geralmente utilizam a abordagem de concatenar as sentenças relevantes \cite{artex,SASI}. Esse método é simples e rápido, mas pode criar vários problemas de coesão e coerência devido à possibilidade das frases selecionadas não apresentarem conexão entre si e assim, incorrer na perda do fluxo e da compreensão do texto. 

A abordagem por abstração utiliza métodos mais complexos e elaborados para gerar sentenças (fusão de sentenças, geração de paráfrases, etc) \cite{Filippova,seno1}.

Analisamos os trabalhos relacionados às principais metodologias utilizadas na literatura para a sumarização de textos com o intuito de analisar diferentes metodologias e compará-las com os sistemas propostos nesta dissertação. As seções seguintes discorrem 
sobre: Fatoração de Matrizes (seção \ref{sc:fdm}), Grafos (seção \ref{sc:grf}), Métodos Heurísticos (seção \ref{sc:heur}), \ac{PLI} (seção 
\ref{sc:pli}) e Submodular (seção \ref{sc:subm}). A seção \ref{sc:abst} mostra alguns trabalhos de sumarização por abstração de sentenças 
voltados ao Português Brasileiro e Inglês.

\section{Fatoração de Matrizes}
\label{sc:fdm}

A \ac{DVS} é um dos métodos de fatorização de matrizes bem estudado na literatura. A \ac{DVS} é usada em álgebra linear para minimizar erros 
computacionais em operações com matrizes de grande porte \cite{David}. A decomposição fatoriza a matriz $\textbf{M}$ em três matrizes, 
$\textbf{I}$, $\textbf{A}$ e $\textbf{X}$, de tal forma que $\textbf{M = I }\times \textbf{A} \times \textbf{X}^T$. A matriz $\textbf{I}$ 
é uma matriz unitária de dimensão $m \times n$ cujas colunas são vetores ortonormais. A matriz $\textbf{A}$ é uma matriz diagonal $n \times 
n$, cujos elementos diagonais são valores singulares não negativos, ordenados de forma decrescente. Finalmente, a matriz transposta de 
$\textbf{X}$ ($\textbf{X}^T$) é uma matriz ortogonal de dimensão $n \times n$, cujas colunas são chamadas de vetores singulares à direita. 

A \ac{LSA} é um método não supervisionado de obtenção de espaço vetorial representando a semântica de um grande corpus de textos. Esse 
método representa uma coleção de documentos através da distribuição dos termos do documento em uma matriz $\textbf{M}$. A \ac{LSA} utiliza a \ac{DVS} 
para decompor a matriz $\textbf{M}$ e obter as matrizes $\textbf{I}$, $\textbf{A}$ e $\textbf{X}$.

Os primeiros (maiores) valores na diagonal da matriz $\textbf{A}$ são postos a zero, resultando em uma espécie de análise de componentes 
principais. Esse processo reduz eficazmente a dimensionalidade das palavras para cada vetor \cite{Landauer}. 

No contexto da aplicação em sumarização automática, o documento é modelado como uma matriz $\textbf{M}$ representando a ocorrência dos 
termos por sentença (i.e., uma matriz com $m$ linhas, que representam os termos únicos, e com $n$ colunas, que representam cada frase). A 
\ac{DVS} é utilizada para reduzir o número de linhas (i.e., os termos correlacionados são agrupados em conceitos, capturando fenômenos como 
a sinonímia entre termos), preservando a estrutura de similaridade entre as colunas (i.e., entre as frases). 

Um algoritmo simples pode ser usado para selecionar a(s) melhor(es) frase(s), com base na decomposição \ac{DVS} e no uso da matriz 
de valores singulares à direita $\textbf{X}^T$. Cada sentença $i$ é representada pelo vetor coluna $[x_{j1}, x_{j2}, \ldots, x_{jn}]^T$ da 
matriz $\textbf{X}^T$. Uma abordagem simples é selecionar o primeiro vetor singular direito da matriz $\textbf{X}^T$ e em 
seguida a(s) frase(s) que têm o maior valor índice no vetor. Após isso, se necessário, efetua-se o mesmo processo para o segundo vetor 
singular direito da matriz, até se chegar ao número desejado de frases a serem selecionadas para a construção do sumário \cite{Gong}. 

Na proposta original de \cite{lee}, a \ac{FMN} decompunha uma matriz $\textbf{M}$ em duas matrizes não-negativas $\textbf{Y}$ e 
$\textbf{Z}$, de forma que $\textbf{M} = \textbf{Y} \times \textbf{Z}$, onde $\textbf{Y}$ é uma matriz não-negativa de características 
semânticas (i.e., a matriz dos termos) e $\textbf{Z}$ é uma matriz não negativa de variáveis semânticas (i.e., a matriz das frases). Um dos 
algoritmos mais populares para encontrar decomposições \ac{FMN} é baseado numa regra de atualização multiplicativa que atualiza 
iterativamente as matrizes $\textbf{Y}$ e $\textbf{Z}$ até obter a convergência de uma função objetiva ou até o algoritmo exceder um 
determinado número de passos. 

\section{Grafos}
\label{sc:grf}

Os métodos relacionados à Teoria dos Grafos consideram um grafo não-direcionado $G = (V,E)$ composto por um conjunto de vértices $V$ e um 
conjunto de arestas $E$. Existem diversas formas de modelizar o texto utilizando grafos. Filippova \cite{Filippova} utilizou cada palavra 
como um nó no grafo e o fluxo das sentenças para determinar a existência de arcos entre os vértices. \cite{SASI} consideram cada vértice 
representando uma sentença no texto e as arestas representam a existência de similaridade entre dois vértices (sentenças). \cite{Baralis} 
modelizam o texto utilizando grafos para representar os termos dos textos.

Mihalcea desenvolveu algoritmos de classificação baseados em grafos, tais como \textit{PageRank} \cite{Mihalcea}. Esses algoritmos foram 
utilizados com sucesso nas redes sociais, na análise do número de citações ou no estudo da estrutura da Web. Eles permitem tomar decisões 
acerca da importância de um vértice, baseando-se na informação global advinda da análise recursiva do grafo completo e não na análise local 
do vértice. No âmbito da sumarização automática, observa-se que o documento é representado por um grafo de unidades textuais (frases) 
conectadas entre si através de relações oriundas de cálculos de similaridade. As frases são em seguida selecionadas segundo critérios de 
centralidade ou de prestígio no grafo, e agrupadas a fim de produzir os extratos do texto \cite{Ferreira}.

Baralis \textit{et al.} também utilizaram a modelagem de grafos para sumarizar textos \cite{Baralis}. Entretanto, sua metodologia é composta 
por: processamento do texto, correlação do grafo, indexação do grafo e seleção de sentenças. O processamento realiza a 
\textit{stemming} e a remoção dos \textit{stopwords}. A correlação do grafo é feita a partir dos conjuntos de itens frequentes no texto. A 
indexação do grafo dá-se através de um algoritmo baseado no \textit{PageRank} para ponderar os nós do grafo. A geração do resumo seleciona 
as sentenças com melhor ponderação baseada na indexação dos nós.

\begin{figure}[!htb]
  \begin{floatrow}
     \centering
      \includegraphics[width=12cm]{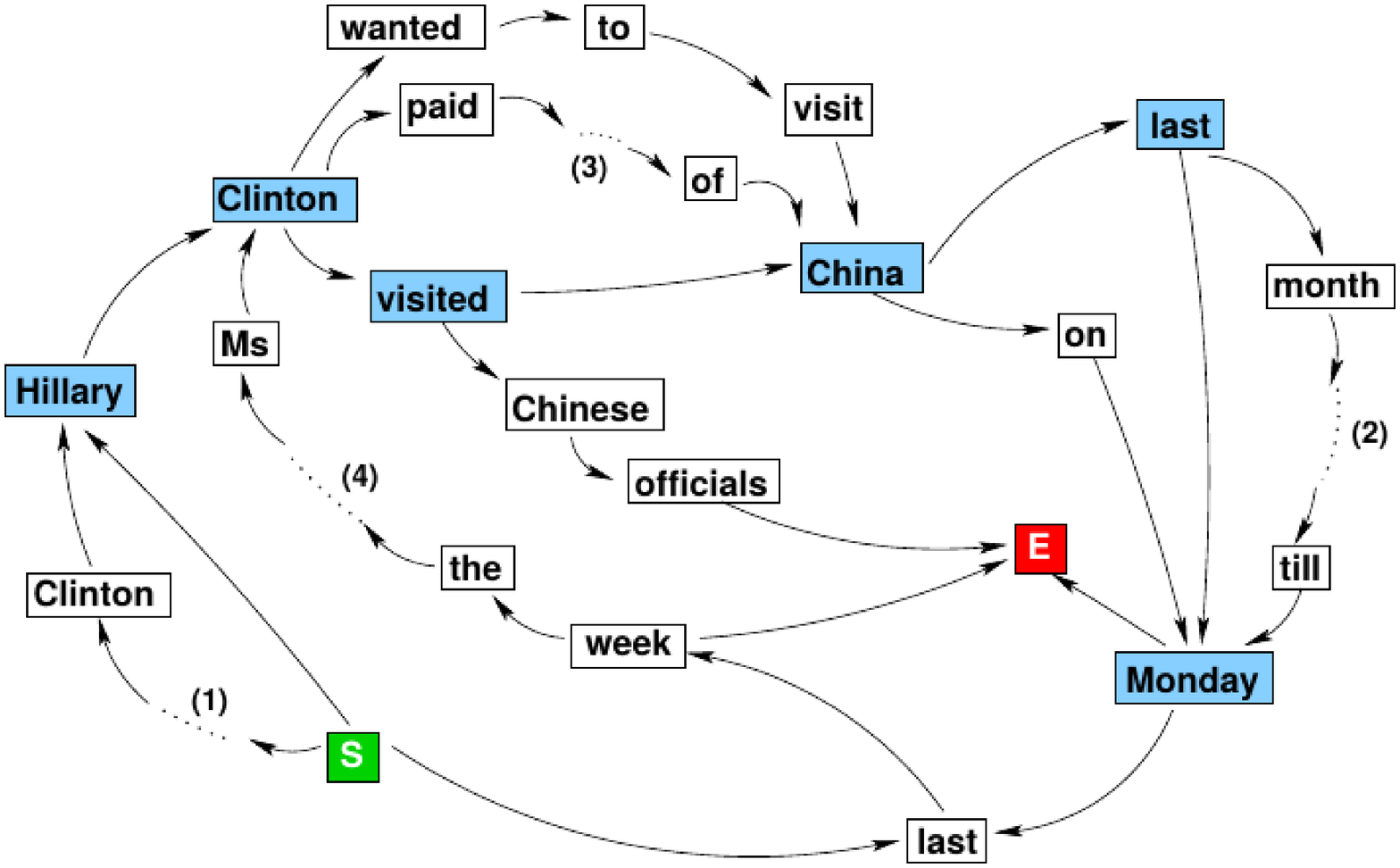}
     \caption{\label{fg:compressao}Exemplo de compressão de sentenças utilizando grafos \cite{Filippova}.}
  \end{floatrow}
\end{figure}

Filippova descreve outro método através da fusão multissentença, apoiado em uma estrutura de grafo para representar palavras e frases 
\cite{Filippova}. Os arcos do grafo são ponderados com a frequência de palavras e a posição nas frases. A fusão é realizada através do 
cálculo do menor caminho no grafo de palavras. A figura \ref{fg:compressao} exemplifica a fusão utilizando várias sentenças similares para 
criar uma nova sentença com o conteúdo que as representa.

O trabalho de Pontes \textit{et al} usa a Teoria dos Grafos concomitante com a divergência \ac{JS} para criar resumos multi-documentos por 
extração \cite{SASI}. O sistema representa o texto através de um grafo onde as frases correspondem aos vértices e as arestas representam a 
similaridade entre duas sentenças (vértices). Portanto, o grupo de vértices interligados no grafo representa sentenças com conteúdo similar. 
O sistema calcula o conjunto estável máximo do grafo visando criar o resumo com frases contendo informações gerais do \textit{cluster} e 
sem redundância. Por isso, o sistema seleciona geralmente uma frase de cada grupo, reduzindo a quantidade de sentenças muito similares.

\section{Métodos Heurísticos}
\label{sc:heur}

Torres desenvolveu vários métodos heurísticos para sumarização por extração. Em \cite{artex}, o texto é representado como um \ac{VSM} e cada 
sentença possui um vetor que descreve o tema do texto. A partir desse vetor, calcula-se a relevância de cada frase para construir o resumo. 
O trabalho de \cite{cortex} utiliza diferentes cálculos numéricos para determinar a relevância das sentenças. \cite{enertex} calculam a 
energia textual das sentenças que representa a relevância de cada uma delas com relação ao texto. Esses métodos serão melhores descritos no 
capítulo \ref{ch:lit}.

\section{Programação Linear Inteira}
\label{sc:pli}

Uma forma de melhorar a qualidade do resumo é maximizar a seleção das sentenças mais relevantes. A \acf{PLI} modela a sumarização de textos 
de forma a maximizar a qualidade da extração de sentenças por meio de análises do texto.

A \ac{PLI} consiste em maximizar ou minimizar uma função sujeita a um conjunto de restrições\footnote{A seção \ref{sc:spli} descreverá 
melhor a modelagem da \ac{PLI} para a sumarização de textos}. Mcdonald \cite{mcdonald2006} considerou a informatividade e a redundância das 
sentenças como pontos fundamentais para a \ac{SAT}. Ele avalia a qualidade do resumo a partir da relevância das frases com a inserção de uma 
penalidade para as sentenças redundantes do resumo.

\cite{benoit} basearam-se no modelo de Mcdonald para abordar o modelo de escalas de forma. Os autores tratam a redundância das sentenças 
sem necessitar de um número quadrático de variáveis, facilitando a modelagem e resolução do problema. Eles também modelam esse problema na 
formulação PLI realizando a compressão e a seleção de sentenças simultaneamente.

Já \cite{Dimitrios2012} adicionaram o modelo \ac{SVR} para mensurar a relevância das sentenças a partir do treinamento dos sistemas com 
resumos produzidos por humanos. Esse modelo avalia a relevância das sentenças baseando-se em sua diversidade.

\section{Submodular}
\label{sc:subm}

Seja $B$ um conjunto finito e seja $f$ uma função de partes de $B$ em $\mathbb{R}$. A função $f$ é submodular se, para quaisquer partes $C$ e 
$D$ de $B$, observa-se a relação exposta na equação \ref{eq:subrel}.

\begin{equation}
\label{eq:subrel}
 f(C \cup D) + f(C \cap D) \leq f(C) + f(D)
\end{equation}

Seja um conjunto de objetos $B$ e uma função $f$ que retorna um valor real para qualquer subconjunto $C \subseteq B$. O 
objetivo é escolher um subconjunto com tamanho limitado a um certo valor que maximize a função \cite{krause}. A \ac{SAT} por extração 
seleciona um subconjunto (conjunto de sentenças) $C \subseteq B$ para representar o documento de tal forma que o resumo tenha o tamanho 
máximo de $L$ palavras.

\cite{Lin2011} projetaram uma classe de funções submodulares voltadas à compactação de sentenças objetivando a diversidade e a 
representatividade do corpus. As funções são não-decrescentes, monótonas e submodulares, possibilitando um desempenho ideal de fator 
constante. Outra possibilidade é a aprendizagem supervisionada de funções submodulares para a extração de sentenças. \cite{sub2} aplicam 
esse método de aprendizagem a diversos métodos de compactação submodulares e demonstram sua eficácia com base na análise de vários conjuntos 
de dados.

\section{Sumarização por abstração}
\label{sc:abst}

A sumarização por abstração geralmente usa métodos mais complexos de análise, pois eles permitem verificar o conteúdo do documento e 
possibilitam que o sistema sumarizador possa ``saber'' quais informações estão presentes no mesmo. Possui uma abordagem baseada na estrutura 
e na semântica. A abordagem estruturada codifica a informação mais importante do documento por meio de esquemas cognitivos tais como 
modelos, regras de extração, e estruturais, tais como conhecimento base e estrutura da frase. O método semântico, por sua 
vez, concentra-se na identificação de frases nominais e verbais por meio de dados linguísticos para que o sistema de geração da linguagem 
natural possa criar o resumo do texto analisado. 

A sumarização por abstração geralmente usa a fusão de frases ou geração de paráfrases para revisar as informações e evitar redundâncias em 
um texto \cite{seno1}.

\begin{figure}[!htb]
  \begin{floatrow}
     \centering
      \includegraphics[width=11cm]{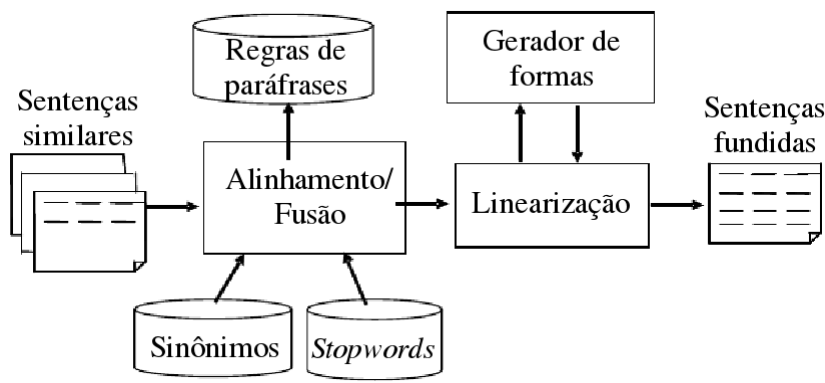}
     \caption{\label{fig:seno}Estrutura para fusão de sentenças similares.}
  \end{floatrow}
\end{figure}

A fusão por união preserva a mensagem geral do \textit{cluster}, enquanto a fusão por interseção analisa as informações redundantes 
presentes no mesmo. Seno propôs um método de fusão de sentenças similares em Português utilizando uma abordagem simbólica e independente de 
domínio (figura \ref{fig:seno}) \cite{seno1}. Esse método permite mesclar as sumarizações por união e interseção de um \textit{cluster} de 
documentos.  Já no trabalho de \cite{seno2} é descrita uma metodologia para identificar informações comuns entre frases em Português usando 
o conhecimento lexical, sintático e regras semânticas de paráfrase.

\begin{figure}[!htb]
  \begin{floatrow}
     \centering
      \includegraphics[width=12cm]{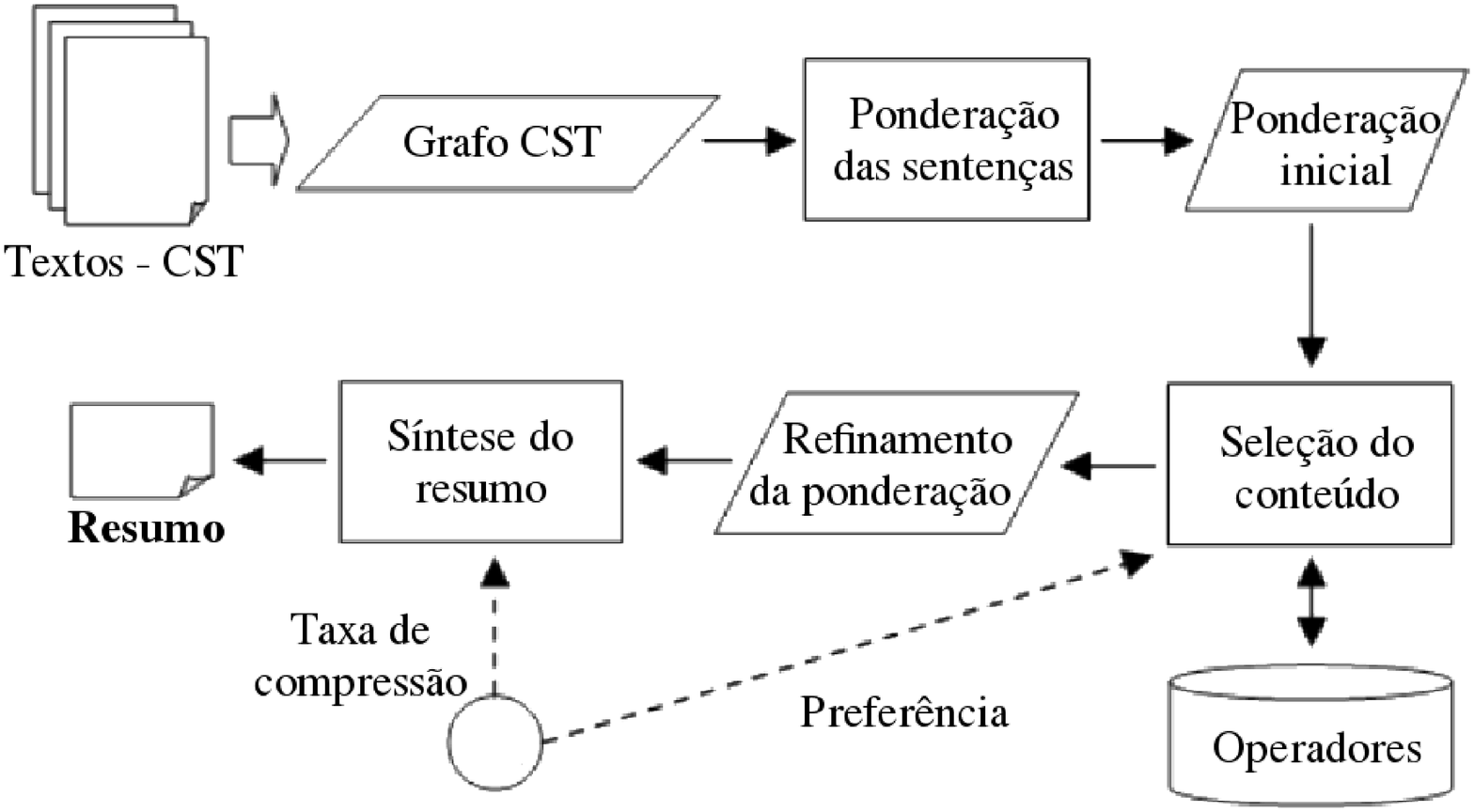}
     \caption{\label{fg:cstsumm}Arquitetura do sistema CSTSumm \cite{cstsumm}.}
  \end{floatrow}
\end{figure}

Em \cite{cstsumm} é desenvolvido um método de compactação multi-documento baseado no modelo \ac{CST}. Esse método analisa a redundância,
a complementaridade e a contradição entre as diferentes fontes de informação para calcular a relevância das sentenças. Em seguida, os 
operadores são utilizados no auxílio do processo de reponderação das sentenças a fim de criar um resumo com algumas características 
escolhidas pelo usuário. Por fim, o resumo é gerado a partir das sentenças de maior ponderação (figura \ref{fg:cstsumm}).

\begin{figure}[!htb]
  \begin{floatrow}
     \centering
      \includegraphics[width=12cm]{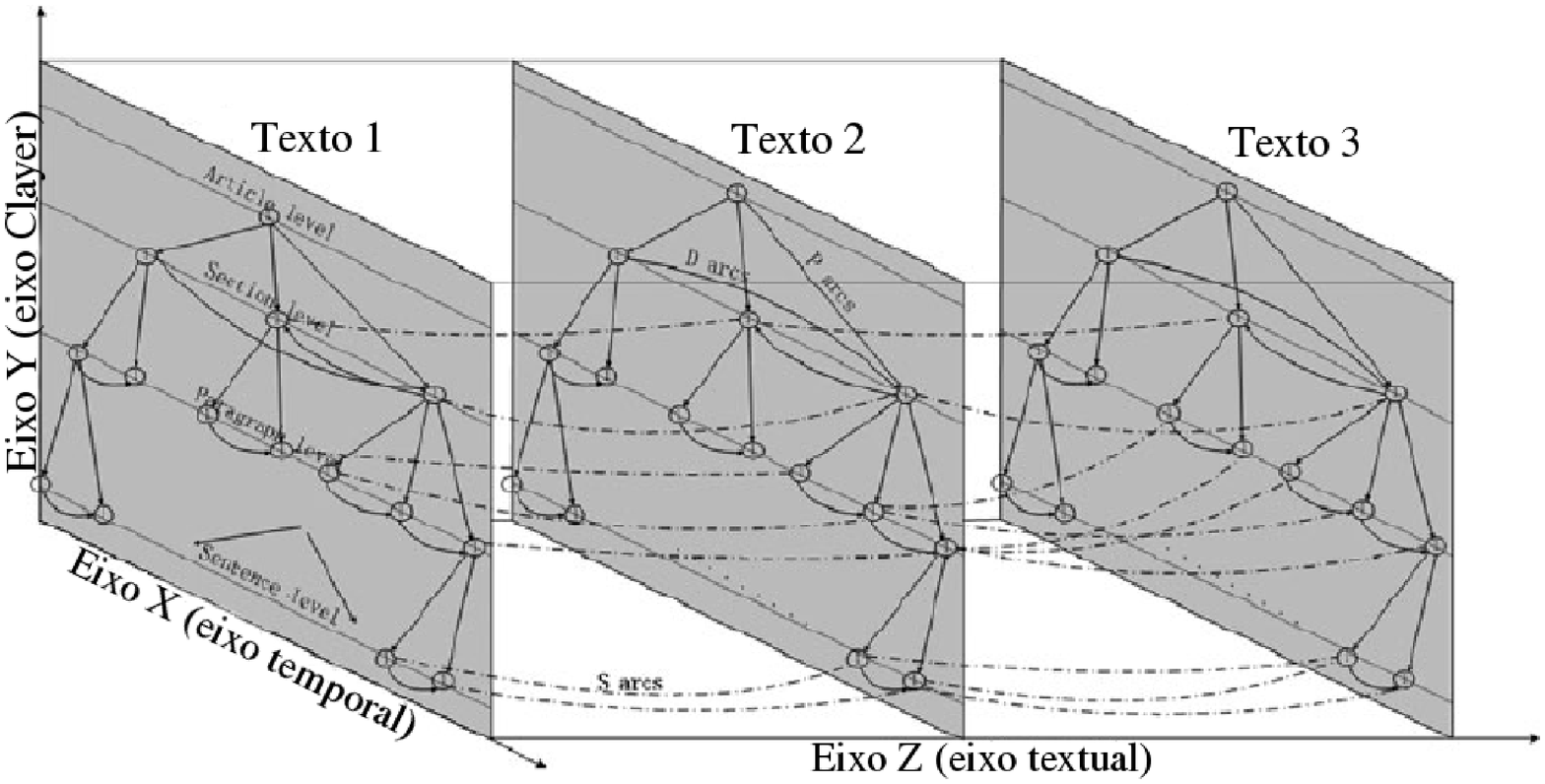}
     \caption{\label{fig:mrs}Estrutura \ac{MRS} \cite{Xu}.}
  \end{floatrow}
\end{figure}

O trabalho de \cite{Xu} usa os conceitos de árvore tópica hierárquica, retórica e relação temporal para calcular as inter-relações entre as 
unidades do texto. Ele utiliza ainda a estrutura \acf{MRS} (figura \ref{fig:mrs}) para representar o texto em diferentes níveis de 
granularidade (incluindo frases, parágrafos, seções e documentos). Seu algoritmo de extração realiza etapas de ponderação e remoção de nós a 
fim de selecionar as sentenças mais importantes. Primeiramente, ele executa o algoritmo de ponderação dos nós para, em seguida, utilizar um 
algoritmo de \textit{clustering} no intuito de identificar sentenças similares e remover as que são redundantes.

Barzilay \textit{et al.} desenvolveram um método de geração automática de resumos através da identificação e sintetização de elementos 
semelhantes em um conjunto de documentos \cite{Barzilay2}. Esse método seleciona o conteúdo dos textos a partir da similaridade das frases 
relacionadas com o tema discutido a fim de criar um novo texto. No trabalho de \cite{Barzilay} é descrita uma abordagem para fusionar frases 
por meio da técnica de geração \textit{text-to-text}\footnote{A geração \textit{text-to-text} é reconfiguração de um texto através da: 
reformulação, reorganização, combinação ou divisão das sentenças e eliminação do conteúdo.} de modo a sintetizar informações 
repetidas de vários documentos. Esse método utiliza um alinhamento sintático nas frases para identificar informações comuns às mesmas. Após 
a etapa de identificação, as sentenças são processadas e um novo texto é gerado mantendo a mesma ideia original.

Uma maneira de calcular a similaridade entre frases é através da co-ocorrência de palavras. Em \cite{He} é proposto um método de fusão 
usando métricas de similaridade, co-ocorrência \textit{Skip}-bigrama\footnote{\textit{Skip}-bigramas são quaisquer pares de palavras na ordem da frase.} e densidade de informação para calcular o valor das sentenças e selecionar os conteúdos mais relevantes. Hennig e Albayrak 
desenvolveram um modelo multi-documento para resumir um texto analisando a co-ocorrência de termos e bigramas\footnote{Bigramas são pares 
de palavras sequenciais em uma frase.} usando a divergência \ac{JS} \cite{Hennig}.

Hovy e Lin desenvolveram o sistema sumarizador SUMMARIST, que combina conhecimentos de conceitos do mundo (dicionários, ontologias e 
recursos similares) com PLN \cite{summarist}. O SUMMARIST é baseado na identificação da temática do documento e em sua interpretação. 
Através de técnicas de recuperação de informação e de dicionários, o sistema identifica os principais tópicos abordados no texto e faz a 
interpretação dos dados através de técnicas estatísticas, psicologia cognitiva, léxicos e dicionários. Por fim, o sumário é criado através 
de métodos de geração de frases.
\chapter{Modelagem do Problema}
\label{ch:model}

Este capítulo aborda a modelagem matemática utilizada nesta dissertação. A seção \ref{sc:saco} descreve o modelo de saco de palavras para 
identificar as palavras e suas distribuições nos documentos analisados. A seção \ref{sc:rel} descreve formas de avaliação da relevância 
inicial de palavras e sentenças. A seção \ref{sc:sim} e \ref{sc:div} discorrem sobre formas de calcular a similaridade e a divergência entre 
sentenças, respectivamente. Por fim, a seção \ref{sc:gr} descreve os conceitos gerais e os problemas (abordados neste trabalho) envolvendo a 
Teoria de Grafos.

\section{Saco de palavras}
\label{sc:saco}

Os métodos propostos nesta dissertação são baseados em um pré-tratamento específico de palavras e um modelo de saco de palavras para 
representar o texto. A modelagem de saco de palavras organiza e representa o modo como as palavras estão distribuídas nas sentenças. 
Utiliza-se uma matriz representada por $\textbf{S}_{[N_S \times N_P]}$ construída a partir do texto, onde $N_S$ é o número de sentenças e 
$N_P$ é o número de palavras distintas no documento. A célula $S_{ij}$ representa a frequência da palavra $j$ na frase $i$ ($FP_{ij}$).

\begin{equation}
\begin{split}
 \textbf{S}=\left(
\begin{array}{cccc}
S_{11} & S_{12} & \ldots & S_{1N_P}\\
S_{21} & S_{22} & \ldots & S_{2N_P}\\
\vdots & \vdots &  & \vdots \\
S_{N_S1} & S_{N_S2} & \ldots & S_{N_SN_P}\\
\end{array}
\right), 
S_{ij}=\left\{
\begin{array}{cc}
FP_{ij}, & \textrm{se }\exists\textrm{ a palavra j na frase i}\\
0, & \textrm{caso contrário}\\
\end{array}
\right.
\end{split}
\end{equation}

\section{Relevância das palavras e frases}
\label{sc:rel}

Uma forma de verificar a relevância de uma palavra ou de uma frase em relação ao texto é através do \ac{TF-ISF} \cite{ets}, que é uma variação do \ac{TF-IDF}\footnote{\acf{TF-IDF}}. Normalmente, as palavras mais frequentes representam ideias relevantes no texto. Entretanto, uma boa parte das mesmas são artigos, preposições ou conjunções, e assim serão referenciadas como \textit{stopwords}. Os \textit{stopwords} não são 
relevantes para a compreensão do texto e sua remoção auxilia sua análise. As palavras menos frequentes também podem ser 
relevantes. Então, uma melhor 
forma de analisar as palavras é através da \ac{TF-ISF}. Essa métrica analisa a frequência das palavras e sua distribuição nas sentenças. 
A relevância de uma palavra $w$ no texto, $tfisf(w)$, é dada pela equação \ref{tfisf}.

\begin{equation}
 \label{tfisf}
  tfisf(w) = tf(w) \times \log \frac{N_S}{n_{w}},
\end{equation}

\noindent onde $tf(w)$ corresponde a frequência da palavra $w$, $N_S$ é a quantidade de sentenças e $n_w$ é a quantidade de sentenças que 
possuem a palavra $w$. Consequentemente, a relevância de uma sentença $s$, $rel(s)$, é determinada pela equação \ref{sent}.

\begin{equation}
 \label{sent}
  rel(s) = \sum_{w \in s} tfisf(w)
\end{equation}

\section{Similaridade}
\label{sc:sim}

A métrica de similaridade de sentenças verifica a semelhança entre as mensagens transmitidas por cada sentença. O processo de análise de 
similaridade verifica o significado semântico das sentenças e analisa a semelhança das mensagens transmitidas. Entretanto, o processo de 
análise das sentenças exige uma verificação complexa da estrutura sintática e semântica das palavras.

Uma alternativa para avaliar a similaridade entre duas sentenças é analisar as palavras em comum entre elas. A subseção \ref{ssc:cos} 
descreve o cálculo da similaridade baseado na determinação do cosseno e a subseção \ref{ssc:jac} baseia-se nas palavras em comum para 
avaliar o nível de similaridade.

\subsection{Similaridade do cosseno}
\label{ssc:cos}

O método da similaridade do cosseno baseia-se no cálculo do cosseno do ângulo entre dois vetores. Essa métrica avalia duas sentenças e 
calcula o ângulo entre dois vetores representados por elas. Sejam as sentenças $P$ e $Q$ com as palavras $p_1, p_2, \dots, p_n$ e $q_1, q_2, 
\dots, q_n$, respectivamente. O produto escalar $PQ$ é dado pela equação \ref{eq:mult}, em que $p_k$ é 1 quando a palavra $p_k$ existe na sentença $P$, e, caso contrário, $p_k$ é 0. O mesmo ocorre para $q_k$ em relação à sentença 
$Q$. 

\begin{equation}
 \label{eq:mult}
  P.Q = p_1 \times q_1 + p_2 \times q_2 + \dots + p_n \times q_n
\end{equation}

A norma da frase é definida pela equação \ref{eq:norma}.

\begin{equation}
 \label{eq:norma}
 ||P|| = \sqrt[2]{p^2_1 + p^2_2 + \dots + p^2_n}
\end{equation}

A similaridade do cosseno é calculada segundo a equação \ref{eq:cos}. Seu valor varia entre $[0,1]$. Quanto maior for o valor do cosseno
maior é a semelhança entre as sentenças.

\begin{equation}
 \label{eq:cos}
 cos(P,Q) = \frac{P . Q}{||P|| \times ||Q||}
\end{equation}

\subsection{Similaridade de Jaccard}
\label{ssc:jac}

O coeficiente de similaridade de Jaccard calcula a similaridade entre conjuntos de forma estatística. Essa métrica é baseada na quantidade 
de elementos em comum entre dois conjuntos. Sejam duas sentenças $P$ e $Q$ cujos elementos são as palavras de cada uma delas. O coeficiente 
de similaridade de Jaccard, $Jac(.)$, é dado pela equação \ref{eq:jac} e o mesmo varia entre $[0,1]$, sendo 1 quando duas sentenças são 
iguais.

\begin{equation}
\label{eq:jac}
 Jac(P,Q) = \frac{|P \cap Q|}{|P \cup Q|}
\end{equation}

Outra medida possível é a distância de Jaccard, $dis_J(.)$, que calcula a dissimilaridade entre dois conjuntos através do complemento de Jaccard (equação \ref{eq:cjac}).

\begin{equation}
\label{eq:cjac}
 dis_J(P,Q) = 1 - Jac(P,Q) = \frac{|P \cup Q| - |P \cap Q|}{|P \cup Q|}
\end{equation}

\section{Divergência}
\label{sc:div}

A métrica de divergência também necessita da análise semântica para avaliar corretamente o significado e a diferença entre duas sentenças. Por isso, foram analisados alguns métodos estatísticos (\cite{dkl,Juan-Manuel}) para avaliar a divergência entre conjuntos de probabilidades. As subseções \ref{ssc:dkl} e \ref{ssc:djs} descrevem um método assimétrico e outro simétrico, respectivamente, para mensurar a divergência entre duas sentenças.

\subsection{Divergência de Kullback-Leibler}
\label{ssc:dkl}

Sejam $P$, $Q$ e $W$ conjuntos de palavras, onde $P$ e $Q$ representam as palavras de duas sentenças de um documento e $W$ é formado por sua união. 
$Pr(P,w)$ representa a probabilidade da palavra $w$ na sentença $P$ em relação ao conjunto $W$. Respectivamente, $Pr(Q,w)$ será a probabilidade da palavra $w$ na frase $Q$ em relação ao conjunto $W$.

A divergência \ac{KL} determina a informação perdida ao aproximar um conjunto de probabilidades do conjunto $P$ em relação ao conjunto de probabilidades do conjunto $Q$ (equação \ref{DKL}) \cite{dkl}. Ela permite calcular a diferença entre dois conjuntos de palavras distintas. 

\begin{equation}
  \label{DKL}
  D_{KL}(P||Q) = \frac{1}{2}\sum_{w \in W}\left( Pr(P,w)\log{\frac{Pr(P,w)}{Pr(Q,w)}}\right)
\end{equation}

Essa divergência é assimétrica ($D_{KL}(P||Q) \neq D_{KL}(Q||P)$), assumindo valores diferentes dependendo da análise realizada. Seu valor varia entre $[0,\infty)$ e suas probabilidades são semelhantes quando seus valores são próximos a zero.

\subsection{Divergência de Jensen-Shannon}
\label{ssc:djs}

A divergência \acf{JS} é uma versão simétrica e suavizada da \ac{KL}, fornecendo uma maneira mais estável para mensurar a diferença entre dois conjuntos de probabilidades (equação \ref{DJS}) \cite{Juan-Manuel, Saggion}. A divergência \ac{JS} pode assumir a mesma variação de valores que a \ac{KL}.

\begin{equation}
\label{DJS}
\begin{split}
  D_{JS}(P||Q) = \frac{1}{2}\sum_{w \in W}\left( Pr(P,w)\log{\frac{2 \times Pr(P,w)}{Pr(P,w)+Pr(Q,w)}} 
  + Pr(Q,w)\log{\frac{2 \times Pr(Q,w)}{Pr(P,w)+Pr(Q,w)}} \right)
\end{split}
\end{equation}

No caso de haver somente uma palavra em uma das frases analisadas, utiliza-se uma suavização para evitar valores nulos \cite{smoothBook}. Se uma palavra $w$ não está presente na frase $Q$, então $Pr(Q,w)$ é calculada pela equação \ref{smooth}, onde $\beta = 1.5 * voc$, $voc$ é o número de palavras distintas em $W$, $\gamma$ controla a relevância da palavra ausente na frase e $N_P$ é a quantidade de palavras em $W$. Caso a palavra $w$ esteja presente somente na frase $Q$, o cálculo da probabilidade $Pr(P,w)$ é simétrico a equação \ref{smooth}.

\begin{equation}
 \label{smooth}
  Pr(Q,w) = \left( \frac{Pr(P,w) + \gamma}{N_P + \gamma \times \beta} \right)
\end{equation}

\section{Grafos}
\label{sc:gr}

Esta seção descreve uma breve introdução sobre a Teoria dos Grafos (subseção \ref{ssc:conc}) e os problemas da clique e do subconjunto independente de vértices (subseção \ref{ssc:psi}) abordados em alguns algoritmos utilizados nos sistemas desenvolvidos.

\subsection{Conceitos gerais}
\label{ssc:conc}

\begin{figure}[!htb]
\centering
\subfigure[]{
    \includegraphics[scale=0.8]{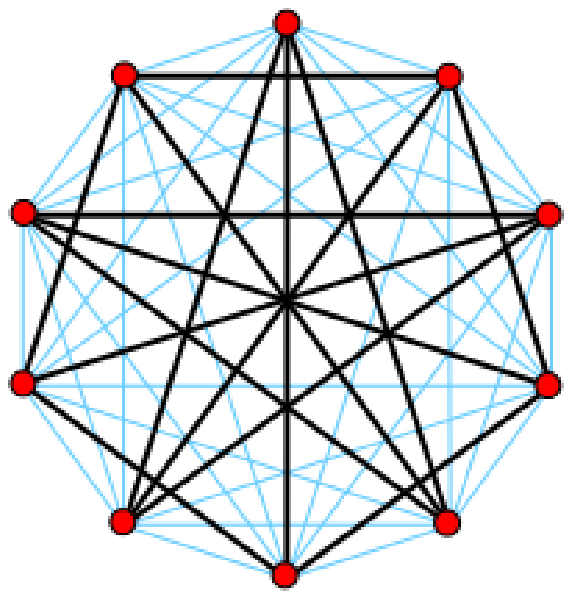}
    \label{fig:subfig1}
}
\ \ \ \ \ \ \ \ \ 
\subfigure[]{
    \includegraphics[scale=0.8]{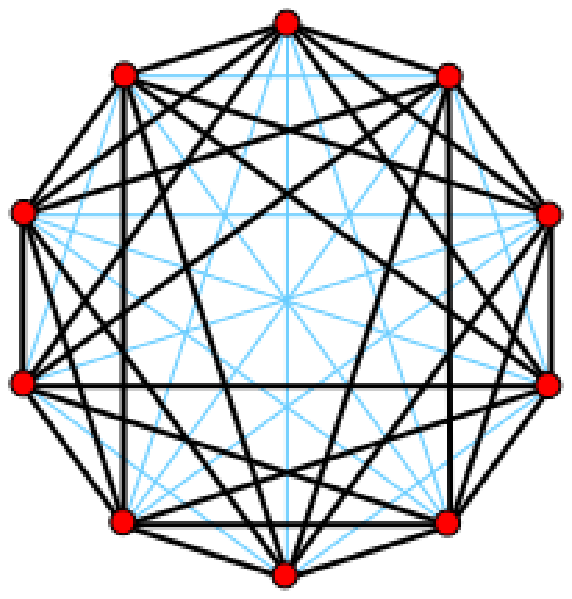}
    \label{fig:subfig1}
}
\caption{Exemplo de um grafo G (a) e seu complemento $\bar{G}$ (b).}
\label{fig:comp}
\end{figure}

Considere um grafo $G = (V, E)$, onde $V$ é o conjunto de vértices e $E$ o conjunto de arestas não orientadas de $G$ (ver figura \ref{fig:comp}.a). A cardinalidade ($n$) de um grafo é definida pela quantidade de vértices e o complemento do grafo $\bar{G}$ é o grafo $(V, \{u,v\} \in (V\times V)\backslash E: u \neq v)$ (figura \ref{fig:comp}.b\footnote{https://it.wikipedia.org/wiki/Grafo\_complemento}).

O grafo $G$ pode ser denso ou esparso. Ele é denso quando o número de arestas é próximo a $n(\frac{n-1}{2})$ (quando existe uma aresta entre todo par de vértices). Quando há poucas arestas, o grafo é denominado esparso. A vizinhança de um conjunto de vértices $U$ é o conjunto de todos os vértices que possui em algum vizinho em $U$. O grau de um vértice $v$ é definido pela quantidade de arestas que ele possui. 

Um subgrafo de um grafo $G$ é qualquer grafo $H$ tal que $V(H) \subseteq V(G)$ e $E(H) \subseteq E(G)$. O subgrafo de $G$ induzido por um subconjunto $U$ de $V(G)$ é o grafo $(U, J)$, em que $J$ é o conjunto de todas as arestas de $G$ que têm ambas as extremidades em $U$.

\begin{figure}[!htb]
\centering
\subfigure[Conexo]{
    \includegraphics[scale=0.19]{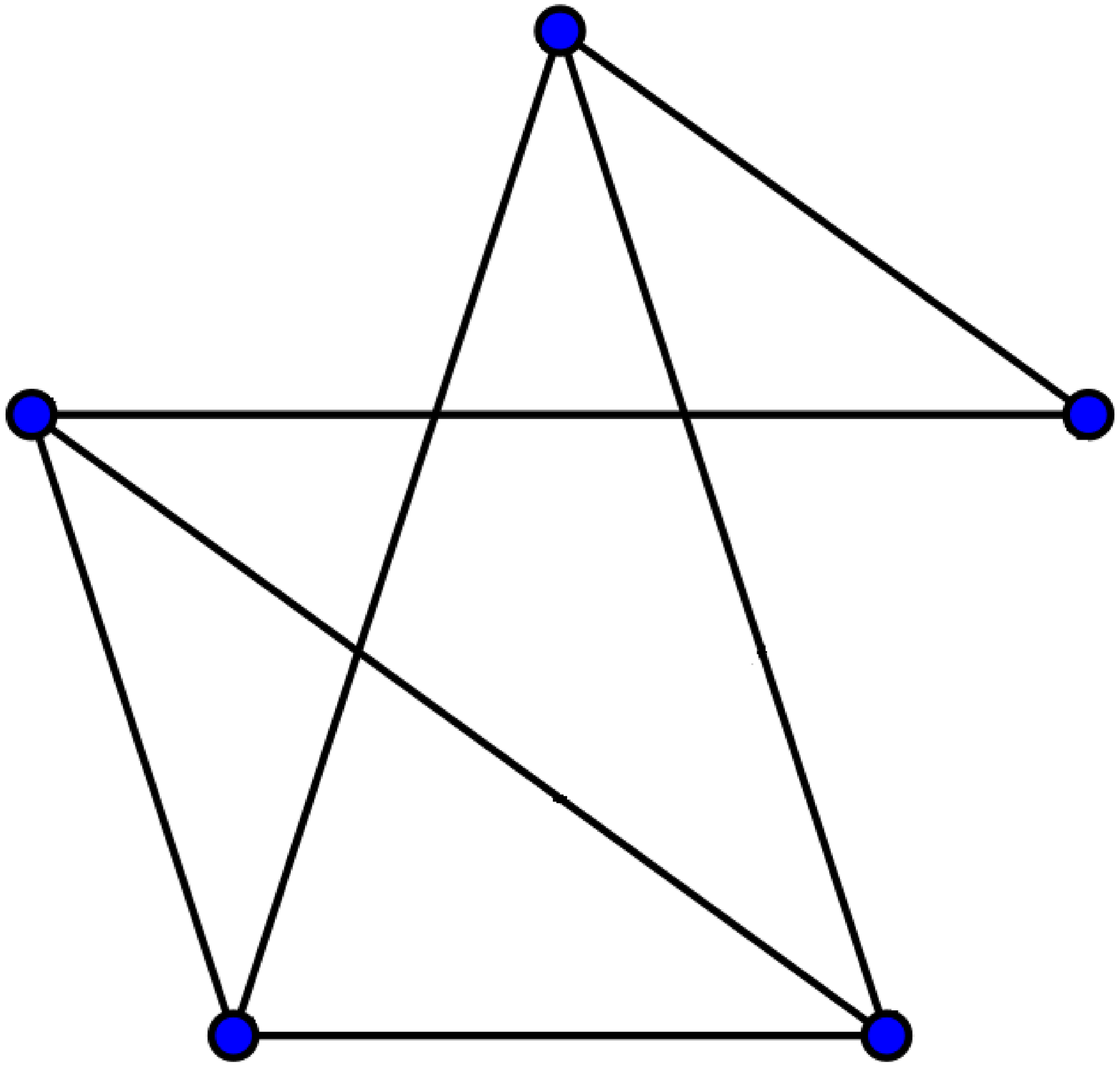}
    \label{fig:subfig1}
}
\ \ \ \ \ \ \ \ \ 
\subfigure[Desconexo]{
    \includegraphics[scale=0.19]{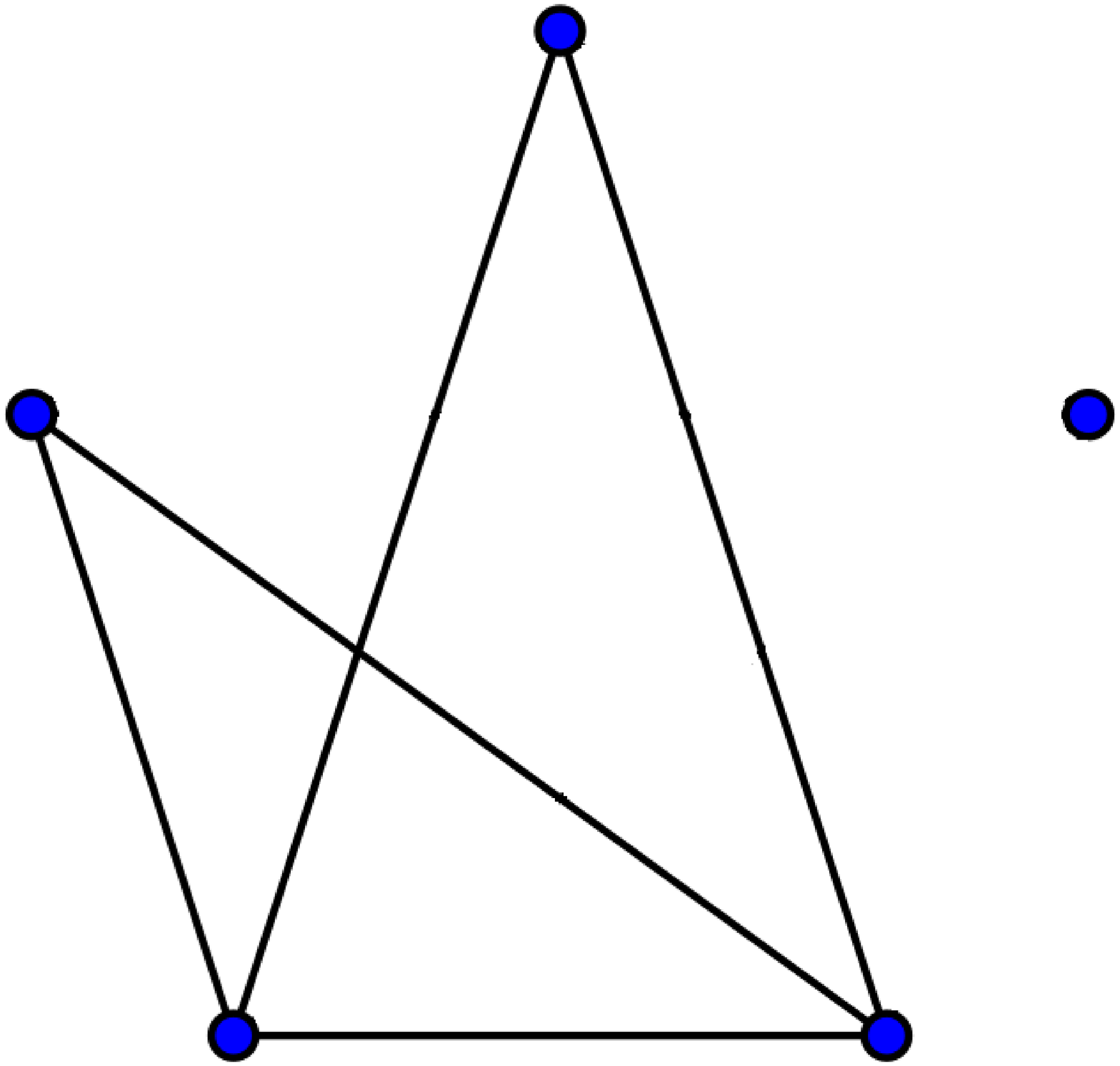}
    \label{fig:subfig1}
}
\caption{Exemplos de grafo conexo (a) e desconexo (b).}
\label{fig:conexo}
\end{figure}

Um caminho (para o caso de um grafo simples\footnote{1-grafo, multiplicidade de arestas entre dois vértices igual a 1.}) com a origem no vértice $v_i$ e o fim no vértice $v_f$ é definido por um conjunto finito de arestas consecutivas partindo do vértice $v_i$ e chegando ao vértice $v_f$.
Portanto, há um caminho entre os vértices $v_1$ e $v_2$ se existir um conjunto de arestas que interligue esses dois vértices diretamente ou indiretamente. Um grafo é conexo se para qualquer par de vértices $\{v_1,v_2\}$, existe um caminho com início em $v_1$ e término em $v_2$. A figura \ref{fig:conexo} exemplifica um grafo conexo e 
outro desconexo. 

\subsection{Problema da Clique e do Subconjunto Independente}
\label{ssc:psi}

\begin{figure}
\centering
  \includegraphics[scale=0.18]{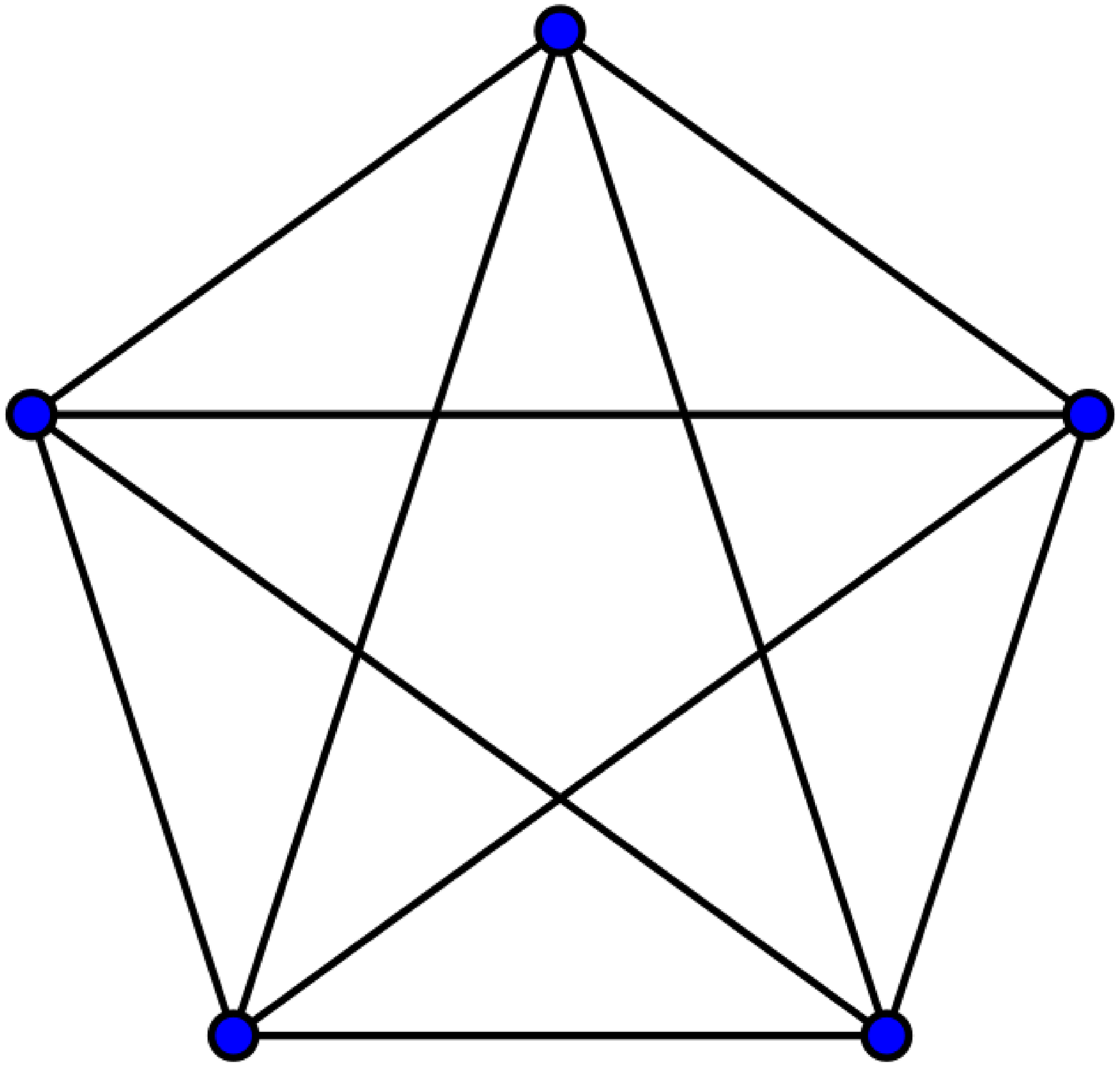}
\caption{Exemplo de Clique.}
\label{fig:clique}
\end{figure}

Seja um grafo $G = (V, E)$. A clique (ou subconjunto completo) é qualquer conjunto de vértices dois a dois adjacentes. Um subconjunto $U$ de vértices é uma clique se o grafo induzido $G[U]$ é completo (figura \ref{fig:clique}). O tamanho da clique é definido pela sua cardinalidade e um dos problemas mais conhecidos é o de identificar a maior clique existente em um grafo, denominado problema da Clique Máxima. Esse problema é NP-Difícil.

\begin{figure}[!htb]
\centering
\subfigure[SIV]{
    \includegraphics[scale=0.18]{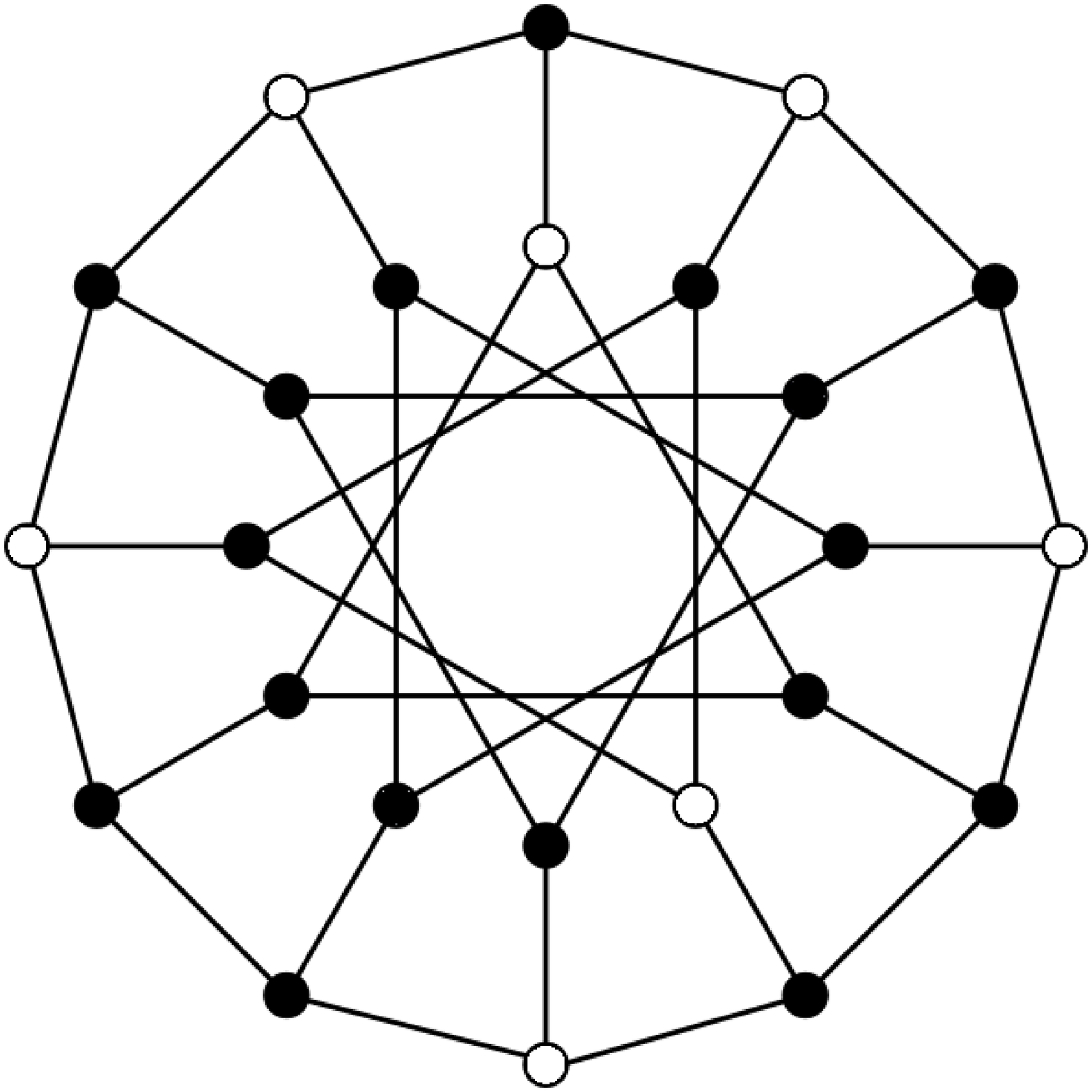}
    \label{fig:subfig1}
}
\ \ \ \ \ \ \ \ \ 
\subfigure[SIM]{
    \includegraphics[scale=0.19]{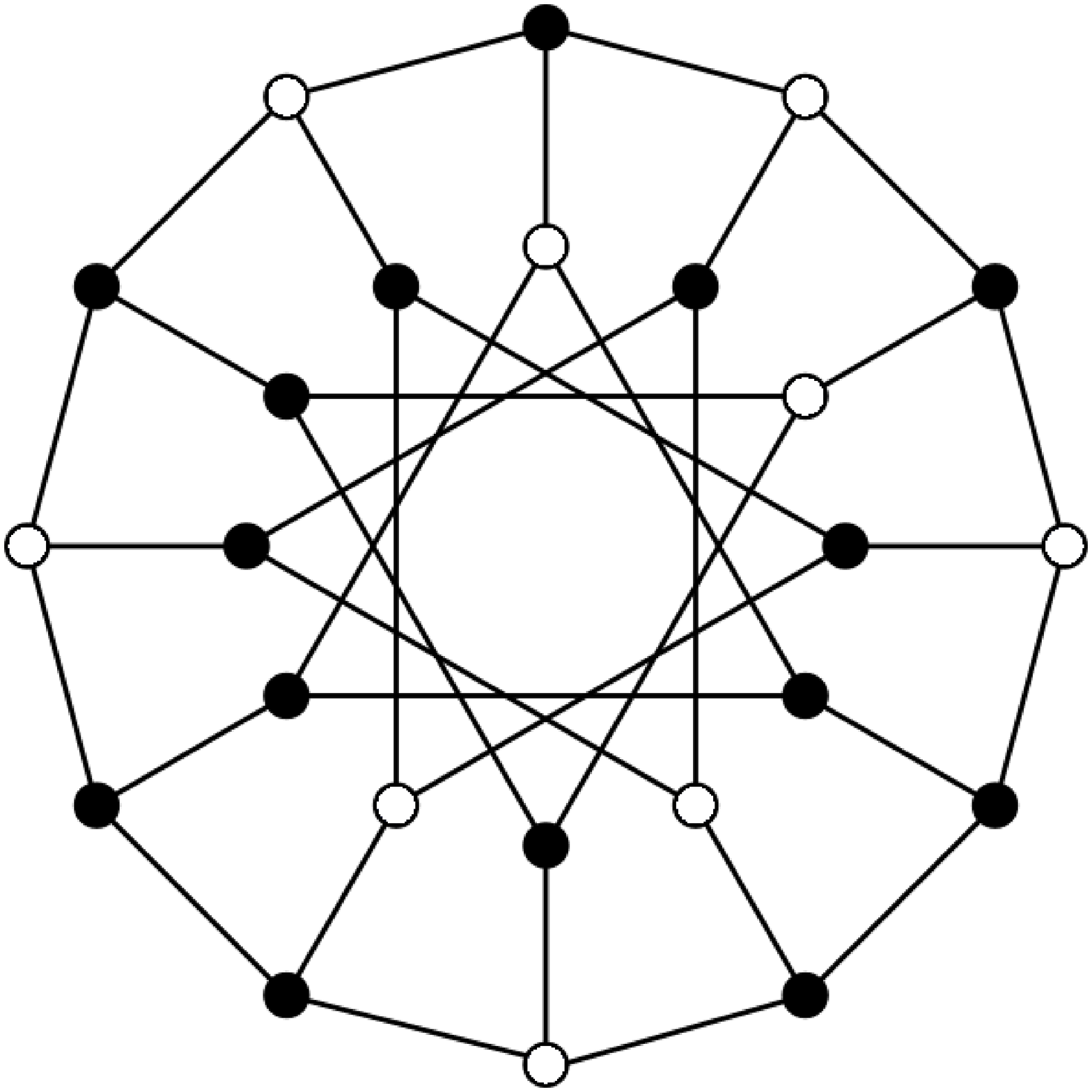}
    \label{fig:subfig1}
}
\caption{Exemplo de \ac{SIV} (a) e \ac{SIM} (b).}
\label{fig:cim}
\end{figure}

Um \acf{SIV} (denotado dessa forma nesta dissertação) é formado por um subconjunto dos vértices do grafo que não são adjacentes \cite{Garey}, ou seja, vértices que não possuem aresta interligando-os (figura \ref{fig:cim}). A Clique e o \ac{SIV} são problemas complementares. Portanto, um conjunto $U$ de vértices é uma clique em um grafo $G$ se e somente se $U$ é \ac{SIV} no grafo complementar $\bar{G}$.

O \acf{SIM} é o subconjunto independente de vértices com maior cardinalidade no grafo. O \ac{SIM} está presente em várias aplicações como coloração de grafos, agendamento de tarefas, atribuição de canais de rádio, entre outras.

\cite{Butenko} abordou a complexidade algorítmica e possíveis soluções para o \ac{SIM} adaptadas às características do grafo. \cite{Rossi} desenvolveram um algoritmo \textit{branch-and-cut} \cite{Laurence} para solucionar o problema do \ac{SIM}. Eles criaram restrições de estrutura geral a partir da execução de algoritmos de separação de clique sobre um grafo modificado obtido a partir da projeção das arestas.
\chapter{Sistemas de Sumarização Automática de Textos na Literatura}
\label{ch:lit}

Este capítulo aborda alguns sistemas \textit{baseline}\footnote{Sistemas \textit{baseline} são sistemas simples que servem de comparação de 
desempenho em relação a outros sistemas.} e da literatura para a sumarização automática de textos. Serão apresentados os sistemas Artex, 
Cortex, Enertex, GistSumm, KLSumm, baseados em \ac{PLI}, LexRank e TextRank. Alguns deles serão comparados com os sistemas desenvolvidos 
nesta dissertação.

\section{Artex}
\label{sc:artex}

\begin{figure}[!htb]
     \centering
      \includegraphics[scale=0.25]{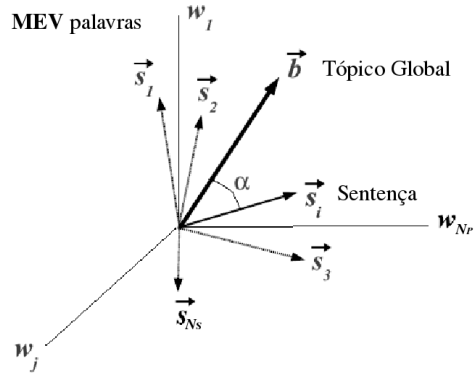}
     \caption{\label{fig:vet} \ac{MEV} do tema global.}
\end{figure}

\ac{Artex} é um sistema de sumarização por extração \cite{artex}. O pré-tratamento é o mesmo descrito no sistema \ac{SASI} (seção 
\ref{sc:sasi}). O sistema recebe a matriz de saco de palavras e cria o \acf{MEV} que possibilita a construção do vetor médio do documento 
(o ``tópico global'') de todos os vetores das sentenças. Ao mesmo tempo, obtém-se o ``peso lexical'' de cada sentença, que é o número de palavras na sentença. Analisa-se a similaridade entre as sentenças em relação ao ``peso lexical" e o ``tópico global", a figura \ref{fig:vet} descreve a \ac{MEV} de um documento em que as 
sentenças vetoriais $\vec{s_i}$ e o vetor do tópico global $\vec{b}$ são representados em um espaço de palavras de dimensão $N_P$.

\begin{figure}[!htb]
     \centering
      \includegraphics[scale=0.25]{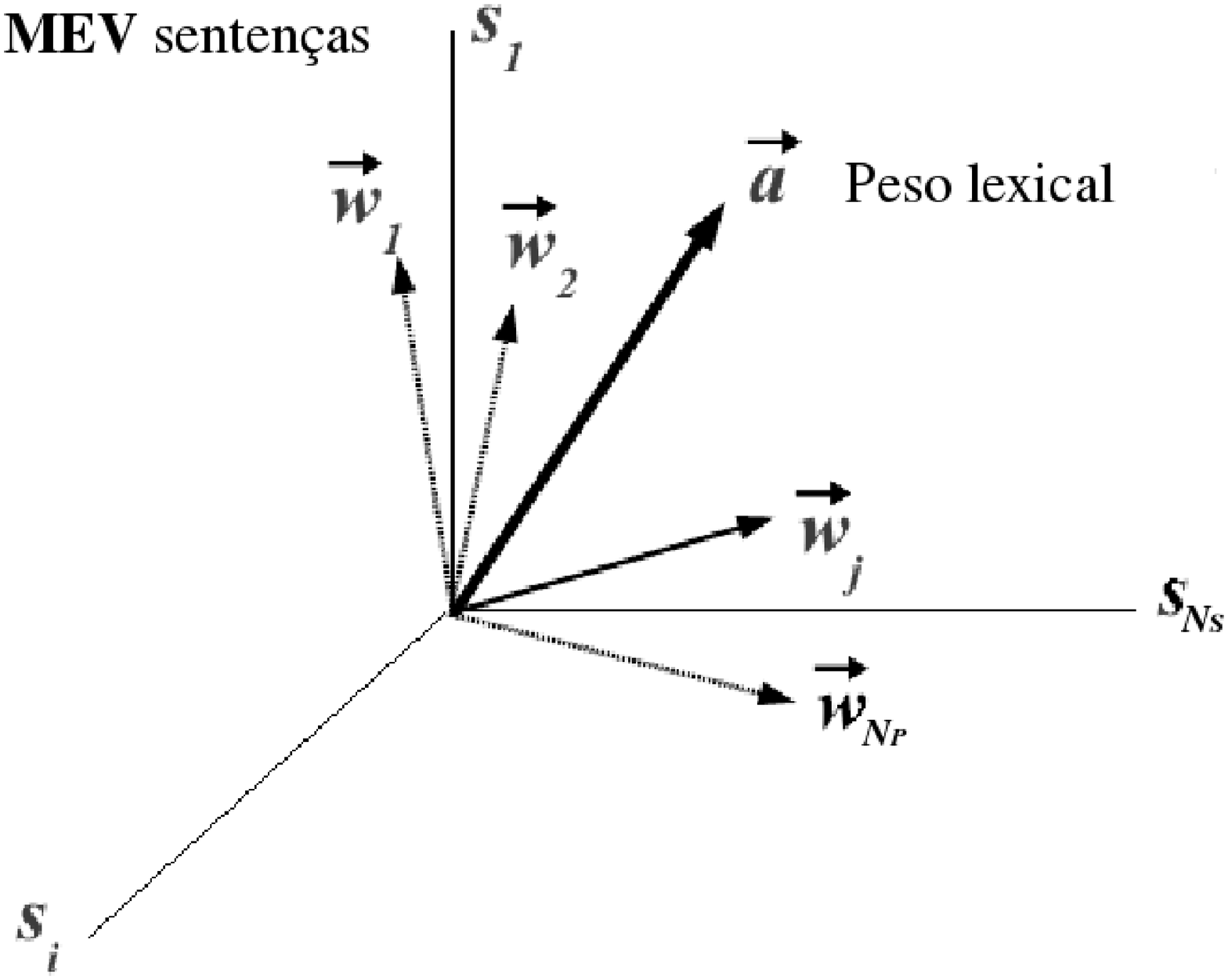}
     \caption{\label{fig:lexical} \ac{MEV} do peso lexical.}
\end{figure}

 A figura \ref{fig:lexical} descreve o peso lexical $\vec{a}$ em um \ac{MEV} de sentenças com dimensionalidade de $N_S$, sendo $\vec{w_i}$ o vetor da frequência da palavra $i$ nas $N_S$ sentenças. A relevância das sentenças é calculada baseada na proximidade do vetor de cada sentença em relação ao vetor de ``tópico global" e ao vetor de ``peso lexical". A relevância ou o peso da sentença $s_i$, denominado $score(s_i)$, é calculada pela equação \ref{eq:artex}.

\begin{equation}
\label{eq:artex}
 score(s_i) = \frac{1}{N_S \times N_P} \left( \sum_{j=1}^{N_P} nos(i,j) \times nod(j) \right) \times a_i ; i = 1,2, ..., N_S;
\end{equation}

\noindent sendo $nos(i,j)$ é o número médio de ocorrências da palavra $j$ na sentença $i$, $nod(j)$ é o número médio de ocorrências da palavra $j$ nas $N_S$ sentenças e $a_i$ é o número médio de ocorrências das $N_P$ palavras na sentença $i$.

Por fim, o sistema gera o resumo composto pelas sentenças com os maiores \textit{scores}.

\section{Cortex}
\label{sc:cortex}

\begin{figure}[!htb]
     \centering
      \includegraphics[scale=0.45]{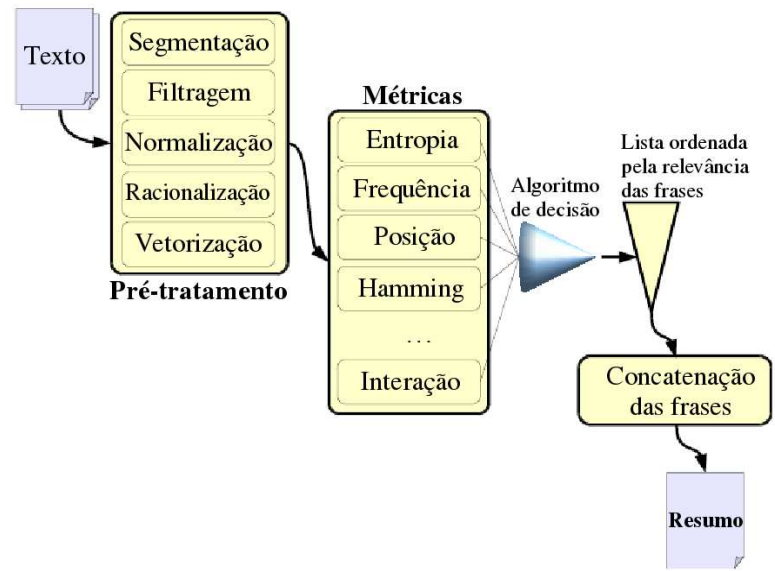}
     \caption{\label{fig:siscortex} Funcionamento do sistema Cortex.}
\end{figure}

O sistema Cortex é um sistema de sumarização automática independente do idioma \cite{cortex}. O sistema aplica  cálculos numéricos para 
ponderar frases e criar um resumo. Inicialmente, realiza-se o processo de pré-tratamento descrito no sistema \ac{SASI} (seção 
\ref{sc:sasi}). O Cortex possui um sistema de decisão que utiliza as métricas de similaridade, de \textit{overlap} e \textit{score} 
normalizadas (entre $[0, 1]$), combinadas a fim de calcular a pontuação de cada sentença (figura \ref{fig:siscortex}). 

O algoritmo de decisão avalia todas as métricas permitindo uma melhor avaliação da importância das sentenças. Utiliza-se o cosseno de 
similaridade para medir o grau de similaridade entre dois vetores representando os documentos e os tópicos abordados.

O \textit{overlap} calcula a quantidade de informação compartilhada de uma sentença (conjunto de palavras $P$) em relação a um determinado 
tópico (conjunto de palavras $Q$) a partir da quantidade de palavras comuns entre eles (equação \ref{eq:ol}).

\begin{equation}
\label{eq:ol}
 overlap(P,Q) = \frac{|P \cap Q|}{|Q|}
\end{equation}

As métricas de similaridade e \textit{overlap} são usadas para refinar o cálculo do Cortex descrito no trabalho de Torres-Moreno \textit{et 
al} \cite{Cor1}. Assim, a ponderação final (\textit{score}) de uma frase $P$ com relação a um documento (conjunto de palavras $W$) e um 
tópico $Q$ é definida pela equação \ref{eq:scorecortex}.

\begin{equation}
\label{eq:scorecortex}
 score(P) = \alpha_{0} \times CORTEX(P,W) + \alpha_{1} \times overlap(P,Q) + \alpha_{2} \times Sim(W,Q); \sum_i \alpha_{i} = 1
\end{equation}

O resumo é criado a partir das sentenças com maiores pontuações e sem redundâncias. A redundância das sentenças é calculada a partir da similaridade de Dice \cite{dice}.

\section{Enertex}
\label{sc:enertex}

O sistema Enertex avalia a energia textual das sentenças, que representa a relevância das mesmas em relação ao texto \cite{enertex}. O 
sistema modela um texto como uma rede neural baseada no modelo de Hopfield e de Ising para calcular a energia textual das sentenças. O 
pré-tratamento do texto é o mesmo descrito no \ac{SASI} (seção \ref{sc:sasi}), com a criação da matriz de palavras $\textbf{S}_{[N_S \times 
N_P]}$. Em seguida, o Enertex utiliza a matriz $\textbf{S}$ para analisar a interação entre os termos do vocabulário do texto a partir da 
regra de Hebb (equação \ref{eq:j}).

\begin{equation}
\label{eq:j}
 \textbf{K} = \textbf{S}^T \times \textbf{S},
\end{equation}

\noindent onde $\textbf{S}^T$ é a matriz transposta de $\textbf{S}$. Baseada nessa interação, a energia textual ($\textbf{Energia}$) é 
calculada através da equação \ref{eq:enertex}.

\begin{equation}
\label{eq:enertex}
 \textbf{Energia} = - \frac{1}{2} \textbf{S} \times \textbf{K} \times \textbf{S}^T; \textbf{Energia}_{ij} \in \textbf{Energia}_{N_S 
\times N_S}
\end{equation}

\noindent em que $\textbf{Energia}_{ij}$ representa a energia da interação entre as sentenças $i$ e $j$. A relevância das sentenças é 
calculada considerando sua energia textual e, por fim, cria-se o resumo (sem redundâncias) a partir das frases consideradas mais relevantes 
pelo sistema.

\section{GistSumm}
\label{sc:gist}

O sistema GistSumm \cite{gist} baseia-se em encontrar a mensagem principal do documento para a criação de seu resumo. O sistema segmenta o texto e calcula a relevância das sentenças através das palavras-chaves e do \ac{TF-ISF} das sentenças. Dessa forma, as sentenças com maior \textit{score} são consideradas as mais relevantes para o documento. O sistema avalia a pontuação das sentenças determinando seus limiares e seleciona as sentenças que possuam palavras com radicais correspondentes às palavras na sentença principal e com um \textit{score} acima do limiar.

\section{KLSumm}
\label{sc:klsumm}

O algoritmo KLSumm baseia-se no cálculo da divergência \ac{KL} (seção \ref{ssc:dkl}) para analisar a divergência entre um sumário e seu 
texto original \cite{klsumm}. O objetivo desse algoritmo é tornar o sumário (conjuntos de palavras $R$) o mais similar possível do texto 
(conjunto de palavras $O$) respeitando o limite de caracteres do sumário (equação \ref{eq:klsumm}). Dessa forma, o sistema seleciona as 
sentenças menos divergentes em relação ao texto, conforme a equação \ref{eq:klsumm}.

\begin{equation}
 \label{eq:klsumm}
 Resumo = min\ D_{KL}(R||O)
\end{equation}

\section{LexRank}
\label{sc:lexrank}
 
O LexRank avalia estocasticamente a relevância das sentenças \cite{lexrank}. O sistema utiliza um grafo de palavras construído a partir do 
texto e o modela como um grafo de sentenças. O sistema é dividido em duas partes: similaridade das sentenças e centralidade de uma sentença 
com relação às demais.

O cálculo da similaridade é baseado no \ac{TF-ISF} e no cosseno de similaridade (equação \ref{eq:lexrank1}), sendo $tf(w,P)$ a frequência 
da palavra $w$ na sentença $P$ e $idf(w)$ o valor \textit{inverse document frequency} (inverso da frequência nos documentos) da palavra $w$.

\begin{equation}
    \label{eq:lexrank1}
    lexrank_{sim}(P,Q) = \frac{\sum_{w \in P,Q}\ tf(w,P) \times tf(w,Q) \times [idf(w)]^2}{\sqrt{\sum_{p_i \in P}[tf(p_i,P) \times 
idf(p_i)]^2} \times \sqrt{\sum_{q_i \in Q}\ [tf(q_i,Q) \times idf(q_i)]^2}}
\end{equation}

A partir da similaridade ($lexrank_{sim}$), calcula-se a relevância (\textit{score}) de cada sentença baseada na centralidade das sentenças 
(equação \ref{eq:lexrank2}). A variável $d$ considera a ponderação inicial dos vértices.

\begin{equation}
	\label{eq:lexrank2}
	p(u) = \frac{d}{N_S} + (1-d) \sum_{v \in viz(u)} \frac{lexrank_{sim}(u,v)}{\sum_{z \in viz(v)}lexrank_{sim}(z,v)} p(v)
\end{equation}

\noindent onde $p(u)$ é a centralidade do vértice e $viz(u)$ são os vértices vizinhos de $u$.

O sistema criará um resumo com as sentenças de maior score e sem redundância.

\section{Programação Linear Inteira}
\label{sc:spli}

\cite{mcdonald2006} propôs um modelo matemático para avaliar a qualidade do resumo baseado na relevância e similaridade das sentenças. Seja 
um conjunto de sentenças de um texto $bf_i$ indicará a existência da sentença $i$ no resumo do texto, $Doc$ será o conjunto de sentenças do 
documento, $Redud_{ij}$ representará a redundância entre as sentenças $i$ e $j$ e por sua vez $Relev_i$ representará a relevância da 
sentença $i$. Dessa forma, um resumo pode ser avaliado pela fórmula descrita na equação \ref{eq:pli}.

\begin{equation}
\label{eq:pli}
 Resumo = \sum_{i \in Doc} Relev_i \times bf_i - \sum_{i,j \in Doc} Redud_{ij} \times bf_i \times bf_j
\end{equation}

Mcdonald modela o problema usando Programação Linear, considerando $L$ como o limite máximo de caracteres do resumo e $l_i$ o tamanho da 
sentença $i$. A análise da redundância é uma função quadrática, portanto é necessário linearizar a ocorrência simultânea das frases $i$ e $j$ ($r_{ij} = bf_i \times bf_j$). Assim, a modelagem \ac{PLI} utilizada por Mcdonald é expressa pelas equações \ref{eq:1}, \ref{eq:2}, \ref{eq:3}, \ref{eq:4} e \ref{eq:5}.

Maximizar:  
  \begin{equation}
  \label{eq:1}
  \sum_{i \in Doc} Relev_i \times bf_i\ -\ \sum_{i,j \in Doc} Redud_{ij} \times r_{ij}
  \end{equation}
	   
Restrições: 
  \begin{equation}
  \label{eq:2}
  \sum_{j \in Doc} l_j \times bf_j \leq L
  \end{equation}
  \begin{equation}
  \label{eq:3}
  r_{ij} \leq bf_i \ \ \ r_{ij} \leq bf_j \ \ \forall i,j
  \end{equation}
  \begin{equation}
  \label{eq:4}
  bf_i\ +\ bf_j\ -\ r_{ij} \leq 1\ \ \forall i,j
  \end{equation}
  \begin{equation}
  \label{eq:5}
  bf_i \in \{0,1\} \ \ \forall i
  \end{equation}
  
\section{Submodular}
\label{sc:submod}

Seja $B$ um conjunto finito e seja $f$ uma função de partes de $B$ em $\mathbb{R}$. A função $f$ é submodular se para quaisquer partes $C$ e 
$D$ de $B$, observa-se a relação  expressa na equação \ref{eq:subsubsub}.

\begin{equation}
\label{eq:subsubsub}
 f(C \cup D) + f(C \cap D) \leq f(C) + f(D)
\end{equation}

Correlacionando a função submodular com a extração automática, definiu-se $B$ como o conjunto com todas as sentenças do documento, $C$ é 
resumo do texto e $f$ é a função de avaliação de qualidade do resumo. A sumarização objetiva encontrar o conjunto de sentenças (conjunto 
$C$), que melhor represente o texto (conjunto $B$) e tenha um tamanho $|C| \leq L$ (tamanho máximo do resumo) (equação \ref{eq:sub}). Esse 
problema é NP-Difícil. Entretanto, é possível determinar uma solução aproximada através  de um algoritmo guloso se a função $f$ for 
submodular \cite{submodular}.

\begin{equation}
\label{eq:sub}
 \max_{C \subseteq B}\{ f(C): |C| \leq L \}
\end{equation}

O trabalho \cite{submodular} descreve quatro funções submodulares para avaliar a qualidade de resumos automáticos. A primeira calcula a 
representatividade de um resumo com relação ao texto (equação \ref{eq:fac}) analisando a similaridade do resumo $C$ com relação ao texto 
$B$, sendo $sim_{ij}$ a similaridade entre as sentenças $i$ e $j$. 

\begin{equation}
\label{eq:fac}
 f_{fac}(C) = \sum_{i \in B}\ \max_{j \in C}\ sim_{ij}
\end{equation}

A segunda métrica utilizada, $f_{corte}$, avalia a similaridade das sentenças do resumo em relação as demais sentenças do texto (equação \ref{eq:cor}). 
A terceira métrica, $f_{pior}$, considera o caso de uma sentença relevante não ser similar às demais. Uma forma de avaliar esse caso é 
através da maximização da similaridade da sentença menos similar do resumo (equação \ref{eq:pi}).

\begin{equation}
 \label{eq:cor}
 f_{corte}(C) = \sum_{i \in B \backslash C}\ \sum_{j \in C}\ sim_{ij}
\end{equation}

\begin{equation}
 \label{eq:pi}
 f_{pior}(C) = \min_{i \in B}\ \max_{j \in C}\ sim_{ij}
\end{equation}

A última métrica, $f_{pen}$, avalia a redundância das sentenças do resumo. Um resumo deve ser conciso e pequeno. Desse modo, a presença de 
sentenças similares não adiciona um conteúdo interessante ao resumo. Portanto, pode-se evitar essa redundância com a adição de penalidades 
às sentenças similares do resumo (equação \ref{eq:pen}), em que $\lambda$ é a taxa de penalidade caso haja redundância de sentenças.

\begin{equation}
 \label{eq:pen}
 f_{pen}(C) = \sum_{i \in B \backslash C}\ \sum_{j \in C} sim_{ij} - \lambda \sum_{i,j \in C:i \neq j} sim_{ij}, \lambda \geq 0
\end{equation}

\section{TextRank}
\label{sc:textrank}

O TextRank é um algoritmo baseado em grafos para mensurar a relevância das sentenças \cite{textrank}. O algoritmo baseia-se na recomendação 
(ou relevância) de cada vértice baseada nas ligações entre os mesmos. Portanto, quanto maior a quantidade de ligações associadas a um 
vértice, mais relevante ele será para o texto.

O algoritmo modela o texto como um grafo de sentenças e cria as arestas baseadas na similaridade entre as sentenças $P$ e $Q$ (equação \ref{eq:simtr}).

\begin{equation}
 \label{eq:simtr}
 sim(P,Q) = \frac{|\{w|w \in P\ \&\ w \in Q\}|}{\log (|P|) + \log (|Q|)}
\end{equation}

O TextRank mensura o \textit{score} de cada vértice de forma recursiva utilizando a vizinhança dos vértices até seus valores se tornarem estáveis (equação \ref{eq:textrank}).

\begin{equation}
 \label{eq:textrank}
 Rec(v_i) = (1-p) + p \times \sum_{v_j \in viz(v_i)} \frac{sim(v_j,v_i)}{\sum_{v_k \in viz(v_j)} sim(v_j,v_k)} Rec(v_j),
\end{equation}

\noindent em que a variável $p$ é um amortecimento ajustado entre 0 e 1, $sim(v_j,v_i)$ é a similaridade entre os vértices $v_j$ e $v_i$, e $viz(v_i)$ representa o conjunto de vértices da vizinhança do vértice $v_i$.  

Finalmente, o sistema cria o resumo com as sentenças de maior \textit{score}.
\chapter{Sistemas propostos}
\label{ch:sist}

Este capítulo aborda os sistemas sumarizadores LIA-RAG, \ac{RAG}, \ac{SASI} e \ac{SUMMatrix}. Os sistemas utilizam a Teoria de Grafos (LIA-RAG, \ac{RAG} e \ac{SASI}) e matrizes de similaridade (\ac{SUMMatrix}) juntamente com métricas de similaridade para avaliar a divergência entre as sentenças e selecionar as principais informações de um texto. Os sistemas desenvolvidos são sumarizadores genéricos, monodocumento e multidocumento, indicativos com as informações-chaves e geram os resumos através da extração de sentenças. Os sistemas nesta dissertação possuem características particulares e assim, requereram o uso de corpus específicos ao problema configurado. A seção \ref{sc:sasi} descreve a motivação e o funcionamento geral dos sistemas \ac{SASI}. As seções \ref{sc:rag} e \ref{sc:ragc} descrevem o funcionamento dos sistemas \ac{RAG} e LIA-RAG. Por fim, a seção \ref{sc:summatrix} descreve o sistema SUMMatrix e suas características.

\section{\acf{SASI}}
\label{sc:sasi}

Um texto é composto por diversas sentenças, que podem ser agrupadas de acordo com o grau de similaridade 
entre elas. Cada grupo aborda uma etapa/ideia do texto. Considera-se neste trabalho que uma sentença similar a todas (ou à maioria) 
pertencentes ao seu grupo possui conteúdo ``relevante'' com relação ao grupo. Dessa forma, é possível criar um resumo com a ideia geral do 
texto utilizando as sentenças com maior similaridade identificadas em cada grupo. Uma outra abordagem possível consiste em obter resumos 
contendo apenas as informações mais importantes, ou seja, analisar somente os maiores grupos de sentenças similares visto que suas 
informações são constantemente discutidas no texto.

O sistema \ac{SASI} analisa e cria grupos de sentenças similares do texto para identificar e gerar um resumo com as principais sentenças de cada grupo. 

\subsection{Funcionamento}
\label{ssc:algoSasi}

Inicialmente, executa-se o pré-tratamento do documento realizando o processo de leitura e o reconhecimento de caracteres do texto. Em 
seguida, realiza-se a segmentação das sentenças e palavras, a filtragem de \textit{stopwords} e o processo de \textit{stemming} a fim de 
remover palavras irrelevantes e reduzir as palavras às suas raízes. Por fim, cria-se uma matriz de palavras correlacionando sua frequência 
em cada sentença. De posse dessa matriz de palavras, o \ac{SASI} calcula a divergência das sentenças. 

Realiza-se, igualmente, uma filtragem das sentenças sem relevância para o texto e verifica-se a divergência entre elas. Para isso, foi 
utilizada a abordagem descrita na seção \ref{sc:div}. Então, sempre dois conjuntos de palavras $P$ e $Q$ são analisados concomitantemente. 
Inicialmente, $P$ e $Q$ referenciam, respectivamente, o texto completo e cada frase isolada do mesmo. Dessa forma, calcula-se a divergência 
entre os 
conjuntos $P$ e $Q$, $D_{JS}(P||Q)$, para verificar se uma frase é importante para o documento por meio da análise das palavras existentes:

\begin{itemize}
 \item Se a divergência da frase for grande em relação ao texto completo, a frase será descartada;
 \item Se o documento possuir um título: analisa-se a similaridade entre o título e a frase descartada, caso a divergência entre eles seja pequena, a frase fará parte do documento novamente.
\end{itemize}

Em seguida, o \ac{SASI} cria um grafo $G$ em que os vértices representam as sentenças inicialmente relevantes. Nesse caso, os conjuntos $P$ e $Q$ correspondem a pares de sentenças e sua divergência 
será igualmente utilizada para a criação e para a ponderação das arestas. Caso a divergência entre duas sentenças seja ``pequena'' 
(parâmetro do algoritmo), uma aresta interligando-as será inserida. 

\begin{figure}[!htb]
     \centering
      \includegraphics[width=13cm]{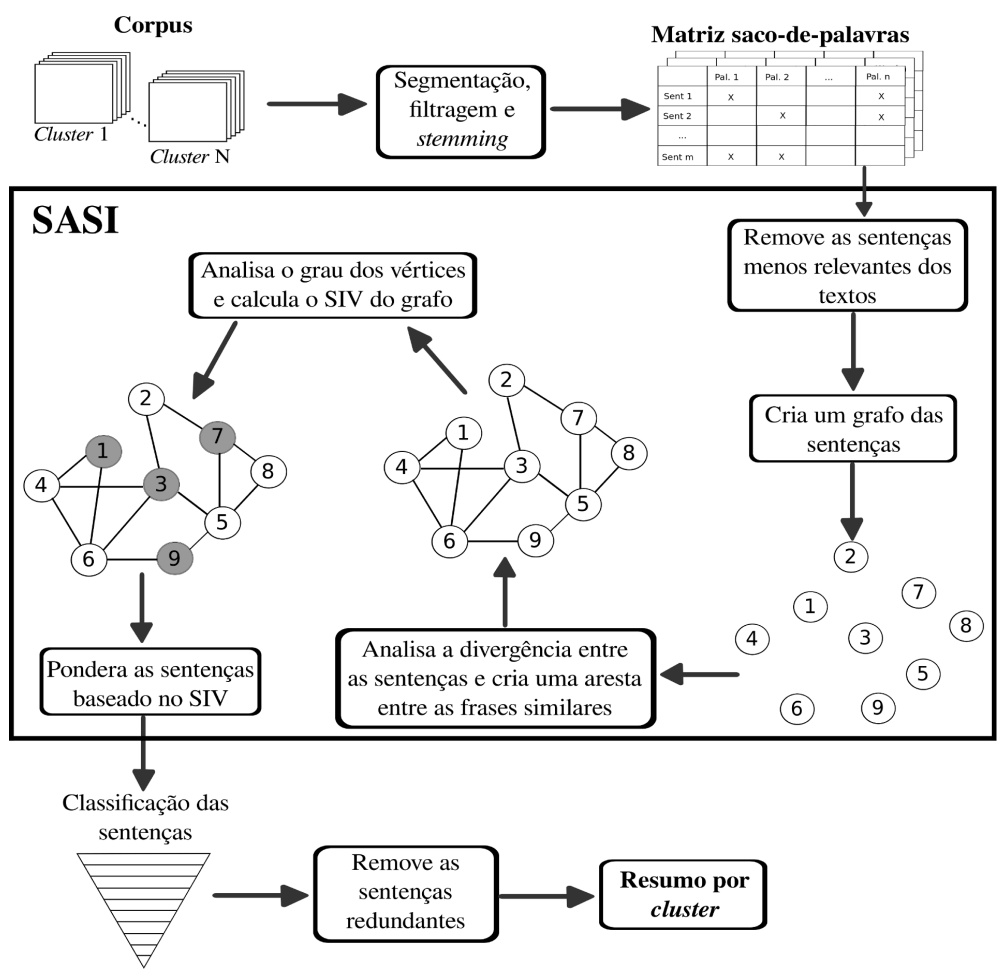}
     \caption{\label{fig:sasi} Funcionamento do sistema SASI.}
\end{figure}

A sumarização objetiva a produção de um resumo de tamanho pequeno e com as principais informações do texto. Desse modo, o \ac{SASI} deverá escolher somente as sentenças com mais informações adicionais para construir o resumo. Portanto, a escolha de vértices adjacentes implica na seleção de frases com conteúdo redundante, o que não é interessante para esse tipo de resumo. O sistema deverá selecionar as frases com conteúdo distinto entre si, descartando as sentenças repetitivas ou que trazem pouca informação adicional ao resumo. A determinação de um \acf{SIV} do grafo fornece uma possível solução para o problema devido suas características de selecionar vértices não conectados entre si. Os vértices com maior grau tem preferência no resumo pois, teoricamente, são similares a um número maior de sentenças e, portanto, são mais relevantes. 
Dessa forma, o sistema ordena os vértices de acordo com o grau deles em ordem decrescente. Em seguida, o SASI avalia os vértices nessa ordem e seleciona aqueles que mantenham as características do \ac{SIV}.
O resumo é composto pelas frases que compõem o \ac{SIV} excluindo as frases que são redundantes com base no coeficiente de Dice \cite{dice} (figura \ref{fig:sasi}).

\subsection{Exemplo}
\label{ssc:exemploSasi}

O texto da tabela \ref{tb:ex_sasi} retrata uma notícia sobre a política brasileira. Inicialmente realiza-se os processos de segmentação, filtragem, 
\textit{stemming} e criação da matriz de palavras. Os textos não possuem títulos, então todas as sentenças são consideradas relevantes. O 
sistema recebe a matriz e cria um grafo de sentenças. O \ac{SASI} calcula a divergência entre as sentenças e caso a divergência entre elas 
seja abaixo de $0,20$ (valor obtido de forma empírica), adiciona-se uma aresta interligando os dois vértices, i.e., as duas sentenças. Em 
seguida, o sistema calcula o \ac{SIV} privilegiando os vértices de maior grau. Por fim, o resumo do texto é 
composto pela concatenação das sentenças selecionadas no \ac{SIV}, sem redundâncias. Nesse exemplo, o resumo com 100 
palavras é composto pelas frases: $S-9$, $S-13$, $S-18$, $S-19$ e $S-21$.

\begin{tabela}[!htb]
\begin{tabular}{|p{0.9\columnwidth}|}
\hline
 {\footnotesize \textbf{Texto}}\\
 {\footnotesize S-1 Termina hoje, às 20 horas, o prazo para que os deputados acusados de participar do esquema dos sanguessugas renunciem 
para escapar da abertura de processo por quebra de decoro parlamentar.} \\
 {\footnotesize S-2 A expectativa de lideranças da Câmara e do Conselho de Ética é que pouco mais de 10\% dos 69 deputados denunciados no 
relatório parcial da CPI dos Sanguessugas abrirão mão de seus mandatos.} \\
 {\footnotesize S-3 Integrante da cúpula da Câmara que, nos últimos dias, conversou com ao menos 30 parlamentares acusados no caso calcula 
que sete podem renunciar - Nilton Capixaba (PTB-RO), Marcelino Fraga (PMDB-ES), César Bandeira (PFL-MA), Benedito Dias (PP-AP), Carlos
Nader (PL-RJ), João Caldas (PL-AL) e Reginaldo Germano (PP-BA).} \\
 {\footnotesize S-4 Ex-líder do PP, Pedro Henry (MT) cogitou sair da função, mas teria desistido da ideia. }\\
 {\footnotesize S-5 Até ontem, só Coriolano Sales (PFL-BA) havia apresentado renúncia. }\\
 {\footnotesize S-6 Ele não quis arriscar a chance de assumir a prefeitura de Vitória da Conquista. }\\
 {\footnotesize S-7 Segundo colocado em 2004, Sales processou seu adversário por abuso do poder econômico e aguarda resultado. }\\
 {\footnotesize S-8 ``Não dá para avaliar quantos vão renunciar'', disse ontem o presidente do Conselho de Ética, Ricardo Izar (PTB-SP). }\\
 {\footnotesize S-9 Ele vai instaurar o processo contra os deputados envolvidos com a máfia dos sanguessugas amanhã, às 10h30. }\\
 {\footnotesize S-10 Formalizada antes da abertura, a renúncia cessa o procedimento. }\\
 {\footnotesize S-11 Izar pretende que os casos dos 15 parlamentares que receberam depósito na própria conta bancária ou na de parentes 
sejam os primeiros julgados pelo Conselho. }\\
 {\footnotesize S-12 ``Vou instaurar todos os processos juntos, mas a ideia é que os 15 casos mais graves, que têm provas contundentes, 
sejam julgados na frente'', afirmou Izar. }\\
 {\footnotesize S-13 O horário-limite para que o parlamentar renuncie - 20 horas - foi estabelecido pela direção da Câmara a fim de que o 
ato seja oficializado com a sua publicação já no Diário Oficial do Congresso de amanhã. }\\
 {\footnotesize S-14 A maioria dos 69 deputados acusados de envolvimento com a máfia dos sanguessugas é candidato à reeleição e, com a 
renúncia, tentará escapar do risco de cassação e da perda dos direitos políticos por oito anos.} \\
 {\footnotesize S-15 Outros parlamentares resistem à renúncia por considerarem ter chance de não sofrer punição.} \\
 {\footnotesize S-16 Segundo investigações iniciadas pela Polícia Federal, o esquema consistia no desvio de recursos públicos por meio da 
apresentação de emendas parlamentares para a compra superfaturada de ambulâncias.} \\
 {\footnotesize S-17 Dois dos 69 deputados acusados são da Mesa Diretora, mas se afastaram das funções.} \\
 {\footnotesize S-18 Contra João Caldas, por exemplo, pesam 12 pagamentos no total de R\$ 136 mil, alguns dos quais em sua própria conta.}\\
 {\footnotesize S-19 Ele, porém, resiste à ideia de abrir mão do mandato. }\\
 {\footnotesize S-20 Já Capixaba, acusado de ter recebido R\$ 646 mil, considera seriamente a hipótese de renunciar.} \\
 {\footnotesize S-21 No caso dos integrantes da Igreja Universal, a possibilidade de saída do cargo está afastada, pois os envolvidos, 
entre eles Edna Macedo (PTB-SP), foram proibidos pela direção da instituição de concorrer à reeleição.} \\
 {\footnotesize S-22 Eleitos na esteira do deputado Enéas Carneiro (Prona-SP), ex-integrantes do partido suspeitos, como Irapuan Teixeira 
(PTB-PR) e Ildeu Araújo (PP-SP), tiveram votação insignificante em 2002 e não têm chance de reeleição.} \\
 {\footnotesize S-23 Devem preferir manter o resto do mandato.} \\ \hline
\end{tabular}
\caption{Texto integrando o \textit{cluster} do corpus CSTNews \cite{cstnews}. }
\label{tb:ex_sasi}
\end{tabela}

\section{\acf{RAG}}
\label{sc:rag}

As principais ideias de um texto são geralmente analisadas e discutidas várias vezes. No entanto, não é necessário dispor de uma grande quantidade de sentenças similares para que elas tenham importância. O \ac{RAG}, que é um sistema sumarizador automático por extração de sentenças, seleciona as principais sentenças de um texto baseado na similaridade entre as sentenças e na importância das palavras.

\subsection{Funcionamento}
\label{ssc:rag_desc}

Efetua-se o pré-processamento dos textos e obtém-se a matriz de saco de palavras, assim como descrito no funcionamento do sistema SASI (seção \ref{sc:sasi}). O RAG calcula a relevância de cada sentença com base na métrica TF-ISF (equação \ref{tfisf}) e seleciona as sentenças com maior \textit{score}.

\begin{figure}[!htb]
     \centering
     \includegraphics[width=9cm]{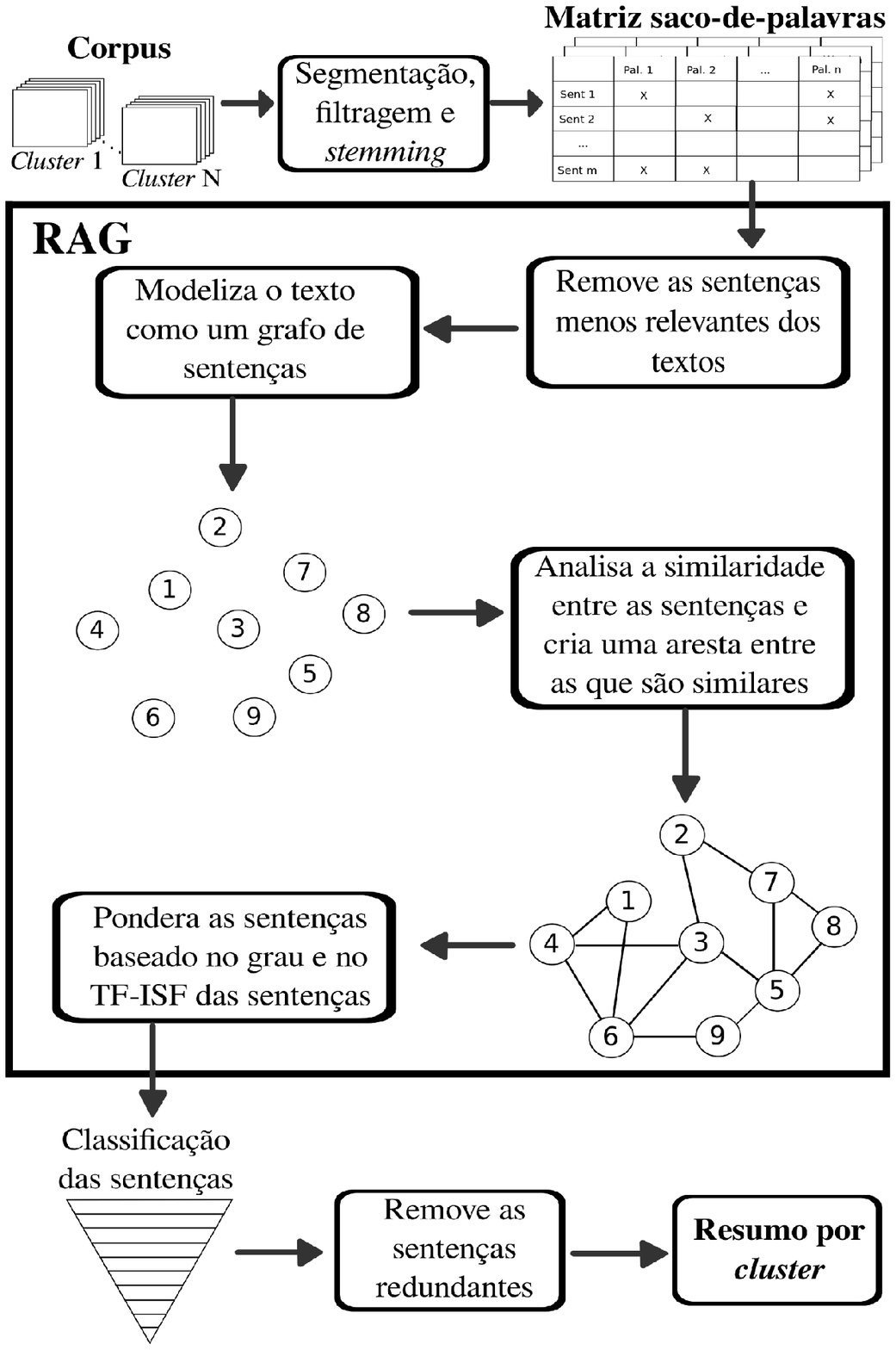}
     \caption{\label{fig:rag} Sistema RAG.}
\end{figure}

O RAG cria um grafo $G$ em que cada vértice representa uma sentença selecionada anteriormente. O texto é analisado e modelado como um grafo de sentenças (vértices). Com base na equação \ref{DJS}, o \ac{RAG} calcula a similaridade entre as sentenças. Se a divergência entre elas é menor do que limiar (obtido por testes empíricos), então o sistema criará uma aresta entre elas. Portanto, os vértices com maior grau terão o conteúdo mais analisado no texto. No entanto, algumas sentenças podem ter um grau pequeno e mesmo assim podem conter informações importantes. O RAG combina o valor do TF-ISF e do grau das sentenças para analisar sua relevância. A relevância da sentença $i$ é definida pela equação \ref{eq:score}, onde $grau(i)$ é o grau de vértice $i$ e $rel(i)$ é a relevância da sentença $i$ (equação \ref{sent}). Em seguida, o sistema cria um resumo de sentenças de pontuação mais elevada, excluindo frases similares (ou redundantes) com base no coeficiente de Dice \cite{dice}. A figura \ref{fig:rag} descreve o sistema RAG.

\begin{equation}
\label{eq:score}
score(i) = grau(i) \times rel(i)
\end{equation}

\subsection{Exemplo}
\label{ssc:exemploRAG}

O exemplo aqui descrito considera o mesmo texto da tabela \ref{tb:ex_sasi}. Inicialmente realiza-se os processos de segmentação, filtragem e 
\textit{stemming}. Em seguida, uma matriz de saco de palavras é criada a partir do texto. O RAG recebe essa matriz, calcula o 
TF-ISF das sentenças para remover as sentenças menos relevantes e modela o texto como um grafo de sentenças. Em seguida, o 
sistema calcula o \textit{score} das sentenças (tabela \ref{tb:rag}) e são selecionadas as de maior pontuação. No exemplo considerado, o 
resumo contendo 100 palavras é composto pelas sentenças: S-3, S-11, S-13, S-14 e S-18.

\begin{tabela}[!htb]
\begin{tabular}{|c|c| |c|c| |c|c|}
\hline
Sentenças 	& Relevância	& Sentenças 	& Relevância  	& Sentenças 	& Relevância	\\\hline
S-1		&	93,7	&   S-9		&  89,1		&   S-17		&  26,5		\\\hline
S-2		&	84,4	&   S-10		&  71,6		&   S-18		&  134,0	\\\hline
S-3		&	113,4	&   S-11		&  127,9	&   S-19		&  92,9		\\\hline
S-4		&	78,0	&   S-12		&  83,4		&   S-20		&  63,2		\\\hline
S-5		&	47,8	&   S-13		&  157,8	&   S-21		&  65,0		\\\hline
S-6		&	86,1	&   S-14		&  115,1	&   S-22		&  53,7		\\\hline
S-7		&	0	&   S-15		&  28,9		&   S-23		&  31		\\\hline
S-8		&	99,1	&   S-16		&  104,5	&   -----	&  ----		\\\hline
\end{tabular}
\caption{Relevância das sentenças segundo o sistema RAG.}
\label{tb:rag}
\end{tabela}

\section{LIA-RAG}
\label{sc:ragc}

O áudio é amplamente utilizado nas transmissões dos rádios e na internet (em notícias, histórias, entrevistas ou conversas). 
Existem várias ferramentas para reconhecimento de voz (áudio sobre um assunto ou uma conversa entre duas ou mais pessoas). 

A tarefa de sumarização de textos orais é mais complexa e envolve a transcrição de áudios em textos. As reuniões e conversas envolvem discussões entre várias pessoas e frequentemente suas vozes se sobrepõem. Essas transcrições possuem diversas problemáticas associadas pois é mais difícil identificar os limites de uma sentença, visto que a mesma pode ser fragmentada, conter rupturas e introduzir erros de reconhecimento de fala. A linguagem utilizada pode ser informal e as declarações podem ser parciais, fragmentárias, não gramaticais, além de possivelmente incluírem muitas reticências e pronomes. No entanto, a fala pode fornecer informações adicionais que evidenciam uma parte específica do texto como, por exemplo, a prosódia \cite{Murray}.

\cite{Mckeown} descreveram aplicações de técnicas de sumarização automática de um texto para a produção de resumos de conversas orais. Eles analisaram alguns trabalhos sobre a produção de resumos de notícias e reuniões. \cite{Murray} analisaram a sumarização extrativa de reuniões multipartidárias. Eles descreveram a relevância marginal máxima e a análise semântica latente para criar um resumo com base em características prosódicas e lexicais.

O sistema LIA-RAG foi desenvolvido a fim de sumarizar textos de conversas através da extração de sentenças. Esse sistema seleciona as principais sentenças da conversa transcrita baseado no sistema \ac{RAG} e usa um pós-processamento para remover os erros contidos em diálogos e tornar o texto mais conciso e compacto.

\subsection{Funcionamento}
\label{ssc:rag_post}

O LIA-RAG recebe um texto transcrito de uma conversa e realiza o pré-processamento do mesmo, pois estas possuem diversas expressões para associar o pensamento e o modo de falar dos personagens. O sistema remove essas expressões e características relacionadas aos participantes da conversa. Em seguida, ele analisa a relevância das sentenças utilizando o sistema \ac{RAG}, que fornece o resumo automático da conversa.

O processo de reconhecimento da voz produz um texto com vários problemas gramaticais (gírias, linguagens cotidianas, expressões e erros de reconhecimento da fala). Um algoritmo de resumo por extração seleciona as frases relevantes. No entanto, as sentenças podem ter alguns problemas gramaticais. Assim, é necessário realizar um tratamento desse resumo. Os principais aspectos analisados nesse processo são:

\begin{itemize}
  \item coloquialismos;
  \item expressões da fala;
  \item datas.
\end{itemize}

\begin{figure}[!htb]
     \centering
      \includegraphics[width=9cm]{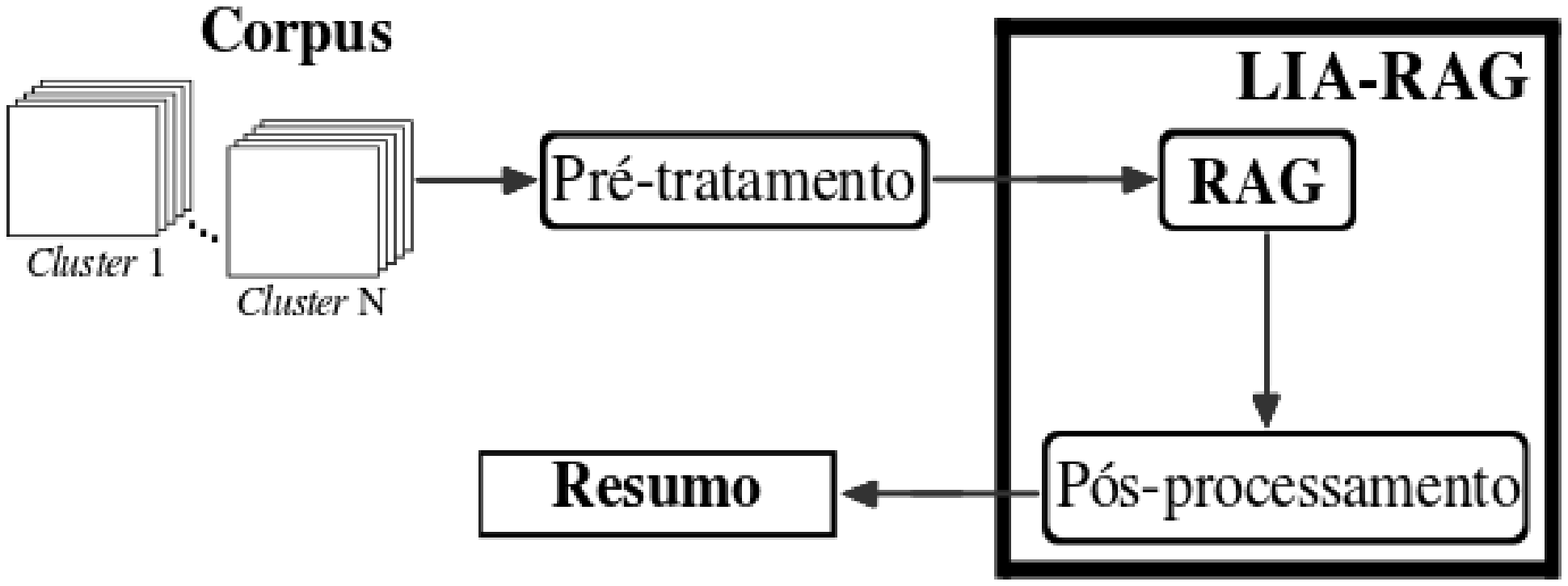}
     \caption{\label{fig:lia-rag} Sistema LIA-RAG.}
\end{figure}

Algumas expressões da fala são usadas para conectar ideias ou conceitos em conversas orais. O sistema LIA-RAG realiza o pós-processamento com o resumo gerado pelo RAG. O sistema remove essas expressões da fala porque muitas vezes elas são transcritas incorretamente (ruído na conversação). Além disso, o sistema elimina vários coloquialismos e palavras duplicadas, substitui algumas palavras erradas por sua forma correta e transforma as datas por extenso na forma ``dia/mês/ano''. A figura \ref{fig:lia-rag} mostra a arquitetura do sistema LIA-RAG.

\section{\acf{SUMMatrix}}
\label{sc:summatrix}

Atualmente, há diferentes jornais e sites de notícias disponibilizando informações sobre diferentes acontecimentos a todo momento. 
Entretanto, a análise de uma notícia a partir de uma única fonte pode fornecer informações imprecisas, duvidosas ou 
desnecessárias. A diversidade dos meios de comunicação possibilita a identificação de informações tendenciosas (desnecessárias ou erradas) 
e compreender corretamente um acontecimento a partir de diferentes perspectivas.

A utilização de várias fontes de informação nos auxilia na compreensão de um evento. Entretanto, a enorme quantidade de notícias torna 
impossível sua leitura de maneira integral. Uma maneira de resolver esse problema é selecionar as principais informações contidas em todos 
os documentos e criar um resumo com as mesmas \cite{Barzilay}.

\begin{figure}[!htb]
     \centering
      \includegraphics[width=11cm]{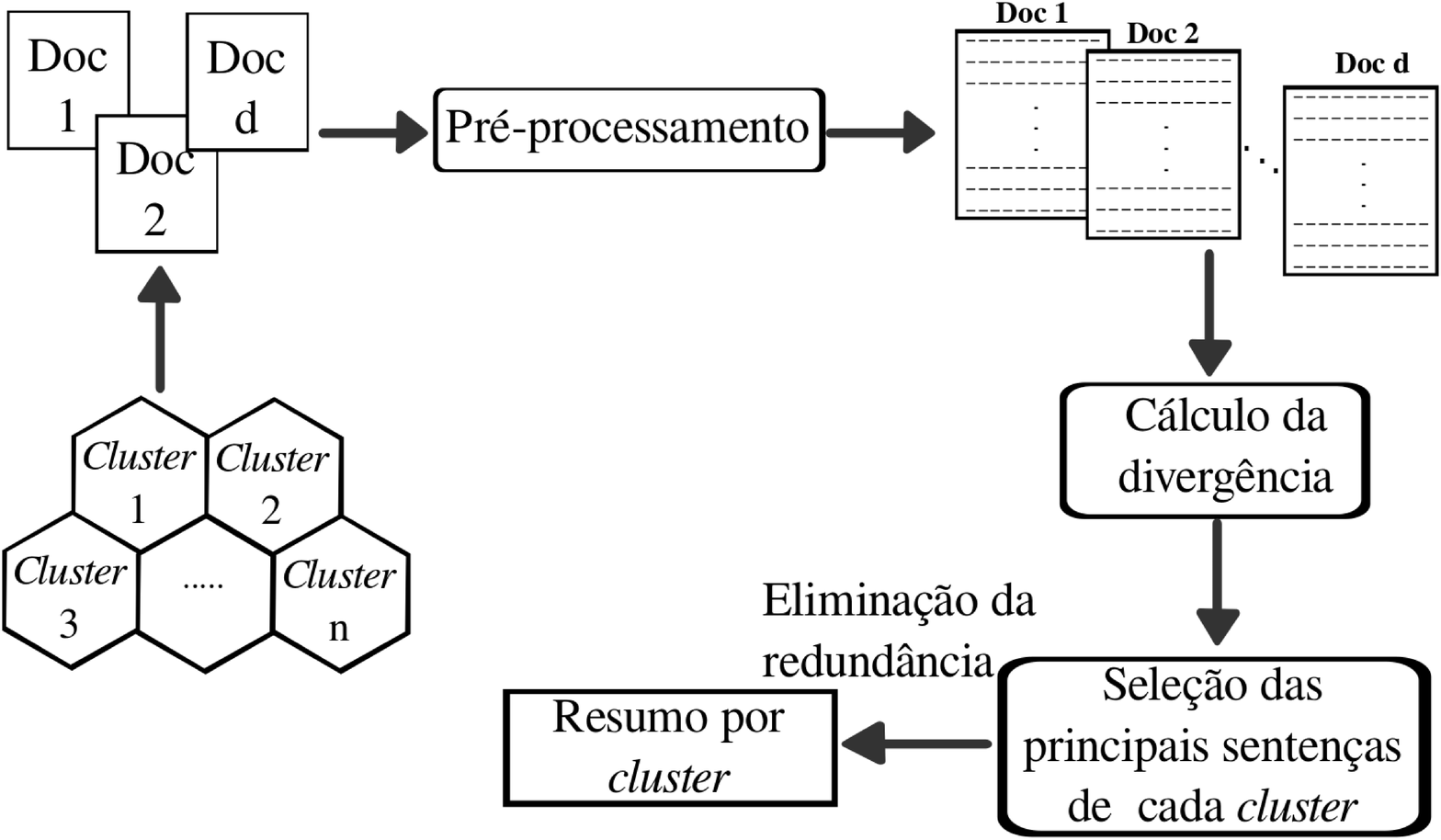}
     \caption{\label{fig:summatrix} Funcionamento do sistema SUMMatrix.}
\end{figure}

O sistema \ac{SUMMatrix} foi desenvolvido no intuito de analisar as sentenças de conjuntos de textos de conteúdo similar. O sistema cria um 
resumo por extração independente do idioma do \textit{cluster}, analisa as sentenças dos textos e calcula a divergência entre as mesmas 
a fim de mensurar sua relevância. Em seguida, o resumo é gerado através da concatenação das sentenças com maior pontuação. A figura 
\ref{fig:summatrix} descreve resumidamente o funcionamento do \ac{SUMMatrix}.

\subsection{Funcionamento}
\label{ssc:algoSummatrix}

O \ac{SUMMatrix} analisa um \textit{cluster} composto de textos relacionados a uma mesma notícia. Esses textos são similares e transmitem a 
mesma mensagem ou uma informação similar. O cálculo da similaridade utiliza unigramas e bigramas. O \ac{SUMMatrix} é iterativo pois analisa 
qualquer quantidade de documentos em um \textit{cluster}. Ele analisa os textos em pares com o intuito de identificar as sentenças 
similares.

O sistema \ac{SUMMatrix} calcula a divergência a partir da divergência \ac{JS} ou \ac{KL} bem como a similaridade do cosseno. Para calcular a divergência, considera-se os conjuntos $P$ e $Q$ que referem-se, respectivamente, ao conjunto de palavras da sentença do primeiro texto e da sentença do segundo texto. $Pr(P,w)$ é a probabilidade da palavra $w$ na sentença $P$ e $Pr(Q,w)$ da sentença $Q$. Além da divergência \ac{JS} ou \ac{KL}, o sistema 
avalia a similaridade do cosseno entre duas sentenças. Assim, a divergência entre duas sentenças $F1$ e $F2$, $Div(F1, F2)$, é calculada pela equação \ref{eq:sim} e seu valor varia entre $0$ e $\infty$. Portanto, quanto menor for o valor da divergência maior será a similaridade entre as sentenças.

\begin{equation}
 \label{eq:sim}
 Div(F1, F2) = [1 - cos(F1, F2)] + D_{JS}(F1||F2)
\end{equation}
%
%

O SUMMatrix analisa os dois primeiros textos, denominados $T1$ e $T2$, e verifica quais sentenças (em pares) são mais similares através do cálculo da divergência descrito na equação \ref{eq:sim}. O sistema cria uma matriz de divergência onde as linhas representam as sentenças do $T1$ e as colunas representam as sentenças do $T2$. O sistema seleciona, em seguida, o par de sentenças $(s_{T1}, s_{T2})$ com menor divergência e calcula a divergência entre ele e o \textit{cluster}. As sentenças selecionadas de cada par são consideradas mais relevantes e adicionadas ao \underline{Resumo Parcial}, o qual contém as principais sentenças dos textos considerados. Nesse momento, o sistema analisa o conjunto \underline{Resumo Parcial} e o texto seguinte, denominado $T3$. O sistema cria uma nova matriz de divergência entre as sentenças dos mesmos. O \ac{SUMMatrix} seleciona os pares de sentenças mais similares, calcula a divergência das sentenças em relação ao \textit{cluster} e acrescenta a sentença mais similar em cada par de sentenças no novo conjunto \underline{Resumo Parcial}. O sistema repete esse processo até que todos os textos tenham sido analisados. O conjunto \underline{Resumo Final} será composto pelas sentenças da última 
versão do conjunto \underline{Resumo Parcial}.

\begin{figure}[!htb]
     \centering
      \includegraphics[width=14cm]{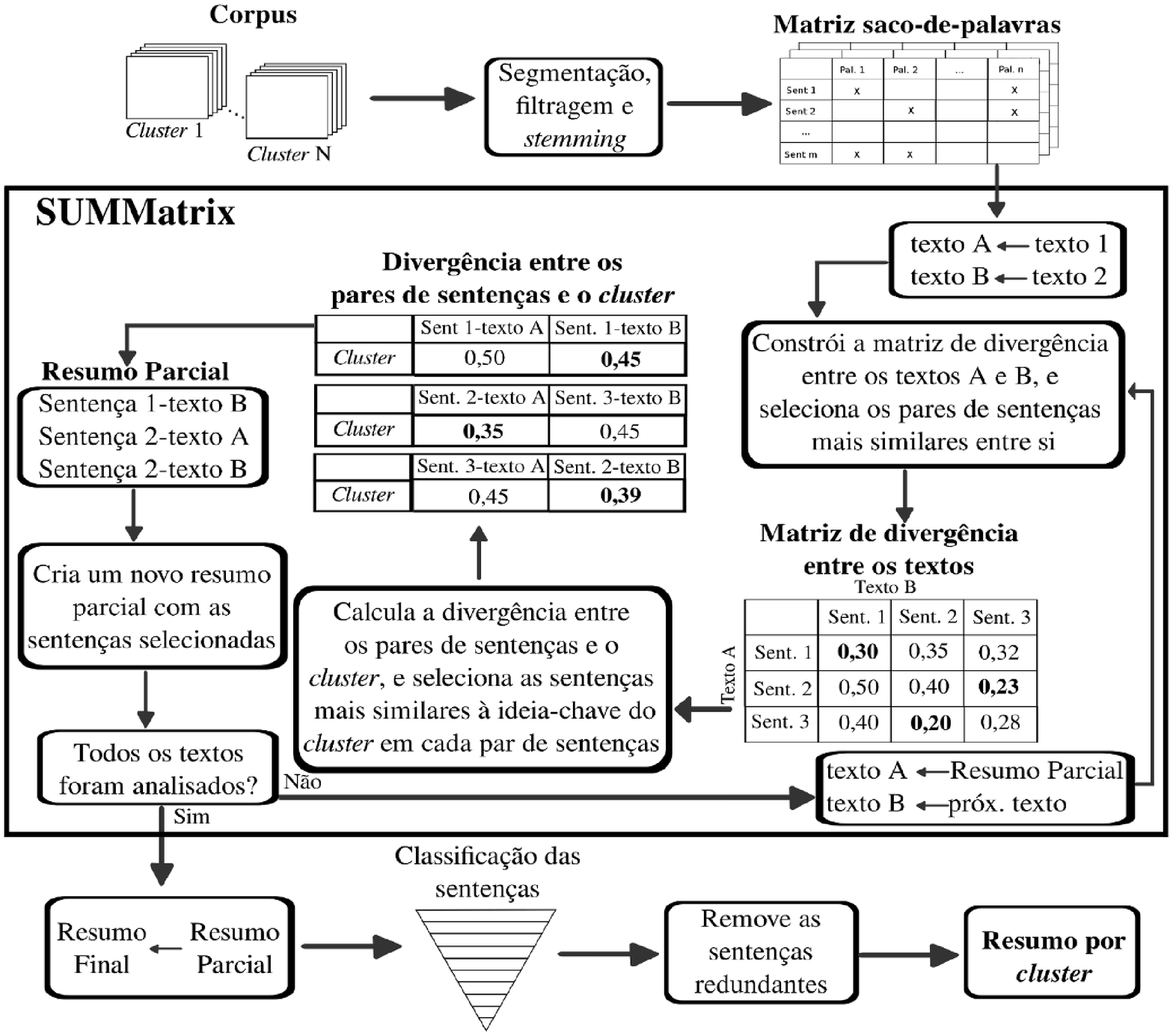}
     \caption{\label{fig:summatrix2} Sistema SUMMatrix.}
\end{figure}

O sistema analisa o conjunto \underline{Resumo Final} e cria o resumo sem sentenças redundantes a partir da similaridade de Dice \cite{dice} e da taxa de compressão definida pelo usuário. A figura \ref{fig:summatrix2} apresenta a arquitetura desse sistema de maneira estruturada.

\subsection{Exemplo}
\label{ssc:exemploSummatrix}

O conjunto de textos descrito na tabela \ref{tb:ex_summatrix} discorre sobre um mesmo acidente de avião ocorrido no Congo. O processo de 
sumarização inicia com o pré-tratamento do texto e com a criação da matriz de palavras. O SUMMatrix calcula a divergência 
entre pares de sentenças dos textos $T1$ e $T2$ e identifica aquelas com menor divergência (tabela \ref{tb:djs1}). 

As sentenças identificadas são comparadas com o \textit{cluster} completo para analisar a divergência entre elas. A sentença de cada par de frases 
mais similar ao \textit{cluster} é selecionada para constituir o \underline{Resumo Parcial} dos textos $T1$ e $T2$. A tabela 
\ref{tb:djs2} mostra a divergência entre as frases selecionadas e o \textit{cluster}. Assim, o resumo parcial da análise dos textos $T1$ e 
$T2$ é composto pelas frases: $T1-2$, $T1-4$, $T1-7$, $T2-1$ e $T2-3$.

\begin{tabela}[!htb]
	\begin{tabular}{|c|ccccccc|}
		\hline
		& $T2-1$ 	& $T2-2$ & $T2-3$& $T2-4$& $T2-5$& $T2-6$& $T2-7$\\\hline
		$T1-1$  &  \textbf{0,44} 	&  0,55	& 0,56	& 0,65	& 0,65	& 0,63	& 0,48	\\
		$T1-2$  &  0,51	&  0,66	& 0,51	& 0,52	& 0,43	& 0,64	& \textbf{0,40}	\\
		$T1-3$  &  0,67	&  0,65	& \textbf{0,35}	& 0,64	& 0,65	& 0,63	& 0,66	\\
		$T1-4$  &  0,55	&  0,44	& 0,62	& 0,60	& 0,68	& 0,67	& \textbf{0,43}	\\
		$T1-5$  &  0,59	&  0,64	& \textbf{0,44}	& 0,63	& 0,63	& 0,60	& 0,65	\\
		$T1-6$  & \textbf{0,38}	&  0,66	& 0,69	& 0,65	& 0,65	& 0,63	& 0,57	\\
		$T1-7$  &  0,49	&  0,65	& 0,68	& 0,64	& 0,64	& 0,63	& \textbf{0,47}	\\
		$T1-8$  &  \textbf{0,54}	&  0,65	& 0,61	& 0,63	& 0,64	& 0,61	& 0,55	\\\hline
	\end{tabular}
	\caption{Divergência entre as frases dos textos $T1$ e $T2$.}
	\label{tb:djs1}
\end{tabela}

\begin{tabela}[!htb]
\begin{tabular}{|c|cc||c|cc|}
\hline
	& $T1-1$&  $T2-1$						&   & $T1-5$	&$T2-3$	\\\hline
\textit{Cluster}	& 0,55	&  \textbf{0,39}		& \textit{Cluster}	&0,60	&\textbf{0,43}	\\\hline
	& $T1-2$	& $T2-7$					&  & $T1-6$	&$T2-1$	\\\hline
\textit{Cluster}	& \textbf{0,53}	&0,55		& \textit{Cluster}	&0,55	&\textbf{0,39}	\\\hline
	& $T1-3$	&$T2-3$						&  & $T1-7$	&$T2-7$	\\\hline
\textit{Cluster}	&0,57	&\textbf{0,43}	& \textit{Cluster}	&\textbf{0,54}	&0,55	\\\hline
	& $T1-4$	&$T2-7$						& & $T1-8$	&$T2-1$	\\\hline
\textit{Cluster}	&\textbf{0,34}	&0,55		& \textit{Cluster}	&0,58	&\textbf{0,39}	\\\hline
\end{tabular}
\caption{Divergência entre as sentenças selecionadas e o \textit{cluster}.}
\label{tb:djs2}
\end{tabela}

Em seguida, inicia-se a análise do \underline{Resumo Parcial} e o texto $T3$. É criada uma nova matriz de divergência entre eles (tabela 
\ref{tb:djs3}). O sistema calcula a divergência entre as frases e seleciona os pares de menor divergência. Novamente, analisa-se os pares de 
sentenças com menor divergência em relação ao \textit{cluster}. Verifica-se a divergência entre as frases e o \textit{cluster} completo e 
seleciona-se as frases de cada par que forem mais similares ao \textit{cluster} para compor o \underline{Resumo Final} do grupo de textos 
(tabela \ref{tb:djs4}). Finalmente, o \underline{Resumo Final} é composto pelas frases: $T1-4$, $T2-1$, $T2-3$, $T3-2$ e $T3-3$.

\begin{tabela}[!t]
\begin{tabular}{|c|ccccc|}
\hline
      & $T3-1$ 	& $T3-2$& $T3-3$& $T3-4$& $T3-5$	\\\hline
 $T1-2$  &  0,52	&  \textbf{0,50}	& 0,55	& 0,68	& 0,64	\\
 $T1-4$  &  0,63	&  0,55	& \textbf{0,15}	& 0,27	& 0,67	\\
 $T1-7$  &  0,50	&  0,68	& \textbf{0,27}	& 0,58	& 0,63	\\
 $T2-1$  &  \textbf{0,01}	&  0,64	& 0,51	& 0,68	& 0,66	\\
 $T2-3$  &  0,65	&  \textbf{0,02}	& 0,69	& 0,64	& 0,68	\\\hline
\end{tabular}
\caption{Divergência entre as frases do \underline{Resumo Parcial}\ e o texto $T3$.}
\label{tb:djs3}
\end{tabela}

\begin{tabela}[!htb]
\begin{tabular}{|c|cc|}
\hline
	& $T1-2$	&$T3-2$				\\\hline
\textit{Cluster}	&0,53	&\textbf{0,44}\\\hline
	& $T1-4$& $T3-3$				\\\hline
\textit{Cluster}	&\textbf{0,34}	&0,44\\\hline
	& $T1-7$	&$T3-3$				\\\hline
\textit{Cluster}	&0,54	&\textbf{0,44}\\\hline
 & $T2-1$	&$T3-1$	\\\hline
 \textit{Cluster}	&\textbf{0,39}	&0,40	\\\hline
 	& $T2-3$	&$T3-2$	\\\hline
 \textit{Cluster}	&\textbf{0,43}	&0,44	\\\hline
\end{tabular}
\caption{Divergência entre as sentenças selecionadas e o \textit{cluster}.}
\label{tb:djs4}
\end{tabela}

\begin{tabela}[!t]
\begin{tabular}{|p{0.9\columnwidth}|}
\hline
 {\footnotesize \textbf{Texto T1:} }\\
 {\footnotesize T1-1 Ao menos 17 pessoas morreram após a queda de um avião de passageiros na República Democrática do Congo.}\\
 {\footnotesize T1-2 Segundo uma porta-voz da ONU, o avião, de fabricação russa, estava tentando aterrissar no aeroporto de Bukavu em meio a 
uma tempestade.}\\ 
 {\footnotesize T1-3 A aeronave se chocou com uma montanha e caiu, em chamas, sobre uma floresta a 15 quilômetros  de distância da pista do 
aeroporto.}\\ 
 {\footnotesize T1-4 Acidentes aéreos são frequentes no Congo, onde 51 companhias privadas operam com aviões antigos principalmente 
fabricados na antiga União Soviética.}\\
 {\footnotesize T1-5 O avião acidentado, operado pela Air Traset, levava 14 passageiros e três tripulantes.}\\ 
 {\footnotesize T1-6 Ele havia saído da cidade mineira de Lugushwa em direção a Bukavu, numa  distância de 130 quilômetros.}\\ 
 {\footnotesize T1-7 Aviões são usados extensivamente para transporte na República Democrática do Congo, um vasto país no qual há poucas 
estradas pavimentadas.}\\ 
 {\footnotesize T1-8 Apenas uma manteve a permissão.}\\ \hline
 {\footnotesize \textbf{Texto T2:} }\\
 {\footnotesize T2-1 Um acidente aéreo na localidade de Bukavu, no leste da República Democrática do Congo (RDC), matou 17 pessoas na 
quinta-feira à tarde, informou nesta sexta-feira um porta-voz das Nações Unidas.}\\
 {\footnotesize T2-2 As vítimas do acidente foram 14 passageiros e três membros da tripulação.}\\
 {\footnotesize T2-3 Todos morreram quando o avião, prejudicado pelo mau tempo, não conseguiu chegar à pista de aterrissagem e caiu numa 
floresta a 15 quilômetros do aeroporto de Bukavu.}\\
 {\footnotesize T2-4 Segundo fontes aeroportuárias, os membros da tripulação eram de nacionalidade russa.}\\
 {\footnotesize T2-5 O avião explodiu e se incendiou, acrescentou o porta-voz da ONU em Kinshasa, Jean-Tobias Okala.}\\
 {\footnotesize T2-6 ``Não houve sobreviventes'', disse Okala.}\\
 {\footnotesize T2-7 O porta-voz informou que o avião, um Soviet Antonov-28 de fabricação ucraniana e propriedade de uma companhia 
congolesa, a Trasept Congo, também levava uma carga de minerais. }\\ \hline
 {\footnotesize \textbf{Texto T3:} }\\
 {\footnotesize T3-1 Um acidente aéreo na localidade de Bukavu, no leste da República Democrática do Congo, matou 17 pessoas na 
quinta-feira 
à tarde, informou hoje um porta-voz das Nações Unidas.}\\
 {\footnotesize T3-2 As vítimas do acidente foram 14 passageiros e três membros da tripulação.}\\
 {\footnotesize T3-3 Todos morreram quando o avião, prejudicado pelo mau tempo, não conseguiu chegar à pista de aterrissagem e caiu numa 
floresta a 15 Km do aeroporto de Bukavu.}\\
 {\footnotesize T3-4 O avião explodiu e se incendiou, acrescentou o porta-voz da ONU em Kinshasa, Jean-Tobias Okala.}\\
 {\footnotesize T3-5 ``Não houve sobreviventes'', disse Okala.} \\ \hline
\end{tabular}
\caption{\textit{Cluster} com 3 textos de diferentes jornais relatando um mesmo acidente no Congo (corpus CSTNews \cite{cstnews}).}
\label{tb:ex_summatrix}
\end{tabela}
\chapter{Corpus e métodos de avaliação}
\label{ch:cma}

Este capítulo descreve os corpus utilizados nos testes descritos no capítulo \ref{ch:avalia}. Serão apresentadas suas características e os 
métodos utilizados para  avaliar a qualidade dos sistemas desenvolvidos e dos referenciados na literatura. A seção \ref{sc:corpus} descreve 
os corpus e a seção \ref{sc:ava} descreve as métricas para avaliar a qualidade dos resumos.

\section{Corpus}
\label{sc:corpus}

O corpus é um conjunto de textos com características específicas. As subseções seguintes descrevem os corpus utilizados e suas características.

\subsection{Corpus multi-idioma}
\label{ssc:mi}

O corpus multi-idioma é composto por textos de jornais e revistas científicas em inglês, francês e espanhol. Além dessas características, os 
resumos dos textos selecionados estão disponíveis na literatura para outros sistemas sumarizadores \cite{Cor1,Ene2}. O corpus utilizado é 
composto por 13 textos de diferentes tamanhos e assuntos.

\subsection{CSTNews}
\label{ssc:cstnews}

O corpus CSTNews\footnote{Corpus CSTNews: http://www.icmc.usp.br/$\sim$taspardo/sucinto/cstnews.html} é composto por textos jornalísticos
em Português. Esse corpus possui notícias sobre política, esportes, acidentes, mundo, entre outros. Ele contém 50 
\textit{clusters}, cada qual possuindo 2 ou 3 textos relacionados ao mesmo acontecimento. Os textos são provenientes de diferentes 
agências de notícias online populares no país: Folha de São Paulo, Estadão, O Globo, Jornal do Brasil e Gazeta do Povo. Os textos foram 
selecionados durante os meses de agosto e setembro de 2007 \cite{cstnews}. O CSTNews possui 40.839 palavras e 258.818 caracteres.

\subsection{RPM}
\label{ssc:rpm}

O corpus RPM\footnote{Corpus RPM: http://lia.univ-avignon.fr/fileadmin/documents/rpm2/rpm2\_resumes\_fr.html} é composto por 40 
\textit{clusters} em Francês. O corpus possui 20 temáticas e cada uma delas possui 2 \textit{clusters}. Ele é composto por textos de 
assuntos similares que foram coletados em momentos distintos. Esse corpus possui resumos elaborados por profissionais em cada 
\textit{cluster}. O RPM 
possui 143.881 palavras e 905.348 caracteres.

\subsection{DECODA}
\label{ssc:decoda}

O corpus DECODA é composto de conversas em francês entre clientes e agentes que foram registradas em 2009 em um \textit{call center} da 
autoridade de transporte público de Paris (tabela \ref{tb:decoda}) \cite{BECHET}. Os tópicos das conversas consistem em pedidos de 
itinerários, cronogramas, perdidos e achados e reclamações. Os diálogos foram gravados em condições espontâneas e focados no objetivo do 
autor da chamada. Essas gravações são um grande desafio para o reconhecimento automático da fala devido às condições acústicas difíceis, 
tais como ligações de telefones celulares diretamente do metrô.

\begin{tabela}[!htb]
\centering
\begin{tabular}{|l|c|}
\hline
\ \ \ \ \ \ \ \ \ \ \ \ \ Estatísticas	& \textbf{FR} \\ \hline
Conversas		& 100 \\ 
Falas			& 7.905 \\ 
Palavras		& 42.130 \\ 
Tamanho médio		& 421,3 \\ 
Tamanho léxico		& 2.995 \\ 
Número de resumos	& 212 \\ 
Tamanho médio do resumo	& 23	\\ \hline
\end{tabular}
\caption{Estatística do corpus DECODA.}
\label{tb:decoda}
\end{tabela}

O corpus é composto por 1.513 conversas (cerca de 70 horas de conversação). 1.000 conversas foram distribuídas sem sinopses para a formação 
do sistema sem supervisão e 50 outras foram distribuídas com várias sinopses para formar até cinco anotadores. O conjunto de testes consiste 
de 47 conversas traduzidas manualmente e suas correspondentes sinopses, e 53 conversas traduzidas automaticamente assim como suas 
correspondentes sinopses \cite{benoit}.

\section{Métodos de avaliação}
\label{sc:ava}

Os capítulos anteriores descreveram diferentes metodologias  para criar resumos, seja por resumidores profissionais ou por algum tipo 
de sumarizador automático. No entanto, é fundamental analisar a qualidade desses resumos. As subseções seguintes descrevem as principais 
métricas e sistemas encontrados na literatura para avaliar a informatividade dos resumos produzidos.

\subsection{Métricas}
\label{ssc:met}

Uma forma de verificar a qualidade do resumo é viabilizada através da análise de 3 métricas (precisão, cobertura e $medida$-$f$), que 
representam a informatividade do resumo criado. Essas métricas possibilitam identificar o conteúdo do resumo e verificar se o assunto do 
mesmo refere-se ao conteúdo geral do texto original ou se ele aborda somente alguns dados aleatórios. 

A avaliação manual considera diferentes características no texto para a compreensão e para a avaliação do resumo como gramaticalidade, 
informatividade e coerência. A gramaticalidade é a qualidade gramatical de um texto, ou seja, ela identifica se uma frase esta construída 
corretamente. A informatividade representa a porcentagem das informações principais transmitidas no texto. A coerência é a ligação ou 
conexão entre os fatos e/ou as ideias de uma história. Entretanto, algumas dessas avaliações dependem do conhecimento e da opinião, entre 
outros fatores, dos resumidores profissionais. Por isso, a avaliação de resumos pode variar entre diferentes profissionais. Geralmente 
pode-se obtê-la através da análise fornecida por diferentes avaliadores, o que requer tempo e dinheiro. O avaliador lê o texto e seleciona 
suas principais informações. Após a leitura dos resumos fornecidos, ele verifica se o texto é coerente e conciso com relação ao tema 
discutido no texto original. Outro ponto analisado é a informatividade do texto.

Outra forma de avaliação possível feita pelos avaliadores consiste na criação de seus próprios resumos e comparação com os resumos 
candidatos (produzidos automaticamente). Geralmente, utiliza-se as métricas precisão e cobertura para avaliar a qualidade  dos resumos 
assim produzidos. A primeira calcula a fração de sentenças do resumo que são relevantes para o texto. A cobertura avalia a fração de sentenças relevantes do texto que estão presentes no resumo. Uma forma de unir essas duas métricas é dada pela $medida$-$f$ (equação \ref{eq:fscore}).

\begin{equation}
\label{eq:fscore}
 medida\textrm{-}f = 2 \times \frac{precis\widetilde{a}o \times cobertura}{precis\widetilde{a}o + cobertura}
\end{equation}


\subsection{ROUGE}
\label{ssc:rouge}

O sistema \ac{ROUGE} é um avaliador automático de resumos desenvolvido por \cite{rouge} baseado no sistema BLUE \cite{blue}. O
\ac{ROUGE} utiliza resumos de profissionais como referência para avaliar os demais resumos. Ele faz uso de diferentes métricas para 
determinar a similaridade entre os resumos dos profissionais e os resumos candidatos. Serão consideradas apenas as métricas ROUGE-N e 
ROUGE-SU nesta dissertação. A ROUGE-N avalia a co-ocorrência de $n$-gramas. Em outras palavras, ela é a recuperação de $n$-gramas entre
um sumário candidato e um conjunto de resumos de referências ($Sum_{Ref}$). Ela é calculada através da equação \ref{eq:rouge}.

\begin{equation}
 \label{eq:rouge}
 ROUGE\textrm{-}N = \frac{\sum_{S \in Sum_{Ref}} \sum_{n\textrm{-gramas} \in S} Count_{match}(n\textrm{-gramas})}{\sum_{S \in Sum_{Ref}} \sum_{n\textrm{-gramas} \in S} 
Count(n\textrm{-gramas})}
\end{equation}

A ROUGE-N será utilizada para avaliar unigramas (ROUGE-1) e bigramas (ROUGE-2). A ROUGE-SU é uma extensão da ROUGE-S, que avalia também
quaisquer pares de palavras na ordem das sentenças entre os resumos  (\textit{skip-bigrama} ou $skip2$). Dessa forma, a 
ROUGE-S calcula a precisão ($Pr$), a cobertura ($Cb$) e a $medida$-$f$, conforme as equações \ref{eq:pr}, \ref{eq:cb} e \ref{eq:f}, 
respectivamente.

\begin{equation}
\label{eq:pr}
 Pr_{skip2} = \frac{SKIP2(P,Q)}{Comb(m,2)}
\end{equation}

\begin{equation}
\label{eq:cb}
 Cb_{skip2} = \frac{SKIP2(P,Q)}{Comb(n,2)}
\end{equation}

\begin{equation}
\label{eq:f}
 F_{skip2} = \frac{(1+\beta^2) \times Cb_{skip2} \times Pr_{skip2}}{Cb_{skip2}+\beta^2 \times Pr_{skip2}}
\end{equation}

\noindent $SKIP2(P,Q)$ é a quantidade de \textit{skip-gramas} iguais entre $P$ e $Q$, $\beta$ controla a relevância do $P_{skip2}$ e 
$R_{skip2}$, e $Comb$ é a função de combinações possíveis.

Entretanto, a ROUGE-S não valoriza as sentenças dos resumos em que não há uma co-ocorrência das palavras do resumo candidato com relação aos 
sumários de referência. Nesse caso, a ROUGE-SU adiciona a contagem de unigramas permitindo às palavras isoladas uma certa relevância no 
processo de avaliação.

\subsection{FRESA}
\label{ssc:fresa}

Nem sempre tem-se recursos financeiros e tempo para a obtenção de resumos de referência para os textos considerados. Nesse caso, o sistema 
\ac{FRESA} apresenta-se como uma boa opção. FRESA é um avaliador de resumos que analisa a qualidade dos mesmos sem necessitar de resumos 
de referência \cite{fresa}. Ele avalia a qualidade do texto a partir das distribuições de probabilidades $P$ e $Q$,  que 
representam a distribuição do resumo $Res$ e o documento $Doc$, respectivamente. Ele calcula a $D_{JS}$ para unigramas, bigramas, 
bigramas-SU4 e suas combinações para $P$ e $Q$ (equação \ref{DJS}). O \ac{FRESA} utiliza as atribuições explicitas na equação \ref{eq:fre}, 
onde $F_w^{Doc}$ é a frequência da palavra $w$ no documento, $F_w^{Res}$ é a frequência da palavra $w$ no resumo, $N_P$ é o
número de palavras do resumo e do documento, $\beta$ é $1,5 \times voc$, $voc$ é o vocabulário do resumo e do documento  e $\gamma$ é 
parâmetro de suavização. Dessa forma, o \ac{FRESA} avalia a precisão, a cobertura e a $medida$-$f$ para cada resumo de acordo com o texto 
original.

\begin{equation}
\label{eq:fre}
 P_w = \frac{F_w^{Doc}}{|Doc|}; Q_w = \left\{
\begin{array}{cc}
\frac{F_w^{Res}}{|Res|}, & \textrm{se w }\in Res\\
\frac{F_w^{Doc} + \gamma}{N_P + \gamma \times \beta}, & \textrm{caso contrário}\\
\end{array}
\right\}
\end{equation}
\chapter{Avaliação Experimental}
\label{ch:avalia}

Os testes foram realizados em um computador com processador i5@2.6 GHz e 4 GB de memória RAM no sistema operacional GNU/Linux Ubuntu
15.04 64-bits. Os algoritmos que compõem os sistemas LIA-RAG, RAG, SASI e SUMMatrix foram implementados na linguagem de programação
Perl. Foram desenvolvidos igualmente dois sistemas \textit{baselines} (resumos produzidos aleatoriamente, \textit{base-rand}, e contendo as 
primeiras sentenças, \textit{base-first}, dos textos) para auxiliar na avaliação dos sistemas. As seções a seguir descrevem os resultados 
obtidos dos sistemas propostos nesta dissertação para cada corpus analisado, sendo que alguns sistemas foram testados com um corpus 
específico tendo em vista as particularidades do problema considerado (áudio ou texto, mono ou multi-documento, etc).

\section{Avaliação do sistema SASI (Corpus multi-idioma)}
\label{sc:ami}

O corpus multi-idioma foi utilizado para avaliar o desempenho do sistema \ac{SASI}. Foram considerados na comparação dos resultados os 
valores fornecidos pelos sistemas sumarizadores Cortex, Enertex e REG. O REG modela o documento como um grafo e atribui uma ponderação às 
arestas para gerar o resumo do texto \cite{REG}.

As similaridades foram analisadas entre: duas frases ($D_{Frase}$); uma frase e o título ($D_{Titulo}$); e uma frase e o documento 
($D_{Texto}$). A tabela \ref{tb:sim} descreve o nível de similaridade relacionado ao valor da divergência JS e aos valores selecionados 
(parametrização) nos experimentos aqui descritos. Analisou-se o $\gamma$ variando entre $0,01$ e $0,20$ e, a partir dos resultados obtidos, 
selecionou-se $\gamma = 0,1$.

\begin{tabela}[htbp]
\centering
\begin{tabular}{|c|c|c|c|c|c|c|c|c|c|}
\hline
\multirow{2}{*}{Parâmetros} & \multicolumn{ 4}{c|}{Nível de similaridade} \\ \cline{2-5}
 & Fraca & Média & Forte & \multicolumn{1}{|l|}{\textbf{Selecionado}} \\ \hline
$D_{Frase}$ & $JS > 0,75$ & $0,25 < JS < 0,75$ & $JS < 0,25$ & \textbf{0,32} \\ \hline
$D_{Titulo}$ & $JS > 0,8$ & $0,4 < JS < 0,8$ & $JS < 0,4$ & \textbf{0,6} \\ \hline
$D_{Texto}$ & $JS > 0,95$ & $0,65 < JS < 0,95$ & $JS < 0,65$ & \textbf{0,9} \\ \hline
\end{tabular}
\caption{Nível de similaridade relacionado à divergência JS} \label{tb:sim}
\end{tabela}

A tabela \ref{tabela2} mostra 4 textos\footnote{Textos disponíveis em: http://www.lia.univ-avignon.fr/chercheurs/torres/recherche/cortex e 
http://en.wikipedia.org/wiki/Quebec\_sovereignty\_movement;Monica\_Lewinsky;Nazca\_lines} extraídos do corpus multi-idioma e os resumos 
produzidos pelo SASI, ressaltando a quantidade de frases presentes nos resumos construídos automaticamente para textos de diferentes 
tamanhos e valores do parâmetro $D_{Frase}$. Pode-se observar, a partir dos dados dessa tabela, que quanto maior o valor da $D_{JS}$ entre 
frases menor será o resumo (número de frases) fornecido pelo sistema. Portanto, esse valor dependerá do tamanho do resumo que se deseja 
obter. Deve-se considerar que quanto menor o resumo mais informações serão omitidas e a informatividade poderá ser reduzida.

\begin{tabela}[htbp]
\centering
\begin{tabular}{|c|c|c|c|c|c|c|}
\hline 
\multirow{4}{*}{Textos} & \multicolumn{ 6}{c|}{Quantidade de Frases} \\ \cline{2-7}
& \multirow{3}{*}{Texto Original (frases)} & \multicolumn{ 5}{c|}{Resumos}\\ \cline{3-7}
&  &  \multicolumn{ 5}{c|}{Valor da $D_{Frase}$} \\ \cline{3-7}
&  & 0,30 & \textbf{0,32} & 0,33 & 0,35 & 0,40 \\ \hline
\textit{Mars} & 12 & 6 & \textbf{4} & 4 & 4 & 2 \\ \hline
\textit{Puces} & 31 & 14 & \textbf{9} & 8 & 5 & 3 \\ \hline
\textit{Lewinsky} & 32 & 13 & \textbf{7} & 7 & 5 & 3 \\ \hline
\textit{Quebec} & 45 & 18 & \textbf{15} & 14 & 12 & 6 \\ \hline
\end{tabular}
\caption{Tamanho dos resumos obtidos para um conjunto de valores da $D_{JS}$.} \label{tabela2}
\end{tabela}

O gráfico \ref{desempenho} descreve o desempenho obtido em segundos pelo sistema SASI na produção de resumos a partir dos textos da tabela 
\ref{tabela2}, com a quantidade de palavras variando entre 11 e 214. Os sistemas obtiveram os seguintes tempos de execução total para o corpus 
analisado: Cortex, 58,27 segundos; Enertex, 56,80 segundos; REG, 55,70 segundos; e SASI, 38,78 segundos.

\begin{grafico}[!htb]
     \centering
      \includegraphics[width=11cm]{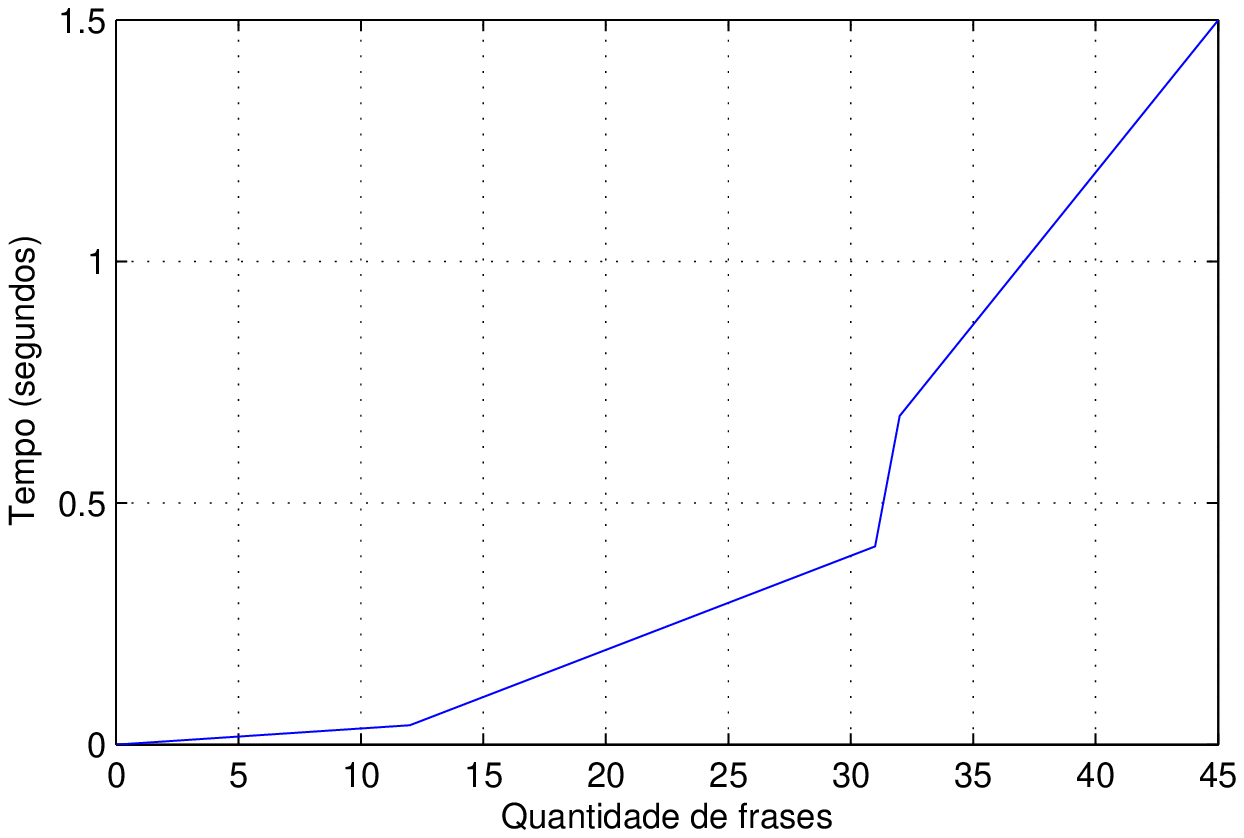}
     \caption{Desempenho do sistema SASI para o corpus multi-idioma.}
     \label{desempenho}
\end{grafico}

Analisou-se, ainda, a qualidade dos resumos através da informatividade (tabela \ref{tab3}) calculada através da taxa de acerto das 
frases fornecidas pelos sistemas sumarizadores em relação aos resumos dos profissionais. O SASI obteve o melhor resultado ao resumir o texto 
\textit{Quebec} e um desempenho semelhante aos outros sistemas com relação ao texto \textit{Puces}. No caso dos textos \textit{Lewinsky} e 
\textit{Mars}, o SASI teve uma taxa de acerto inferior aos demais sistemas pois os SIV calculados não corresponderam aos SIM dos grafos 
analisados, o que implica que a ponderação das frases não foi suficientemente significativa para selecionar outros elementos importantes dos 
textos. Nos demais textos do corpus, os resumos tiveram uma taxa de informatividade semelhante aos resultados obtidos com a sumarização de 
\textit{Puces}. Assim, apesar do SASI ter obtido um desempenho inferior em 69\% com relação aos textos analisados, a informatividade dos 
documentos foi preservada, o que é extremamente importante na construção de um resumo de forma automática. 

\begin{tabela}[htbp]
\centering
\begin{tabular}{|c|c|c|c|c}
\hline
\multirow{2}{*}{Textos} & \multicolumn{ 4}{c|}{Sistemas} \\ \cline{2-5}
 & SASI & Cortex & Enertex & \multicolumn{ 1}{c|}{REG} \\ \hline
\textit{Mars} & 67\% & 100\% & 100\% & \multicolumn{ 1}{c|}{100\%} \\ \hline
\textit{Puces} & 63\% & 75\% & 63\% & \multicolumn{ 1}{c|}{63\%} \\ \hline
\textit{Lewinsky} & 43\% & 57\% & 57\% & \multicolumn{ 1}{c|}{57\%} \\ \hline
\textit{Quebec} & 73\% & 55\% & 46\% & \multicolumn{ 1}{c|}{55\%} \\ \hline
\end{tabular}
\caption{Análise da precisão dos resumos gerados automaticamente.} 
\label{tab3}
\end{tabela}

O SASI é uma ferramenta que tem por base uma heurística simples e que se apoia em cálculos estatísticos menos complexos do que outros sistemas utilizados no domínio do PLN no âmbito da sumarização automática de documentos. A integração de novas regras sintáticas e semânticas para pré-tratamento dos textos proverá melhorias no desempenho do SASI. Além disso, com base nos testes realizados, o desenvolvimento de um método mais refinado para cálculo do SIV que assegure resultados mais próximos de um SIM priorizando sempre um dos vértices de maior grau em cada grupo, impactará de maneira positiva na qualidade dos sumários produzidos, visto que será escolhido um numero maior de ``frases importantes'' do texto e com conteúdo distinto, o que evitará que frases relevantes sejam descartadas no processo.

\section{Avaliação do sistema SUMMatrix (Corpus CSTNews)}
\label{sc:acstnews}

Os resumos dos profissionais produzidos para os textos do corpus CSTNews foram usados como referência para avaliar a qualidade dos sistemas 
Artex, Cortex, Enertex, GistSumm, LexRank, \ac{SUMMatrix} e \textit{base-rand}.

Os sistemas avaliadores \ac{FRESA} e \ac{ROUGE} foram utilizados para analisar a qualidade dos resumos produzidos automaticamente. Os 
resumos produzidos foram de qualidade média e permitiram a compreensão do texto. A tabela \ref{tb:fresa} mostra 
os resultados dos sistemas com a avaliação do \ac{FRESA} usando o processo de \textit{stemming} e uma taxa de compressão de $20\%$. O 
sistema Cortex obteve os melhores resultados em todas as características. O \ac{SUMMatrix} obteve bons resultados e foi melhor que os 
sistemas \textit{baseline} (\textit{base-rand}), GistSumm e LexRank (gráfico \ref{fig:fresa}, os melhores sistemas estão posicionados à 
direita e no topo do gráfico.). 

\begin{tabela}[!htb]
\centering
\begin{tabular}{|c|ccc|}
\hline
 \textbf{Sistemas} 	& \textbf{FRESA-1}    		& \textbf{FRESA-2}    		& \textbf{FRESA-4}     	\\ \hline
 Artex 			& 0,70258    			& 0,64898    			& 0,64193\\ 
 \textit{baseline} 		& 0,66725    			& 0,60514    			& 0,59671\\ 
 Cortex			& \textbf{0,71277}    		& \textbf{0,65648}    		& \textbf{0,64892}\\ 
 Enertex		& 0,69969    			& 0,64833    			& 0,64121\\ 
 GistSumm		& 0,63873    			& 0,59095    			& 0,58714\\ 
 LexRank		& 0,66545    			& 0,60799    			& 0,60106\\ 
 \small SUMMatrix	& 0,68707    			& 0,63087    			& 0,62464\\ \hline
\end{tabular}
\caption{Experimentos com o CSTNews para resumos sem referências.}
\label{tb:fresa}
\end{tabela}

A tabela \ref{tb:rouge} mostra a avaliação do sistema \ac{ROUGE} usando as mesmas características usadas na avaliação FRESA. Nessa 
avaliação, o \ac{SUMMatrix} obteve os melhores resultados para todos os parâmetros do \ac{ROUGE}. O gráfico \ref{fig:rougeEvaluation} mostra 
a qualidade dos resumos descritos na tabela \ref{tb:rouge}, baseado nas métricas ROUGE-2 e ROUGE-SU. Nesse caso, \ac{SUMMatrix} foi o
melhor sistema seguido por GistSumm e LexRank. Os sistemas Artex, Cortex e Enertex tiveram desempenho similares.

\begin{grafico}[!htb]
     \centering
     \includegraphics[width=11cm]{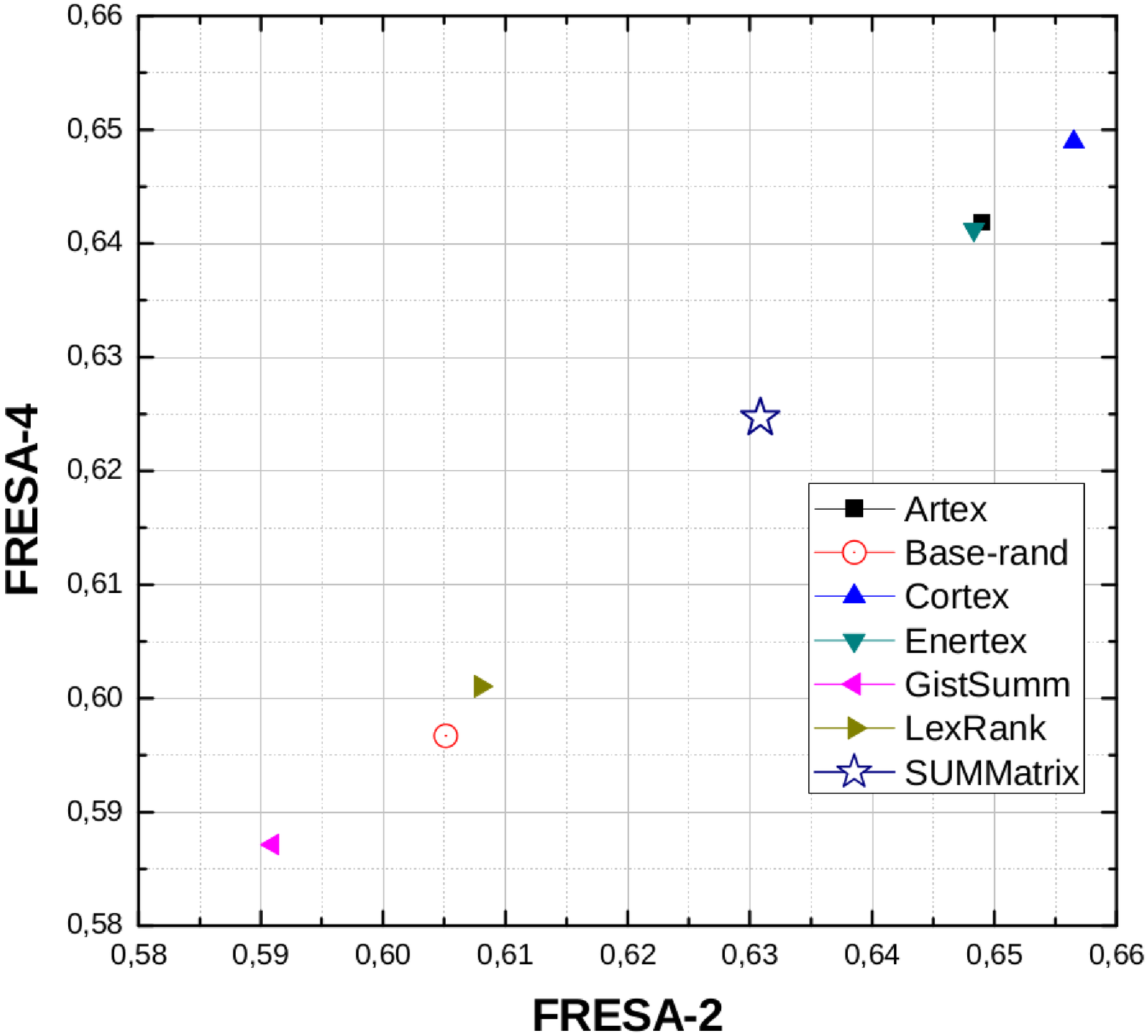}
     \caption{Avaliação FRESA dos sistemas usando CSTNews.}
     \label{fig:fresa}
\end{grafico}

\begin{tabela}[!htb]
\centering
\begin{tabular}{|c|ccc|}
\hline
 \textbf{Sistemas} 	& \small \textbf{ROUGE-1} 	& \small \textbf{ROUGE-2} 	& \small \textbf{ROUGE-SU} 	\\ \hline
 Artex 			& 0,43871 			& 0,23515 			& 0,16102\\ 
 \textit{base-rand} 		& 0,42877 			& 0,17859 			& 0,15773\\ 
 Cortex			& 0,44270 			& 0,23586 			& 0,16279\\ 
 Enertex		& 0,43688 			& 0,23853 			& 0,16099\\ 
 GistSumm		& 0,44629 			& 0,22395 			& 0,19043\\ 
 LexRank		& 0,46765 			& 0,21586 			& 0,18396\\ 
 \small SUMMatrix	& \textbf{0,47749} 		& \textbf{0,25011} 		& \textbf{0,19141}\\ \hline
\end{tabular}
\caption{Experimentos com o CSTNews usando resumos de profissionais.}
\label{tb:rouge}
\end{tabela}

O tempo de execução do \ac{SUMMatrix} foi maior que o dos outros sistemas (tabela \ref{tb:time}). Entretanto, ele obteve um excelente 
resultado quando avaliado com o ROUGE (teste mais relevante pois utiliza os resumos de referência) e bons resultados com o FRESA.

\begin{grafico}[!htb]
     \centering
     \includegraphics[width=11cm]{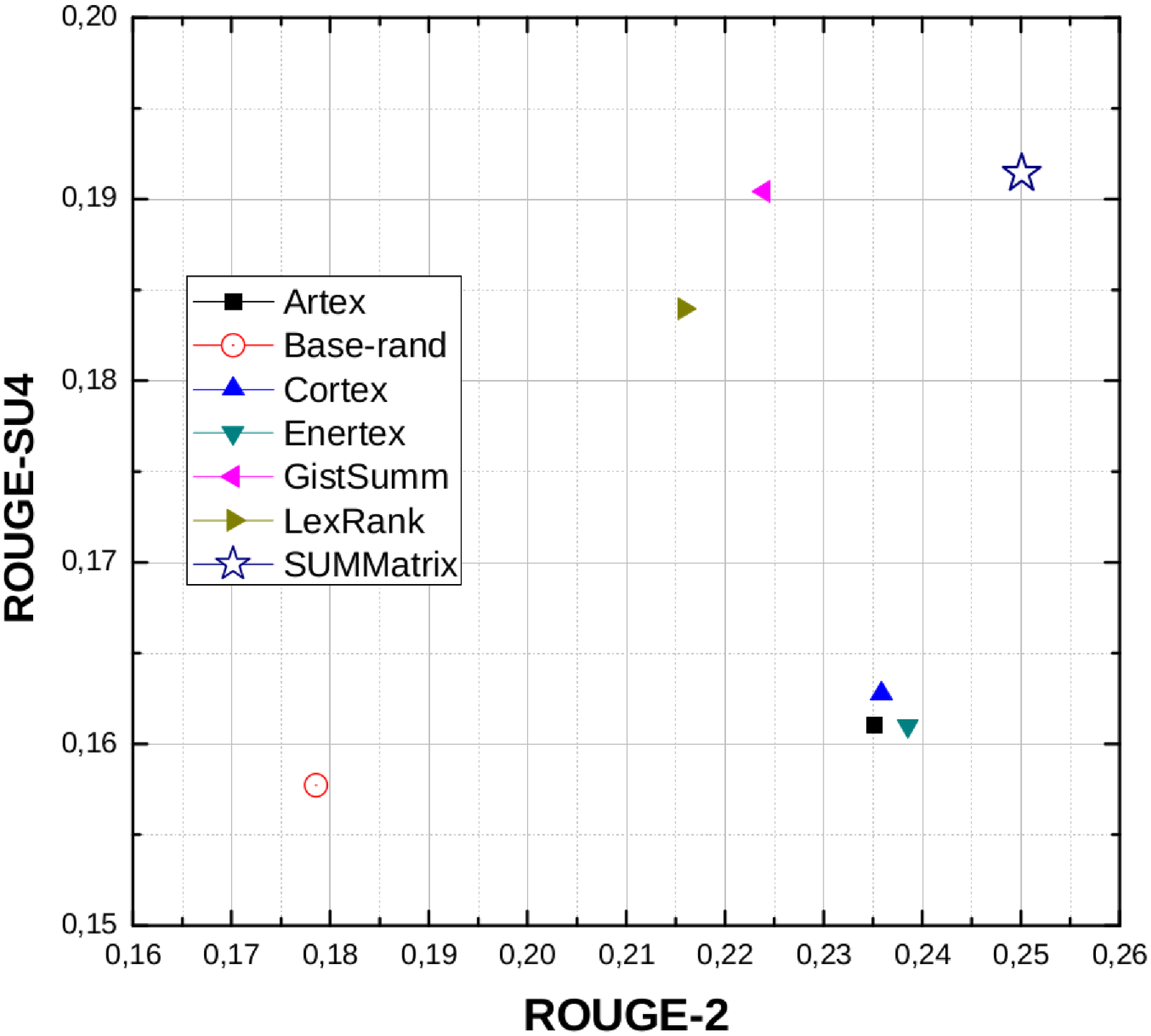}
     \caption{Avaliação ROUGE dos sistemas usando CSTNews.}
     \label{fig:rougeEvaluation}
\end{grafico}

Foram realizados, ainda, os mesmos testes com uma taxa de compressão de $30\%$. Nesse cenário, os resultados obtidos foram similares para o 
FRESA e para o ROUGE.

\begin{tabela}[h]
\centering
\begin{tabular}{|cccc|}
\hline
\small \textbf{Artex} 		& \small \textbf{base-rand} 	& \small \textbf{Cortex} 	& \small \textbf{Enertex}\\ \hline
 11,7s    			& \textbf{10,5s}		& 13,1s 			& 12,4s\\ \hline
 \small \textbf{GistSumm}    	& \small \textbf{LexRank}  	&  \multicolumn{2}{c|}{\small \textbf{SUMMatrix}} \\ \hline
 50,765s    			& 25,0s   			&  \multicolumn{2}{c|}{242,4s}  \\ \hline
\end{tabular}
\caption{\label{tb:time} Tempo de execução dos sistemas usando o corpus CSTNews.}
\end{tabela}

\section{Avaliação do sistema RAG (Corpus RPM)}
\label{sc:arpm}

O sistema RAG foi utilizado para avaliar todos os \textit{clusters} do corpus RPM e produzir automaticamente resumos com as principais 
informações do grupo analisado. Para esse corpus, analisou-se os resultados (resumos) produzidos pelos sistemas Artex, Cortex, Enertex, 
LexRank, RAG, \textit{base-first} e \textit{base-rand}.

Para avaliar a qualidade dos resumos, fez-se uso do sistema ROUGE (seção \ref{ssc:rouge}). A tabela \ref{tb:rouge} mostra os resultados 
dos sistemas com a avaliação ROUGE em que os resumos são compostos por 100 palavras para um \textit{cluster} (tabela \ref{tb:rouge1c}) e 
para dois \textit{clusters} (tabela \ref{tb:rouge2c}) de cada temática. Considerando os dois casos, o sistema Enertex obteve os melhores resultados para os parâmetros 
ROUGE-2 e ROUGE-SU. O sistema RAG obteve bons resultados e foi melhor que os sistemas Artex, \textit{base-first}, \textit{base-rand}, 
Cortex e LexRank.

\begin{tabela}[h]
\centering
\begin{tabular}{|c|ccc|}
\hline
 \textbf{Sistemas} 	& \textbf{ROUGE-1}	& \textbf{ROUGE-2}    	& \textbf{ROUGE-SU}     	\\ \hline
 Artex 			& 0,36406		& 0,10953 		& 0,13532 \\ 
 \textit{base-first} 		& 0,33915		& 0,08126 		& 0,11443 \\ 
 \textit{base-rand} 		& 0,31885		& 0,05697 		& 0,09917 \\ 
 Cortex			& 0,36810		& 0,10605 		& 0,13407 \\ 
 Enertex		& 0,37697		& \textit{0,12057} 	& \textbf{0,14778} \\ 
 LexRank		& \textbf{0,38965}	& \textbf{0,12201} 	& 0,14424 \\ 
 RAG			& \textit{0,37852}	& 0,11795 		& \textit{0,14490} \\ \hline
\end{tabular}
\caption{\label{tb:rouge1c} Avaliação ROUGE dos sistemas utilizando um único \textit{cluster} do RPM.}
\end{tabela}

O gráfico \ref{fig:rougeRPMn} sumariza os dados com relação à qualidade de cada resumo baseado nas métricas ROUGE-2 e ROUGE-SU com relação aos resultados obtidos utilizando os dois \textit{clusters} do corpus RPM.

\begin{tabela}[h]
\centering
\begin{tabular}{|c|ccc|}
\hline
 \textbf{Sistemas} 	& \textbf{ROUGE-1}	& \textbf{ROUGE-2}    	& \textbf{ROUGE-SU}     	\\ \hline
 Artex 			& 0,36205		& 0,10420 		& 0,13280 \\ 
 \textit{base-first} 		& 0,33805		& 0,08098 		& 0,11592 \\ 
 \textit{base-rand} 		& 0,32138		& 0,06353 		& 0,10082 \\ 
 Cortex			& 0,36432		& 0,10342 		& 0,13279 \\ 
 Enertex		& \textit{0,36867}	& \textbf{0,11255} 	& \textbf{0,14112} \\ 
 LexRank		& \textbf{0,37139}	& 0,11107 		& 0,13654 \\ 
 RAG			& 0,36607		& \textit{0,11119}	& \textit{0,13864} \\ \hline
\end{tabular}
\caption{\label{tb:rouge2c} Avaliação ROUGE dos sistemas utilizando dois \textit{clusters} do RPM.}
\end{tabela}

\section{Avaliação dos sistemas LIA-RAG e RAG (Corpus DECODA)}
\label{sc:adecoda}

\textsl{``Multiling é uma iniciativa dirigida para sistemas de sumarização multi-lingue \textit{benchmarking}, para incentivar a pesquisa e 
alavancar o estado da arte na área''}\footnote{http://multiling.iit.demokritos.gr/pages/view/1517/multiling-2015-call-for-participation}. A 
iniciativa \textit{MultiLing} 2015 possuiu as seguintes tarefas: \textit{Multilingual Multi-document Summarization}, \textit{Multilingual 
Single-document Summarization}, \textit{Online Forum Summarization} e \textit{Call Centre Conversation Summarization} (CCCS). A tarefa 
piloto CCCS consiste em \textsl{``criar sistemas que analisem conversas de \textit{call centers} e gerem resumos textuais refletindo o 
motivo do cliente estar ligando, como o agente responde às questões, os passos para solucionar os problemas e o estado de resolução do 
problema"} \cite{benoit}.

\begin{grafico}[!htb]
     \centering
     \includegraphics[width=11cm]{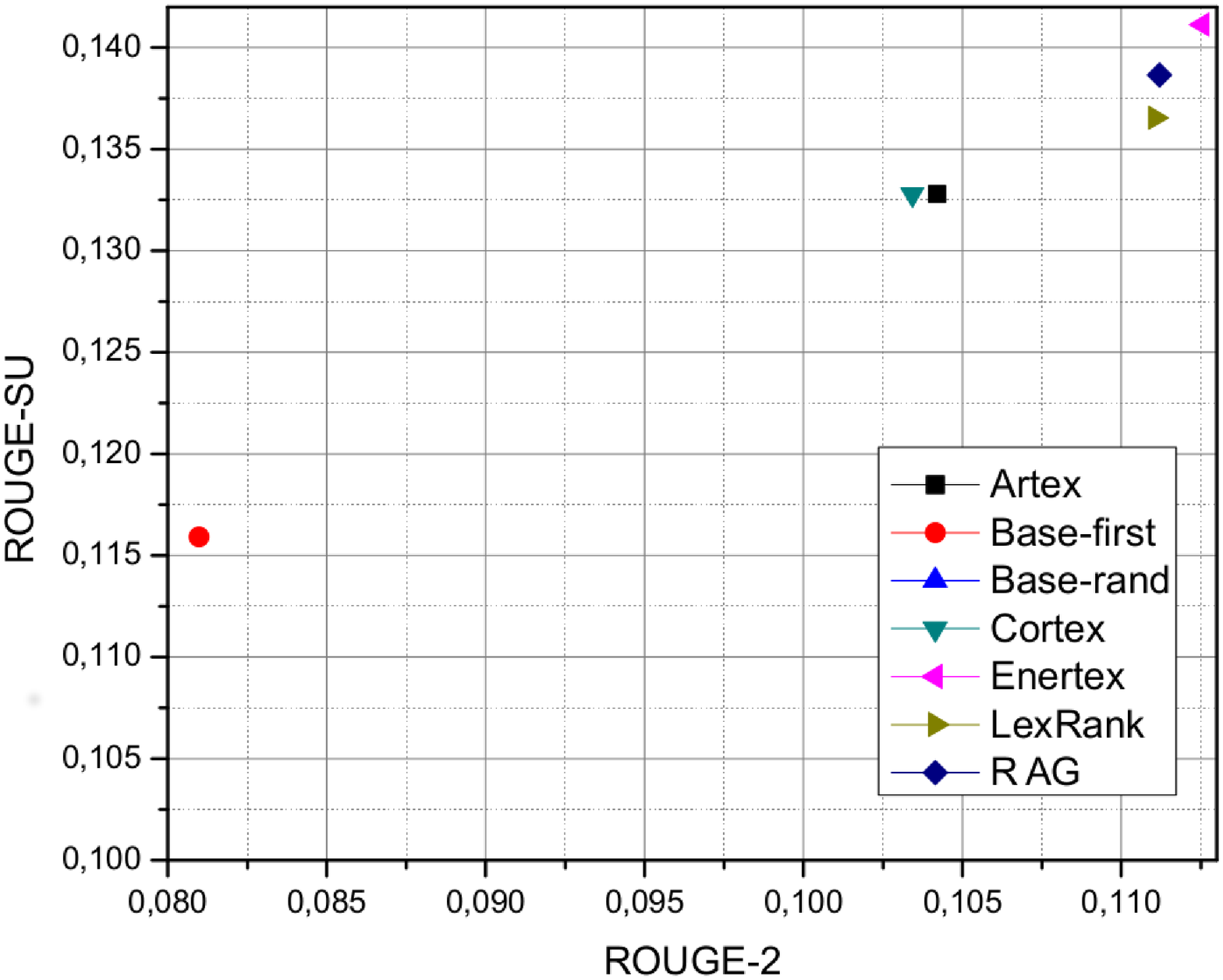}
     \caption{\label{fig:rougeRPMn} Avaliação ROUGE dos sistemas usando o corpus RPM.}
\end{grafico}

Foi utilizado o corpus Francês DECODA \cite{BECHET}. Os sistemas devem produzir resumos com a ideia principal das conversas. 
\textsl{``Os tópicos das conversas variam entre itinerário e pedidos de horários, perdidos e achados e reclamações ''}\cite{benoit}. Cada 
resumo tem 7\% do número de palavras de cada conversa transcrita. Comparou-se os sistemas LIA-RAG e RAG com os sistemas 
\textit{base-first}, \textit{base-rand} e os demais sistemas da competição (NTNU:1, NTNU:2 e NTNU:3). 

O \textit{Multiling} CCCS utilizou o sistema \ac{ROUGE}\footnote{Os parâmetros de execução: ROUGE 1.5.5 -a -l 10000 -n 4 -x -2 4 -u -c 95 
-r 1000 -f A -p 0.5 -t 0} para avaliar a qualidade dos sistemas. A tabela \ref{tb:rouge_training} exibe os resultados obtidos usando os 
sistemas citados anteriormente e o corpus de treinamento\footnote{O corpus de treinamento é utilizado para configurar e testar o sistema 
para as características específicas do corpus de avaliação.}. Ambas as versões do RAG foram melhores que os demais sistemas da competição.  
O pós-processamento do LIA-RAG permitiu melhorar os resultados do RAG por meio da redução dos erros e produzir um resumo mais informativo
e conciso.

\begin{tabela}[h]
\centering
\begin{tabular}{|c|ccc|}
\hline
 \textbf{Sistemas} 	& \small \textbf{ROUGE-1}    	& \small \textbf{ROUGE-2}    	& \small \textbf{ROUGE-4}     	\\ \hline
 \textbf{LIA-RAG:1}	& \textbf{0,1893}  		& \textbf{0,0628}  		& \textbf{0,0683} \\ 
 RAG		& 0,1833  		& 0,0614  		& 0,0654 \\ 
 \textit{base-first} 		& 0,1578  			& 0,0556  			& 0,0583 \\ 
 \textit{base-rand} 		& 0,1170  			& 0,0310  			& 0,0371 \\ 
\hline
\end{tabular}
\caption{\label{tb:rouge_training} Avaliação dos sistemas usando o corpus de treinamento DECODA.}
\end{tabela}

O corpus Francês de teste tem 100 conversas transcritas com 42.130 palavras e 212 resumos. A performance oficial ROUGE-2 para os sistemas 
participantes do CCCS é exibida na tabela \ref{tb:rouge_test} \cite{benoit}. O sistema LIA-RAG obteve os melhores resultados.

\begin{tabela}[h]
\centering
\begin{tabular}{|c|c|}
\hline
 \textbf{Sistemas} 	&  \textbf{ROUGE-2} 	\\ \hline
 \textbf{LIA-RAG:1}	&  \textbf{0,037} \\
 NTNU:1 		&  0,035 \\ 
 NTNU:3			&  0,034 \\ 
 NTNU:2 		&  0,027 \\ \hline
\end{tabular}
\caption{\label{tb:rouge_test} Avaliação dos sistemas usando o corpus de teste DECODA.}
\end{tabela}

\section{Análise geral}
\label{sc:analise}

Os 4 sistemas (LIA-RAG, RAG, SASI e SUMMatrix) desenvolvidos nesta dissertação possuem diferentes abordagens e são mais indicados para 
certos tipos de corpus e aplicações. Cada corpus possui características específicas e, portanto, a metodologia de um sistema pode ser melhor 
que a de um outro para um dado corpus.

O corpus multi-idioma é muito variável pois aborda diferentes temas em diversos idiomas. Essa característica dificulta a identificação da 
relevância das sentenças. Os sistemas Cortex, Enertex e REG tiveram um desempenho melhor para os documentos de 
pequeno tamanho. Entretanto, o SASI obteve os melhores resultados à medida em que o tamanho dos documentos a serem resumidos 
aumentou. Portanto, o SASI conseguiu mensurar  com eficiência e eficácia a relevância das sentenças em documentos de diferentes temáticas.

O corpus CSTNews é composto por grupos de notícias jornalísticas similares sobre um mesmo acontecimento. Textos jornalísticos normalmente 
possuem as principais informações descritas nas primeiras sentenças e, frequentemente, as principais informações estão presentes em todas 
notícias. A característica do SUMMatrix de identificar e priorizar as informações que estão presentes na maioria dos documentos permite 
produzir resumos com as informações de destaque das notícias. O SUMMatrix não obteve os melhores resultados na avaliação FRESA. Entretanto, 
essa avaliação analisa o resumo com relação ao texto integral de uma notícia, que não é o mais importante para textos jornalísticos pois os 
mesmos podem conter informações irrelevantes ou desnecessárias sobre um acontecimento. A análise de textos jornalísticos foca-se na produção 
de resumos com os dados mais relevantes. Assim, o sumário não precisa ser o melhor na análise do FRESA, mas deve ser o melhor na análise do 
sistema ROUGE, que compara o resumo candidato com os produzidos por humanos. Nesse cenário, o SUMMatrix obteve os melhores resultados no 
sistema ROUGE para o corpus CSTNews.

O corpus RPM é dividido em dois \textit{clusters} por temática. Os textos de cada \textit{cluster} foram selecionados em diferentes 
períodos podendo envolver diferentes acontecimentos. O sistema RAG baseia-se na análise das palavras e na quantidade de sentenças similares 
entre si para avaliar a relevância das mesmas no texto. Apesar das sentenças poderem referenciar acontecimentos diferentes, a temática das 
sentenças pode ser similar e juntamente com a análise das palavras, é possível identificar a importância das sentenças com relação ao 
\textit{cluster}. Baseando-se nessa metodologia, o sistema RAG obteve o segundo melhor desempenho na maioria das métricas quando avaliado 
 pelos sistemas FRESA e ROUGE. 

O corpus DECODA trata de conversas sobre um assunto específico em centrais de atendimento ao cliente. Os textos contém  a narração do 
problema ocorrido e uma possível solução proposta (se foi disponibilizada durante a conversa). As expressões vocais e outros erros de 
conversas transcritas (por exemplo, ``bah'', ``hein'', ``ah'', expressões sem relevância para o texto, abreviações erradas, falhas no 
reconhecimento de algumas palavras, entre outros erros) ocupam espaço no resumo e não possuem importância. Essas expressões 
reduzem a informatividade e seu tratamento propicia uma melhoria da qualidade dos resumos produzidos. Por isso, o sistema LIA-RAG gerou 
sumários legíveis e com as informações mais pertinentes do texto. Os resultados do LIA-RAG foram melhores que os resultados dos sistemas 
NTNU da competição \textit{Multiling} CCCS.
\chapter{Conclusão e Trabalhos Futuros}
\label{ch:conc}

O \acf{PLN} se tornou fundamental para a sociedade devido à quantidade de informações presentes no cotidiano. Atualmente, há um grande 
investimento em todas os campos de estudo do \ac{PLN}, com destaque para a área de \acf{SAT}. Empresas como a Google estão investindo nesse 
domínio e nas facilidades obtidas com sua utilização.

Os sistemas desenvolvidos neste trabalho (LIA-RAG, RAG, SASI e SUMMatrix) conseguiram produzir resumos compreensíveis com as principais 
sentenças dos documentos originais e em diferentes idiomas. Com relação aos demais sistemas sumarizadores referenciados na literatura, os 
sistemas desenvolvidos tiveram excelente desempenho, posicionando-se entre os melhores (que fornecem resumos mais informativos).

Mesmo produzindo resumos com boa informatividade, os resumos por extração são compostos por sentenças isoladas e nem sempre é possível 
obter uma construção sintática e semântica coerente entre as sentenças. Os resumos por abstração possuem uma melhor estrutura sintática e 
semântica, mas sua dificuldade reside em construir resumos para um cenário multi-idioma.

\section{Trabalhos futuros}
\label{sc:tf}

Os próximos trabalhos visam desenvolver um sumarizador \textit{cross-lingue} envolvendo os idiomas Espanhol, Francês, Inglês e Português 
em contexto multicultural. A relevância multicultural é fundamental para analisar documentos de diferentes regiões, pois palavras similares 
podem apresentar diferentes significados em regiões distintas. Assim, pretende-se analisar o idioma do utilizador do sistema de forma a 
produzir automaticamente um resumo adaptado à sua língua e cultura.

Cogita-se a utilização da representação \textit{Word Embedding}\footnote{\textit{Word Embedding} é o conjunto de modelagens de um idioma e 
suas técnicas de aprendizagem em \ac{PLN}, em que as palavras e as frases são mapeadas para vetores reais em um espaço contínuo de baixa 
dimensionalidade.} para analisar e representar um \textit{Big Data}\footnote{\textit{Big Data} é um conjunto de dados muito grande e/ou 
complexo.} de documentos a fim de mensurar a relevância das sentenças e seu significado em diferentes idiomas. Serão utilizados, 
igualmente, técnicas de sumarização por abstração no idioma do utilizador do sistema. As sentenças serão produzidas por meio dos processos 
de compressão e fusão multi-frases objetivando-se uma melhora na sua concisão e informatividade. Concomitante a essas técnicas, pretende-se 
desenvolver alguns métodos heurísticos/meta-heurísticos e híbridos buscando maximizar a informatividade e a legibilidade do resumo.

\singlespacing

\bibliographystyle{abnt-alf}
\bibliography{referencias}

\end{document}